%% file: main.tex
\documentclass{article}

    \usepackage[final]{neurips_2025}

\usepackage[utf8]{inputenc} 
\usepackage[T1]{fontenc}    
\usepackage{hyperref}       
\usepackage{url}            
\usepackage{booktabs}       
\usepackage{amsfonts}       
\usepackage{nicefrac}       
\usepackage{microtype}      
\usepackage[dvipsnames]{xcolor}
\usepackage{amsmath}
\usepackage{amssymb}
\usepackage{mathtools}
\usepackage{amsthm}
\usepackage{graphicx}
\usepackage{tabularx}
\usepackage{svg}
\usepackage{csquotes}
\usepackage{floatrow}
\usepackage{pgf}
\usepackage{adjustbox}
\usepackage{enumitem}
\usepackage{makecell}
\usepackage{array,multirow}
\usepackage{cancel}
\usepackage{wrapfig}
\usepackage{microtype}
\usepackage{xspace}
\usepackage{pifont}
\usepackage{scalerel}
\usepackage{tikz}
\usepackage{colortbl}
\usepackage{subfigure}
\usepackage{soul}
\usepackage[capitalize]{cleveref}
\usepackage[textsize=tiny]{todonotes}
\usepackage{pgfplots}
\pgfplotsset{compat=newest}

\crefname{section}{Sec.}{Secs.}
\Crefname{section}{Section}{Sections}
\Crefname{table}{Table}{Tables}
\crefname{table}{Tab.}{Tabs.}

\theoremstyle{plain}
\newtheorem{theorem}{Theorem}[section]
\newtheorem{proposition}[theorem]{Proposition}

\theoremstyle{definition}

\theoremstyle{remark}

\newcolumntype{Y}{>{\centering\arraybackslash}X}

\newcommand\blfootnote[1]{%
\begingroup
\renewcommand\thefootnote{}\footnote{#1}%
\addtocounter{footnote}{-1}%
\endgroup
}

\newcommand{\mypar}[1]{\vspace{3pt}\noindent\textbf{#1~}}

\definecolor{cliptta_color}{rgb}{1, 0.97, 0.92}
\definecolor{lightgray}{rgb}{0.95, 0.95, 0.95}

\definecolor{grayblue}{rgb}{0.471, 0.569, 0.784}

\hypersetup{
    colorlinks = true,
    linkcolor = {BrickRed},
    citecolor = {grayblue},
    urlcolor = {grayblue},
    allbordercolors = {white}
} 

\input{math_commands.tex}

\title{\CLIPTTA: Robust Contrastive Vision-Language Test-Time Adaptation}

\author{%
  Marc Lafon$^{\star,1,2}$ 
  \And Gustavo A. Vargas Hakim$^{\star, 3}$
  \And Clément Rambour$^{2}$
  \And Christian Desrosier$^{3}$
  \And Nicolas Thome$^{2,4}$
  \And\\[-2em]
  $^1$\normalsize{Conservatoire National des Arts et Métiers, CEDRIC, F-75141 Paris, France}\\
  $^2$\normalsize{Sorbonne Université, CNRS, ISIR, F-75005 Paris, France}\\
  $^3$\normalsize{ETS Montreal, Canada} \\
  $^4$\normalsize{Institut universitaire de France (IUF)}
}

\begin{document}

\maketitle

\input{sec/abstract}
\input{sec/introduction}

\input{sec/related_work}
\input{sec/method}

\input{sec/experiments}
\input{sec/conclusion}


\bibliography{main.bib}
\bibliographystyle{unsrt}



\newpage
\clearpage

\appendix

\input{sec/supplementary}

\end{document}

%% file: math_commands.tex
\usepackage{amsmath,amsfonts,bm}
\usepackage[dvipsnames]{xcolor}

\newcommand{\CLIPTTA}{C\textsc{lip}TTA\xspace}

\newcommand{\CLIPTTAb}{\textbf{C\textsc{lip}TTA}\xspace}

\def\eqref#1{equation~\ref{#1}}

\def\1{\bm{1}}

\newcommand{\tr}[1]{{#1^{\top}}}

\def\rvz{{\mathbf{z}}}

\def\vt{{\bm{t}}}

\def\vx{{\bm{x}}}

\def\vz{{\bm{z}}}

\def\evp{{p}}

\DeclareMathAlphabet{\mathsfit}{\encodingdefault}{\sfdefault}{m}{sl}
\SetMathAlphabet{\mathsfit}{bold}{\encodingdefault}{\sfdefault}{bx}{n}

\newcommand{\Ls}{\mathcal{L}}

\DeclareMathOperator*{\argmax}{arg\,max}

\RequirePackage{xspace}
\makeatletter
\DeclareRobustCommand\onedot{\futurelet\@let@token\@onedot}
\def\@onedot{\ifx\@let@token.\else.\null\fi\xspace}

\def\eg{\emph{e.g}\onedot} 

\def\ie{\emph{i.e}\onedot}

\makeatother

%% file: sec/abstract.tex
\begin{abstract}

Vision-language models (VLMs) like CLIP exhibit strong zero-shot capabilities but often fail to generalize under distribution shifts. Test-time adaptation (TTA) allows models to update at inference time without labeled data, typically via entropy minimization. However, this objective is fundamentally misaligned with the contrastive image-text training of VLMs, limiting adaptation performance and introducing failure modes such as pseudo-label drift and class collapse. We propose \CLIPTTA, a new gradient-based TTA method for vision-language models that leverages a soft contrastive loss aligned with CLIP’s pre-training objective. We provide a theoretical analysis of \CLIPTTA’s gradients, showing how its batch-aware design mitigates the risk of collapse. We further extend \CLIPTTA to the open-set setting, where both in-distribution (ID) and out-of-distribution (OOD) samples are encountered, using an Outlier Contrastive Exposure (OCE) loss to improve OOD detection. Evaluated on 75 datasets spanning diverse distribution shifts, \CLIPTTA consistently outperforms entropy-based objectives and is highly competitive with state-of-the-art TTA methods, outperforming them on a large number of datasets and exhibiting more stable performance across diverse shifts. Source code is available at:  \href{https://anonymous.4open.science/r/CLIPTTA-682D/README.md}{CLIPTTA Repository.} 
\blfootnote{$^\star$ \text{Equal contribution} }
\blfootnote{~Corresponding author: marc.lafon@lecnam.net}

\end{abstract}

%% file: sec/introduction.tex
\section{Introduction}
\label{sec:intro}

Vision-language models (VLMs), such as  CLIP \cite{clip} and ALIGN \cite{align}, are multimodal foundation models with strong zero-shot performance in downstream classification tasks. Yet, their ability to generalize to specialized domains, \eg, medical imaging or corrupted inputs, remains limited without adaptation, making this an active area of research.

Test-Time Adaptation (TTA) addresses the adaptation of pre-trained models to new downstream tasks during inference, without access to ground-truth labels, typically by updating model parameters via gradient-based optimization~\cite{tpt,tda,calip,clipartt,watt}. This label-free adaptation is particularly valuable for deploying VLMs in real-world applications where annotation is scarce and costly, such as medical image processing \cite{onthefly}, human-robot interaction \cite{posetta}, and federated learning \cite{federatedtta}.

\begin{figure}[!t]
\setlength{\belowcaptionskip}{-15pt}
\centering
\scalebox{1.0}{
    \includegraphics[width=\linewidth]{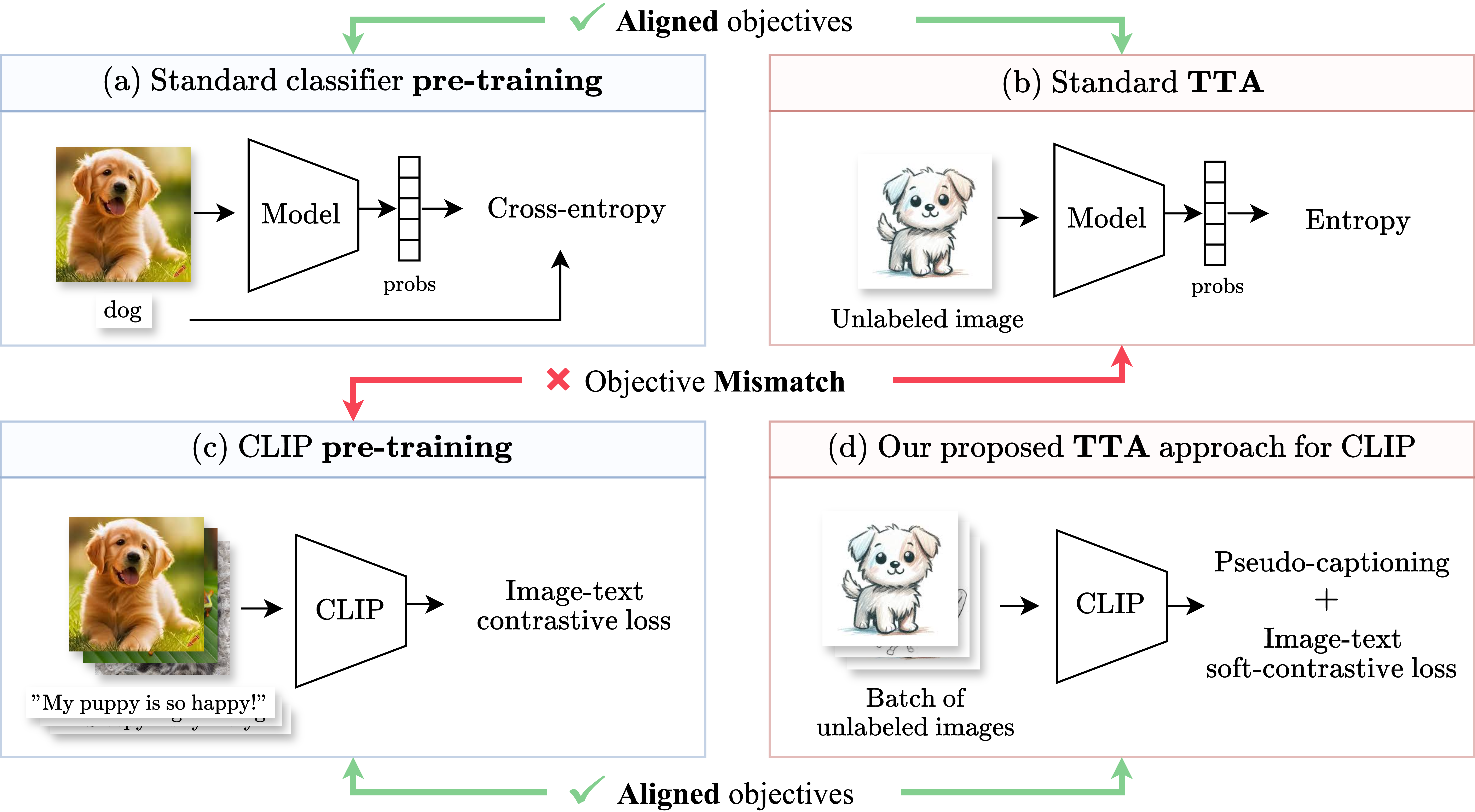}
}
\caption{\textbf{Motivation for \CLIPTTA}. Standard test-time adaptation (TTA) methods often rely on entropy minimization (b), which aligns with the cross-entropy loss used in classifier training (a) but is misaligned with CLIP’s contrastive pre-training (c), hindering adaptation due to incompatible gradient dynamics. \CLIPTTA instead uses a soft contrastive loss aligned with CLIP’s objective, reinforcing alignment between images and their predicted pseudo-captions within the batch (d). Our gradient analysis shows that this contrastive, batch-aware formulation improves robustness to pseudo-label drift and class collapse—two failure modes common to entropy-based TTA methods.}
\label{fig:intro}
\end{figure}

Entropy minimization is the most common TTA objective~\cite{tent,eta,sar,rotta}, as it mirrors the cross-entropy training of standard classifiers. However, it is fundamentally misaligned with the contrastive image-text pre-training objective of VLMs like CLIP, as illustrated in \cref{fig:intro}, potentially hindering adaptation due to differing gradient dynamics. ~Recent works have attempted to improve TTA of CLIP by leveraging visual-textual similarities in a transductive manner \cite{clipartt, watt}, yet the objective misalignment remains unresolved. Furthermore, when labeled data is available, recent work on fine-tuning \cite{flyp} demonstrates that using the exact same loss function as during CLIP pre-training leads to better performance on downstream tasks.

This mismatch in objectives leads to fundamental issues during adaptation: entropy-minimization methods are prone to \emph{pseudo-label drift}, where the model reinforces its own mistakes. This can lead to \emph{class collapse}, where predictions concentrate on a narrow set of classes regardless of the input \cite{survey_collapse,sar}, severely hindering adaptation. Numerous efforts have been made to reduce the adverse impact of pseudo-label misclassification \cite{eta,sar,rotta,tpt}. However, these methods make predictions for each sample independently, without accounting for other predictions in the batch, which limits their robustness. This becomes especially critical when the source model’s accuracy is low or when input batches contain out-of-distribution (OOD) samples that belong to unknown classes \cite{ostta,eta,realm}.

Together, these observations raise a central question:

$\quad$ \emph{How to design an adaptation loss that is more suited for gradient-based test-time adaptation of contrastive vision-language models such as CLIP?}\\

In this work, we introduce \CLIPTTA, a new test-time adaptation method tailored to vision-language models. It employs a soft contrastive image-text loss that mirrors CLIP’s pre-training objective, providing natural continuity in adaptation. As illustrated in \cref{fig:intro}, this design reflects our central assumption: adaptation losses should align with the model’s multimodal contrastive training paradigm. Importantly, the contrastive nature of the \CLIPTTA loss links predictions within a batch, incorporating mechanisms to mitigate the risk of class collapse caused by noisy pseudo-labels. It also demonstrates increased robustness in open-set scenarios, where both in-distribution (ID) and out-of-distribution (OOD) samples are present. We further augment it with a discriminative loss to separate ID from OOD samples, improving performance under open-set conditions.

Our contributions can be summarized as follows:
\vspace{1em}
\begin{itemize}[leftmargin=2em, topsep=1pt, noitemsep]
  \item We introduce \CLIPTTA, a new TTA method for CLIP based on a soft contrastive image-text loss aligned with its pre-training objective, offering a principled alternative to entropy minimization.

  \item We provide a theoretical analysis of C\textsc{lip}TTA’s gradients, showing how its batch-aware design improves robustness to pseudo-label drift and class collapse—two key failure modes of standard gradient-based TTA methods.

  \item We extend \CLIPTTA to open-set adaptation with an Outlier Contrastive Exposure (OCE) loss, improving ID/OOD separation and robustness under distribution shift.
\end{itemize}

We conduct extensive benchmarking across 75 diverse datasets, spanning four types of distribution shifts: corruptions, domain shifts, coarse-grained, and fine-grained classification. Empirical results show that our soft contrastive loss consistently outperforms entropy-based objectives for gradient-based TTA of vision-language models, establishing it as a more effective alternative. In addition, \CLIPTTA is highly competitive with state-of-the-art TTA methods, outperforming them on a large number of datasets and exhibiting more stable performance across diverse shifts. It also achieves notable gains in accuracy and OOD detection under open-set conditions.

%% file: sec/related_work.tex
\section{Related work}
\label{sec:related}

\paragraph{Test-time adaptation} (TTA) seeks to adapt a model to new datasets \emph{on the fly} in the absence of labels. This process is performed on independent data streams that showcase only a small portion of the full data distribution. Aiming to adapt deep classifiers to new domains, TENT~\cite{tent} proposed the widely exploited technique of entropy minimization. The entropy loss is chosen for its link with cross-entropy, with the intent of extending the model's training in an unsupervised way. Building on this principle, several approaches have been proposed: filtering out unimportant samples based on an entropy criterion in ETA~\cite{eta}, and further filtering those with small gradients in SAR~\cite{sar}, minimizing the marginal distribution's entropy across image transformations in MEMO~\cite{memo}, meta-learning the TENT loss via conjugate pseudo-labels~\cite{conjugate}, storing the most confident samples in memory for a \emph{cleaner} adaptation in RoTTA~\cite{rotta}, or combining entropy minimization with a clustering loss constraint in TTC~\cite{ttc}. While these methods rely on additional mechanisms such as filtering or confidence-based selection, \CLIPTTA achieves robustness to pseudo-label drift and collapse by a simple modification of the adaptation objective.
~Contrastive learning approaches have also been explored, such as AdaContrast~\cite{adacontrast}, where a student-teacher model is trained using pseudo-labels obtained from weak and strong image augmentations as in MoCo~\cite{moco}. In contrast, our contrastive adaptation refers to visual-text interactions in the context of VLMs. To the best of our knowledge, this is the first attempt to explore this particular contrastive TTA formulation for VLMs.

\paragraph{TTA for VLMs.} Several methods have been proposed to adapt VLMs to new streams of unseen data. CLIPArTT~\cite{clipartt} introduces a new loss function specifically tailored to VLMs, combining image-to-image and text-to-text similarities to generate pseudo-labels and utilizing a small subset of probable classes to form new image-wise text prompts. WATT~\cite{watt} extends this idea with prompt ensembling and weight averaging. While CLIPArTT's loss better leverages CLIP’s multimodal structure than entropy minimization, it remains heuristically driven and loosely aligned with CLIP’s contrastive training objective. Complementary to these, other methods explore alternative adaptation paradigms. TPT~\cite{tpt} performs adaptation through prompt tuning~\cite{prompt1}: rather than updating the model’s internal weights, it optimizes a small set of text prompts using entropy minimization. Although it uses gradient-based adaptation, this approach is fundamentally distinct from traditional TTA methods that typically update normalization parameters, and it comes with a high computational cost due to its reliance on multiple augmentations per image. TDA~\cite{tda} adopts an even more distinct approach: it operates in a gradient-free manner by building positive and negative caches of past predictions, which are then used as pseudo-labels to simulate few-shot episodes as in \cite{tip}. While TDA achieves strong results on Imagenet variants, we found it to perform poorly under other types of distribution shifts, such as corruptions. In contrast, our approach, \CLIPTTA, requires only a simple modification of the loss function and delivers robust performance across all TTA benchmarks.

\paragraph{Open-set TTA} is a more challenging branch of TTA, where batches are polluted with out-of-distribution (OOD) samples that belong to unknown classes.
~Open-set TTA methods aim at detecting OOD samples from in-distribution (ID) samples, and improve the model's accuracy on ID images. OSTTA~\cite{ostta} uses an entropy heuristic based on a student-teacher model to disregard OOD samples and apply entropy minimization on the ID ones. SoTTA~\cite{sotta} uses the maximum predicted probability to filter and store the most confident samples in memory, and applies TENT on them. On the contrary, STAMP~\cite{stamp} filters samples and their augmentations via entropy, to also preserve them in a memory for entropy minimization. UniEnt~\cite{unient} addresses the problem more explicitly by modeling the samples' outlier score as a mixture of two Gaussian distributions, later using entropy minimization on the ID samples and entropy maximization on the OOD samples. As in the closed-set scenario, these methods do not transfer optimally to VLMs, since entropy does not connect well with CLIP's pre-training loss. Our adaptation loss aligns with CLIP's pre-training, and we propose a discriminative OOD loss that directly aligns with ID/OOD separations metrics.

%% file: sec/method.tex
\section{\CLIPTTA }
\label{sec:method}

\begin{figure*}[t]
\centering
\begin{tikzpicture}[scale=1, every node/.style={scale=1}]
    \node () at (0, 0) {\includegraphics[width=\textwidth]{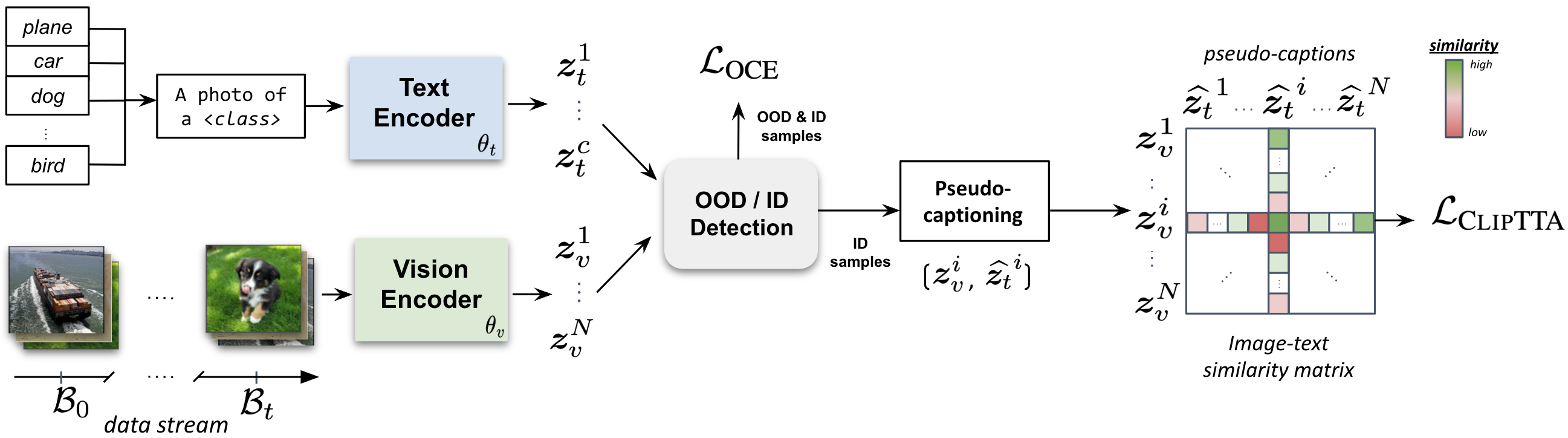} }; 
    \node () at (0.4, 1.3) {\tiny{~Eq.~(\ref{eq:oce})}};
    \node () at (6.375, -0.3) {\tiny{~Eq.~(\ref{eq:cliptta_loss})}};
\end{tikzpicture}
\vspace{-1.5em}
\caption{\textbf{Illustration of \CLIPTTA}. \CLIPTTA in \cref{seq:cliptta_loss} consists of a soft contrastive loss specifically designed for TTA of VLMs like CLIP. We show in \cref{sec:CLIPTTA-analysis} that \CLIPTTA is robust to class collapse and pseudo-label errors. Finally, we add an OCE loss to be robust to OOD samples in batches in \cref{seq:oce_loss} and improve the ID/OOD detection and accuracy in open-set scenarios. 
}
\label{fig:cliptta}
\end{figure*}
\vspace{-0.5em}
We introduce \CLIPTTA, a contrastive test-time adaptation method tailored to VLMs such as CLIP, as illustrated in \cref{fig:cliptta}. By aligning the adaptation objective with CLIP’s image-text contrastive pre-training described in \cref{seq:cliptta_loss}, \CLIPTTA improves robustness to pseudo-label errors and class collapse through its batch-aware formulation, as demonstrated by our gradient analysis in \cref{sec:CLIPTTA-analysis}. Combined with the Outlier Contrastive Exposure loss introduced in \cref{seq:oce_loss}, it improves both OOD detection and accuracy for robust adaptation in open-set scenarios.

\subsection{Contrastive adaptation loss at test-time}
\label{seq:cliptta_loss}

Let us denote CLIP's visual encoder as $f_{\theta_v}^v(\cdot)$ and its textual encoder as$f_{\theta_t}^t(\cdot)$, with model parameters $\theta = (\theta_v, \theta_t)$. Given an image $\vx$ and a textual prompt $\vt$, the normalized visual and text features are $\vz_v = f_{\theta_v}^v(\vx)$ and $\vz_t = f_{\theta_t}^t(\vt)$. To classify an image in a downstream task, we construct class-specific captions of the form ${\vt_c=\text{``A photo of a} <\text{class}>\!\text{''}}$  for each class $c$, and compute the probability of classifying image $\vx_i$ as class $c$: \vspace{-0.2em}
\begin{equation}
q(\vt_c | \vx_i) = \frac{\exp(\tr{{\vz_v^i}}\!\vz_t^c/\tau)}{\sum_{k=1}^C\exp(\tr{{\vz_v^i}}\!  \vz_t^k/\tau)},
    \label{eq:clip_prob}
\end{equation}
\vspace{-0.2em}
\noindent where $\tau$ is a temperature parameter.

Since ground truth captions are unavailable at test-time, we generate pseudo-captions for a batch of  $N$  samples $\{\vx_i\}_{i=1}^N$ by associating each image $\vx_i$ to the caption of its predicted class ${\hat{\vt}_i = \vt_{\hat{c}}, \text{ where }  \hat{c}=\argmax_{c}q(\vt_c | \vx_i)}$. We denote $\widehat{\vz_t}^i$ the representation of $\hat{\vt}_i$. Given two pseudo-labeled image-text pairs $(\vx_i, \widehat{\vt}_i)$ and $(\vx_j, \widehat{\vt}_j)$, we define $p(\hat{\vt}_j | \vx_i)$ and $p(\vx_j | \hat{\vt}_i)$ as the probabilities that $\vx_i$ matches $\hat{\vt}_j$ and that $\hat{\vt}_i$ matches $\vx_j$, respectively:

\begin{equation}
    p(\hat{\vt}_j | \vx_i) = \frac{\exp(\tr{{\vz_v^i}}\!\widehat{\vz_t}^j/\tau)}{ \sum_{l=1}^N \exp(\tr{{\vz_v^i}}\!\widehat{\vz_t}^l/\tau) } \quad \text{and} \quad p(\vx_j |\hat{\vt}_i) = \frac{\exp(\tr{{\vz_v^j}}\!\widehat{\vz_t}^i/\tau)}{ \sum_{l=1}^N \exp(\tr{{\vz_v^l}}\!\widehat{\vz_t}^i/\tau) }.
    \label{eq:clip_batch_vt}
\end{equation}

Although \cref{eq:clip_prob} and \cref{eq:clip_batch_vt} appear similar, \textit{they differ in their softmax normalization}: \cref{eq:clip_prob} normalizes over $C$ classes, while \cref{eq:clip_batch_vt} normalizes over the $N$ predicted classes in the batch. 

A natural strategy for adapting CLIP at test time is to reuse its contrastive loss on pseudo-labeled image-text pairs $(\vx_i, \widehat{\vt}_i)$. However, this assumes pseudo-labels are correct and ignores uncertainty in the predictions. Instead, we retain alignment with CLIP’s training objective while relaxing reliance on hard pseudo-labels. To this end, we introduce a soft contrastive loss that leverages the full distribution over pseudo-captions:
\vspace{-0.2em}
\begin{align}
\begin{split}
\mathcal{L}_{\text{s-cont}}(\theta) \coloneqq \sum_{i=1}^N \Big[&\underbrace{-\sum_{j=1}^N p(\hat{\vt}_j |\vx_i)\log{p(\hat{\vt}_j |\vx_i)}}_{\text{\scriptsize image$\rightarrow$text}}  \quad \underbrace{-\sum_{j=1}^N p(\vx_j |\hat{\vt}_i)\log{p(\vx_j |\hat{\vt}_i)}}_{\text{\scriptsize text$\rightarrow$image}}\Big].
\label{eq:scont_loss}
\end{split}
\end{align}

This loss retains CLIP’s contrastive structure while explicitly modeling uncertainty in pseudo-labels. As shown in ~\cref{fig:cliptta}, the first term computes the entropy over the image-to-text probability distribution (row-wise), and the second term the entropy over the text-to-image probability distribution (column-wise) within the batch. Analogous to entropy minimization, which replaces hard cross-entropy with a soft and uncertainty-aware loss, our soft contrastive loss is a principled extension of the VLMs' contrastive scheme. Furthermore, it demonstrates enhanced robustness to pseudo-label errors, as studied in \cref{sec:CLIPTTA-analysis}. To ensure fair comparisons, we use only the image-to-text term of Eq.~(\ref{eq:scont_loss}) in the main experiments, as most gradient-based TTA methods update only the visual encoder. The effect of simultaneously updating the text encoder is evaluated in \cref{supp_seq:exp_details}.


\vspace{0.2em}
\paragraph{Final training objective.} Following prior TTA research \cite{SWR, ostta, stamp, rotta}, we also incorporate standard techniques such as entropy regularization and a class-wise confident memory (CCM) to enhance adaptation. The regularization loss, based on negative marginal entropy, diversifies the predictions by uniformizing the prediction distribution across classes. Defining $\bar{q}_c = \frac{1}{N} \sum_{i=1}^{N} q(\vt_c | \vx_i)$ as the batch-wise average probability for class $c$ (\ie, over probabilities in \cref{eq:clip_prob}), the regularization loss is $\mathcal{L}_{\text{reg}}(\theta) = \sum_{c=1}^{C} \bar{q}_c \log \bar{q}_c$. ~The final \CLIPTTA loss integrates the soft-contrastive loss Eq.~(\ref{eq:scont_loss}), the regularization term, and the CCM memory. 
~Memory batches $\mathcal{M}$, equal in size to test batches, are used to compute the adaptation loss:

\begin{equation}
    \mathcal{L}_{\text{\CLIPTTA}}(\theta) = \frac{1}{2} \Big[ \mathcal{L}_{\text{s-cont}}(\theta) + \mathcal{L}_{\text{s-cont}}^{\mathcal{M}}(\theta) \Big] 
    + \lambda_{\text{reg}} \mathcal{L}_{\text{reg}}(\theta),
    \label{eq:cliptta_loss}
\end{equation}

\noindent where $\mathcal{L}_{\text{s-cont}}^{\mathcal{M}}(\theta)$ is the soft-contrastive loss computed on the memory batch, and $\lambda_{\text{reg}}$ controls the regularizer's strength. By averaging the loss over current and memory batches, the method effectively leverages confident past predictions to improve adaptation while reducing sensitivity to noisy data.

\subsection{Gradient Analysis}
\label{sec:CLIPTTA-analysis}

 We analyze the gradient of the soft contrastive loss $\mathcal{L}_{\text{s-cont}}$ to understand how it enables robust test-time adaptation, particularly in the presence of pseudo-label errors and class imbalance. The key insight is that, unlike entropy-based losses, $\mathcal{L}_{\text{s-cont}}$ is batch-aware, allowing the model to dynamically correct prediction errors and reducing the risk of class collapse.\\

\begin{proposition}[Gradient of Soft-Contrastive Loss]
Let $N_k$ be the number of samples in the batch pseudo-labeled as class $k$, and $q_{ik}=q(\vt_k | \vx_i)$ as in \cref{eq:clip_prob}. The gradient of $\mathcal{L}_{\text{s-cont}}$  w.r.t. $\vz_v^i$ is:

\begin{align}
\label{eq:grad_soft_contrastive}
\nabla_{\vz_v^i}\mathcal{L}_{\text{s-cont}} = &\sum_{j=1}^N   \beta_{i,j} [-\widehat{\vz_t}^j +\sum_{k=1}^C w_{k,i}~\vz_t^k], \\
\text{where} \quad \beta_{i,j} = p(\hat{\vt}_j |\vx_i)[1 &+ \log{p(\hat{\vt}_j |\vx_i)}],
\quad \text{and}  \quad  w_{k,i} = \frac{N_k \, q_{ik}}{\sum_{c=1}^C N_c \, q_{ic}}. \nonumber
\end{align}
\end{proposition}

\begin{proof}
See \cref{supp_seq:gradient_analysis}.
\end{proof}

This expression shows that the gradient for sample $\vx_i$ aggregates contributions from all pseudo-captions in the batch, each weighted by $\beta_{i,j}$. Each contribution consists of two effects. The first term, $ -\widehat{\vz_t}^j$, acts as an attractive force pulling $ \vz_v^i$ toward pseudo-caption $\hat{\vt}_j$. In contrast, the second term, $\sum_k w_{k,i} \vz_t^k$, introduces a repulsive force that pushes the embedding away from dominant class directions since classes that are more frequently predicted exert stronger repulsion. 

Importantly, the coefficients $\beta_{i,j}$ are key to allowing the gradient of a sample to point toward a class different from its pseudo-label, enabling error correction by leveraging predictions from other samples in the batch, as illustrated on a toy dataset in \cref{supp_seq:toy_quali_analysis}. They amplify the contribution of confident and semantically similar pairs in the gradient update, allowing the model to rely on more reliable examples. For instance, if $\vx_i$ is misclassified as class $k^{\prime}$ but is close to another sample $\vx_j$ whose pseudo-caption reflects the correct class $k$, a large $\beta_{i,j}$ steers the update toward $\vz_t^{k}$. Such correction is not achievable by sample-wise objectives like TENT, which systematically reinforce the predicted class regardless of its correctness.\\


\begin{proposition}[Gradient Vanishing under Class Imbalance]
As one class k dominates the batch ($N_k \rightarrow N$), the gradient of $\mathcal{L}_{\text{s-cont}}$ vanishes:
\begin{align}
||\nabla_{\vz_v^i} \mathcal{L}_{\text{s-cont}}|| \underset{N_k \rightarrow N}{\rightarrow} 0.
\end{align}
\end{proposition}

\begin{proof}
See \cref{supp_seq:gradient_analysis}.
\end{proof}

To further illustrate, consider a binary classification setting with classes $a$ and $b$, where $a$ is the most predicted class in the batch (i.e., $N_a \gg N_b$). In that case, the gradient in \cref{eq:grad_soft_contrastive} becomes: 

\begin{equation}
     \nabla_{\vz_v^i}\mathcal{L}_{\text{s-cont}} = [\beta_{i,a} q_{ib} - \beta_{i,b} q_{ia}] \frac{ N_a N_b }{N_a q_{ia} + N_b q_{ib}} (\vz_t^{b} -\vz_t^{a}).\\
     \label{eq:grad_binary}
\end{equation}

The magnitude of this gradient depends on batch composition. As class imbalance grows, the coefficient $\frac{N_a N_b}{N_a q_{ia} + N_b q_{ib}}$ becomes smaller, reducing the overall gradient magnitude. 

This self-regulation property acts as a built-in dampening mechanism that slows adaptation before collapse occurs, helping prevent convergence to dominant classes, preserving stable updates, and giving the model a chance to recover from poor pseudo-labeling. In contrast, entropy-based objectives such as TENT continue to reinforce dominant class predictions even as imbalance increases, accelerating collapse rather than preventing it (see derivation in \cref{supp_seq:gradient_analysis}).\\



\subsection{Outlier Contrastive Exposure loss}
\label{seq:oce_loss}

In this section, we extend \CLIPTTA to the open-set setting, where the model is exposed to batches composed of images from both \emph{known} classes (ID samples) and \emph{unknown} classes (OOD samples) during adaptation. Our primary objective is to design an effective ID/OOD filtering mechanism to focus adaptation on ID samples only. For that purpose, we use the MCM \cite{mcm} score, defined as ${s_i = \max_{c} q(\vt_c | \vx_i)}$, which is the most popular OOD scoring function in the context of OOD detection for VLMs. For an input image $\vx_i$, we further define the \emph{outlierness} filtering weight:

\begin{equation}
    w_i \, = \, \text{sigmoid}\big(s_i - \alpha\big),
    \label{eq:weights}
\end{equation}

where $\alpha$ is an adaptive and learnable threshold. Using these weights, an image $\vx_i$ will be considered reliable if $w_i>0.5$ and will be regarded as OOD otherwise.

While effective OOD filtering helps to improve TTA performance in an open-set setting, we argue that we can leverage filtered-out OOD samples to improve the ID/OOD detection performance during adaptation. To this end, we introduce the Outlier Contrastive Exposure (OCE) loss that aims at improving the OOD score separation between ID and OOD samples:

\begin{equation}
    \mathcal{L}_{\text{OCE}} \, =\,  -\Bigg[\underbrace{\frac{\sum_{i=1}^N w_i s_i}{\sum_{i=1}^N{w_i}}}_{\mu_{\text{id}}} - \underbrace{\frac{\sum_{i=1}^N (1-w_i)s_i}{\sum_{i=1}^N (1-w_i)}}_{\mu_{\text{ood}}}\Bigg]^2.
    \label{eq:oce}
\end{equation}

In the open-set scenario, our optimization objective then becomes $\min_{\theta,\alpha} \mathcal{L_{\text{\CLIPTTA}}}  + \lambda_{\text{oce}} \mathcal{L}_{\text{OCE}}$, where we update the parameters of the model $\theta$ and the ID / OOD threshold parameter $\alpha$ in an end-to-end fashion. Our OCE loss differs from the UniEnt loss \cite{unient} since it is purely discriminative, enforcing a more direct separation between ID and OOD features, and since it learns the separation threshold $\alpha$.\\
\vspace{2em}

%% file: sec/experiments.tex
\section{Experiments}
\label{sec:experiments}


\paragraph{Datasets.} \CLIPTTA is evaluated on four families of adaptation benchmarks: corruptions (CIFAR-10/100-C, Imagenet-C) with 15 perturbations,  domain shifts (VisDA-C, PACS, OfficeHome, Imagenet-Domains), semantic datasets, including coarse- (CIFAR-10/100) and fine-grained classification (Imagenet, and 10 datasets from the CLIP zero-shot suite). In total, this represents a thorough evaluation over 75 datasets. A detailed description is provided in \cref{supp_seq:exp_details_baselines_datasets}. In open-set TTA, SVHN and Places-365 serve as OOD counterparts for CIFAR-10/100 and Imagenet, respectively.

\paragraph{Metrics.} We report classification accuracy as the primary performance metric. In the open-set setting, we additionally report the area under the ROC curve (AUC) and the false positive rate at a 95$\%$ of ID true positive rate (FPR95) as OOD detection metrics.

\paragraph{Implementation details.} 
~We use ViT-B/16 as CLIP's backbone in all experiments. 
Adaptation is performed with batches of 128 images using the Adam optimizer and a learning rate of $10^{-4}$ over 10 iterations. Experiments are conducted in a non-episodic manner, \ie, without restoring the model's parameters after each batch. Following the standard TTA protocol, we adapt the affine parameters of the visual encoder’s normalization layers. In the open-set setting, we add 128 OOD images per batch, as done in prior work \cite{unient, stamp}.  The regularization and OCE losses' weights are set to $\lambda_{reg}=1$ and $\lambda_{oce}=1$, respectively. We validate that \CLIPTTA is stable to variations of its hyper-parameters in \cref{supp_seq:exp_details}. Experiments were performed on two NVIDIA V100 32GB GPUs.\\

\begin{table*}[!b]
\setlength\tabcolsep{0pt}
    \caption{\textbf{Comparison with gradient-based TTA methods}. \CLIPTTA outperforms entropy minimization methods \cite{tent,eta,sar,rotta} and CLIP-specific TTA methods based on CLIPArTT's loss \cite{clipartt,watt} on all corruptions and domain shift datasets.}    \vspace{-0.5em}
    \label{tab:top1_corruptions}
    \centering
    \begingroup
   \small
    \begin{tabularx}{\linewidth}{>{\small}l  
    >{\small}Y >{\small}Y>{\small}Y>{\small}Y |
    >{\small}Y >{\small}Y>{\small}Y>{\small}Y >{\small}Y}
    \toprule
       & \multicolumn{4}{c}{Corruptions}  & &\multicolumn{4}{c}{Domain shifts}  \\
       \cmidrule(l{4pt}r{4pt}){2-5}
       \cmidrule(l{4pt}r{4pt}){6-10}
      & {\scriptsize C-10-C}  & {\scriptsize C-100-C} & {\scriptsize Imagenet-C} & {\scriptsize Average}& {\scriptsize VisDA-C} & {\scriptsize PACS} & {\scriptsize OfficeHome} & {\scriptsize Imagenet-D} & {\scriptsize Average} \\
      \midrule
      CLIP \cite{clip} &60.2&35.2&25.5&40.3& 87.1&96.1&82.5&59.4&81.3 \\
      \midrule
      TENT \cite{tent}   &56.4&31.4&17.6&35.1& \underline{89.3}&96.6&83.4&60.2&\underline{82.3} \\
      ETA \cite{eta}    &61.3&38.9&26.8&42.3& 88.3&\underline{96.7}&\underline{84.1}&59.9&\underline{82.3} \\
      SAR \cite{sar}    &67.8&43.2&\underline{33.6}&48.2& 87.8&96.2&83.8&60.6& 82.1 \\
      RoTTA \cite{rotta}  &58.0&33.6&24.6&38.7& 83.7&95.8&82.5&61.6&80.9 \\
      \midrule
      CLIPArTT \cite{clipartt} &\underline{68.1}&\underline{48.0}&33.3&\underline{49.8}& 84.1&96.3&82.0&60.7&80.8 \\
      WATT \cite{watt}  &66.0&38.5&26.0&43.5& 87.7&96.2&83.4&61.8& 82.1 \\
      \midrule
      \rowcolor{cliptta_color} \CLIPTTA \scriptsize{(ours)} &\textbf{80.7}&\textbf{52.6}&\textbf{41.1}&\textbf{58.1}& \textbf{89.6}&\textbf{97.5}&\textbf{84.2}&\textbf{63.4}&\textbf{83.7}\\
    \bottomrule
    \end{tabularx}
    \endgroup
    \vspace{-0.5em}
\end{table*}

\subsection{Main results}
\label{sec:expes_main}

\begin{figure}[!t]
\centering    
\vspace{-0.5em}
\setlength\tabcolsep{0pt}
\begin{tabularx}{\textwidth}{c c}
\includegraphics[scale=0.22]{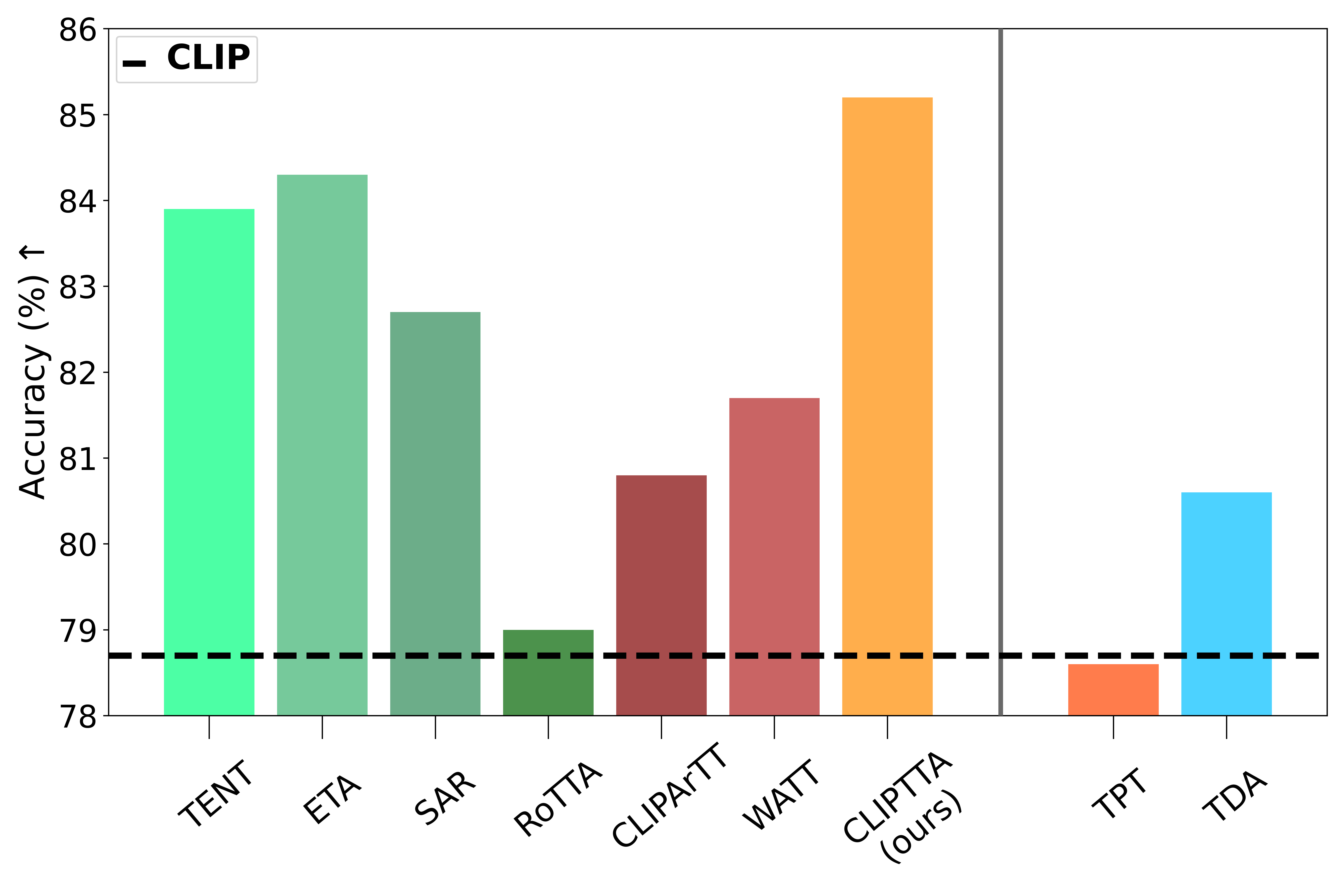}
& \includegraphics[scale=0.22]{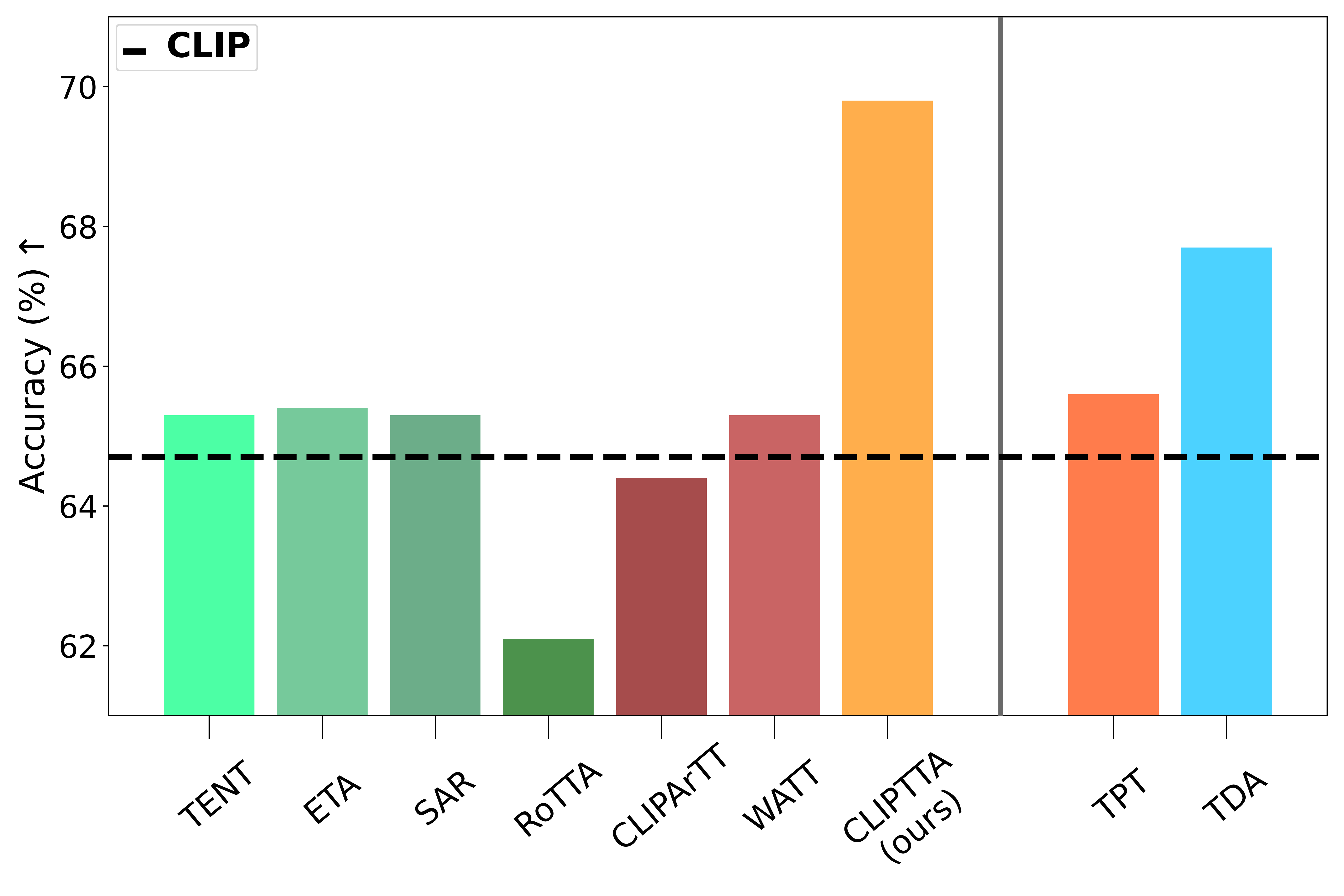} \\[-1ex]
\centering (a)  Coarse-grained results &\centering (b) Fine-grained results
\end{tabularx}
\vspace{-0.75em}
\caption{\textbf{TTA results on semantic datasets}. Top-1 accuracy on coarse-grained datasets (CIFAR-10 and CIFAR-100) (a) and on 11 fine-grained datasets (including Imagenet) (b). Comparison with gradient-based TTA methods \cite{tent,eta,sar,rotta, clipartt,watt} and alternative state-of-the-art TTA methods \cite{tpt,tda}.}
\label{fig:top1_semantic}
\end{figure}

\paragraph{What is the best loss function for TTA of CLIP?} We compare \CLIPTTA with the two prevailing families of test-time objectives:
(i) TENT-style losses, including TENT \cite{tent}, ETA \cite{eta}, SAR \cite{sar} and RoTTA \cite{rotta}, and
(ii) CLIPArTT-derived losses, such as CLIPArTT \cite{clipartt} and WATT \cite{watt}.
Table \ref{tab:top1_corruptions} reports top-1 accuracy under synthetic corruptions and real domain shifts. We highlight two key findings. First, \CLIPTTA substantially improves over zero-shot CLIP, with large gains when initial accuracy is low: +20.5 pts on CIFAR-10-C, +17.4 pts on CIFAR-100-C, and +15.6 pts on Imagenet-C. In contrast,  other methods perform poorly under the same conditions, which can be attributed to the increased likelihood of class collapse and pseudo-label drift when initial accuracy is low. Further analysis and finer-grained experimental evidence are presented in \cref{supp_seq:gradient_analysis} and \cref{supp_seq:toy_quali_analysis}, respectively. Second, \CLIPTTA is the only method that consistently achieves top performance across all benchmarks. While competing methods demonstrate strengths in specific scenarios, they fall short overall. For example, ETA performs best among the TENT-style methods on domain-shift datasets but still lags by 1.4 pts on average. Similarly, CLIPArTT is most competitive on corruption benchmarks but remains 8.3 pts behind. This trend persists across both coarse- and fine-grained datasets (Fig.~\ref{fig:top1_semantic}, with extended results in \cref{supp_seq:exp_details_extended_results}). Altogether, these results establish our soft contrastive loss as the most reliable and broadly effective objective for gradient-based TTA of CLIP.\\

\vspace{1em}

\paragraph{How does \CLIPTTA perform against other CLIP-based TTA methods?} We further benchmark \CLIPTTA against two recent state-of-the-art TTA methods tailored to CLIP: TPT \cite{tpt}, which adapts through text prompt tuning instead of updating normalization parameters, and TDA \cite{tda}, a gradient-free approach that adjusts CLIP’s logits using cached predictions. As shown in Table~\ref{tab:top1_corruptions_tda}, \CLIPTTA improves top-1 accuracy by an average of +19.4 pts over TPT and +15.6 pts over TDA. It achieves state-of-the-art results on nearly all domain-shift benchmarks, with the sole exception of the ImageNet-D suite, where TDA benefits from per-dataset hyperparameter tuning. However, we note that TDA lags far behind on corruption datasets, a standard benchmark in TTA, with an average gap of over 15 pts compared to \CLIPTTA. Figure \ref{fig:top1_semantic} further confirms C\textsc{LIP}TTA’s advantage across both coarse- and fine-grained recognition tasks. While TPT and TDA are designed for single-image batches, our analysis in \cref{supp_seq:exp_details_extended_results} shows that \CLIPTTA remains competitive even in this challenging setting, thanks to its use of the CCM memory and maintains stable performance across a wide range of batch sizes.

\begin{table*}[!b]
\setlength\tabcolsep{0pt}
    \caption{\textbf{Comparison with other CLIP-based TTA methods}. \CLIPTTA outperforms TPT and TDA on most corruptions and domain shifts datasets and is second best on Imagenet-D.}    \vspace{-0.5em}
    \label{tab:top1_corruptions_tda}
    \centering
    \begingroup
   \small
    \begin{tabularx}{\linewidth}{>{\small}l  
    >{\small}Y >{\small}Y>{\small}Y>{\small}Y |
    c 
    >{\small}Y >{\small}Y>{\small}Y>{\small}Y >{\small}Y}
    \toprule
       & \multicolumn{4}{c}{Corruptions}  & &\multicolumn{5}{c}{Domain shifts}  \\
       \cmidrule(l{4pt}r{4pt}){2-5}
       \cmidrule(l{4pt}r{4pt}){7-11}
      & {\scriptsize C-10-C}  & {\scriptsize C-100-C} & {\scriptsize Imagenet-C} & {\scriptsize Average}& & {\scriptsize VisDA-C} & {\scriptsize PACS} & {\scriptsize OfficeHome} & {\scriptsize Imagenet-D} & {\scriptsize Average} \\
      \midrule
      CLIP  \cite{clip} &60.2&35.2&25.5&40.3& &87.1&96.1&82.5&59.4&81.3 \\
      TPT \cite{tpt} &58.0&33.6&24.6&38.7& &85.0&94.0&81.7&62.4&80.8 \\
      TDA \cite{tda}    &63.4&37.4&26.8&42.5& &86.6&96.1&83.0&\textbf{65.0}&\underline{82.8}\\
      \midrule
      \rowcolor{cliptta_color} \CLIPTTA \scriptsize{(ours)} &\textbf{80.7}&\textbf{52.6}&\textbf{41.1}&\textbf{58.1}& &\textbf{89.6}&\textbf{97.5}&\textbf{84.2}&\underline{63.4}&\textbf{83.7}\\
    \bottomrule
    \end{tabularx}
    \endgroup
\end{table*}

\newpage

\newlength{\oldintextsep}
\setlength{\oldintextsep}{\intextsep}

\setlength\intextsep{0pt}
\begin{wraptable}{r}{0.55\textwidth}
\setlength\tabcolsep{0pt}
    \centering
    \begingroup
   \small
    \begin{tabularx}{\linewidth}{>{\small}l  
    >{\small}Y >{\small}Y>{\small}Y
    }
    \toprule
      & {\scriptsize ACC$\uparrow$}  & {\scriptsize AUC$\uparrow$} & {\scriptsize FPR95$\downarrow$} \\
    \midrule
    CLIP   \cite{clip}     &  66.7 & 90.1 & 43.8 \\
    TENT \cite{tent}  &  12.4 & 49.9  & 89.4  \\
    ETA \cite{eta}  &  \underline{67.1} & 89.6  & 46.1 \\
    SAR \cite{sar} &  58.8 & 62.0  & 75.7\\
    CLIPArTT \cite{clipartt} & 31.2  & 61.1  & 87.5\\
    WATT \cite{watt}  & \underline{67.1}  & 87.4  & 53.4\\
    TDA \cite{tda} & 66.8 & 82.1 & 59.8 \\
    \rowcolor{cliptta_color} \CLIPTTA \scriptsize{(ours)}~~  &   \textbf{67.6} & 93.5  & 25.7\\
    \midrule
    OSTTA \cite{ostta} $\dagger$ &  66.9 & 84.9 & 59.2 \\
    SoTTA\cite{sotta} $\dagger$ & 66.7 & 89.3 & 47.1 \\
    STAMP \cite{stamp} $\dagger$  & 29.7& 63.0 & 80.2 \\
    UniEnt \cite{unient} $\dagger$ &   65.2		&  \underline{95.4}		&  \underline{17.1} \\
    \midrule
    \rowcolor{cliptta_color} \CLIPTTA + $\text{\scriptsize{OCE}}$ \scriptsize{(ours)} $\dagger$ ~~ &  \textbf{67.6} & \textbf{97.7} & \textbf{9.7} \\
    \bottomrule
    \end{tabularx}
    \endgroup
    \caption{Open-set TTA results with Imagenet as ID dataset and Places as OOD dataset. $\dagger$ denotes open-set TTA methods.}
    \label{tab:open_set_main}
\end{wraptable}

\paragraph{How does \CLIPTTA perform in the presence of semantic OOD samples?} Table~\ref{tab:open_set_main} presents results on the open-set scenario on Imagenet, where OOD detection needs to be performed alongside classification. First, we note that all closed-set methods, except ours, perform noticeably worse than zero-shot CLIP in OOD detection, highlighting C\textsc{LIP}TTA’s strong robustness to OOD sample contamination during adaptation. Notably, TENT and CLIPArTT suffer severe performance degradation in both classification and OOD detection, likely due to outlier interference in their pseudo-labeling process. Second, when equipped with our OCE loss, \CLIPTTA consistently outperforms specialized open-set TTA methods, which use heuristic OOD detection mechanisms, achieving +2.3 points AUC over UniEnt and +8.4 points AUC over SoTTA. Our soft contrastive objective reliably preserves and improves both accuracy and OOD detection, unlike these entropy-based methods, which tend to degrade CLIP’s initial performance. Additional results on other datasets are reported in \cref{supp_seq:exp_details_extended_results}.\\

\subsection{Model analysis}
\label{subsec:model_an}


\setlength\intextsep{0pt}
\begin{wraptable}{r}{0.5\textwidth}
\setlength\tabcolsep{0pt}
\small
\centering
    \begin{tabularx}{\linewidth}{>{\small}l>{\small}Y >{\small}Y >{\small}Y >{\small}Y >{\small}Y}
        \toprule
        \rowcolor{lightgray} & \scriptsize{C-100} & \scriptsize{C-100-C} & \raisebox{0ex}[0pt]{\scriptsize{IN}}  & \raisebox{0ex}[0pt]{\scriptsize{IN-C}}& \raisebox{0ex}[0pt]{\scriptsize{Avg.}}\\

        \midrule
        \scriptsize{CLIP}  & 68.1  &  35.2  & 66.7 & 25.5 & 48.9\\
        \scriptsize{TENT}   & 72.9 & 31.4 & 66.5 & 17.6& 47.1\\
        \midrule
        $\mathcal{L}_{\text{s-cont}}$& 74.2 & 50.8 & 68.8 & 40.3& 58.5\\
        $\mathcal{L}_{\text{s-cont}}$ \scriptsize{+} $\mathcal{L}_{\text{reg}}$   & 74.9 & 52.4 & 69.1 & 38.6 & 58.8\\
        $\mathcal{L}_{\text{s-cont}}$ \scriptsize{+} $\mathcal{L}_{\text{reg}}$ \scriptsize{+} $\mathcal{M}$ & \textbf{75.3} & \textbf{52.6} & \textbf{69.6} &  \textbf{41.1} & \textbf{59.6}\\
        \bottomrule   
    \end{tabularx}
    \vspace{-0.5em}
\caption{\textbf{Ablation analysis.} Accuracy in the closed-set setting on CIFAR-100, CIFAR-100-C, Imagenet, and Imagenet-C.}
\label{tab:ablations}
\end{wraptable}

\paragraph{Ablation study.} Table~\ref{tab:ablations} presents an ablation of \CLIPTTA’s components on four closed-set benchmarks. The first key observation is that the soft contrastive loss ($\mathcal{L}_\text{s-cont}$) \emph{alone} accounts for the vast majority of the overall performance gains, demonstrating the importance of aligning the adaptation objective with CLIP’s pre-training. In low-accuracy settings, $\mathcal{L}_\text{s-cont}$ significantly outperforms TENT, achieving gains of +19.4 points on CIFAR-100-C and +22.7 points on ImageNet-C, where TENT even degrades CLIP’s performance. This confirms the vulnerability of entropy-based methods to pseudo-label drift and class collapse and supports the enhanced robustness of $\mathcal{L}_{\text{s-cont}}$, as theoretically analyzed in \cref{sec:CLIPTTA-analysis}. On average, $\mathcal{L}_\text{s-cont}$ improves over TENT by +11.4 points. Adding the regularization loss ($\mathcal{L}_\text{reg}$) further improves overall results by +0.3 pts, while incorporating the confident memory ($\mathcal{M}$) brings additional +0.8 pts gains.

\begin{figure}[!t]
\setlength\tabcolsep{0pt}
{\centering
\begin{tabularx}{\textwidth}{c c c}
\includegraphics[width=0.33\linewidth, height=0.28\textwidth]{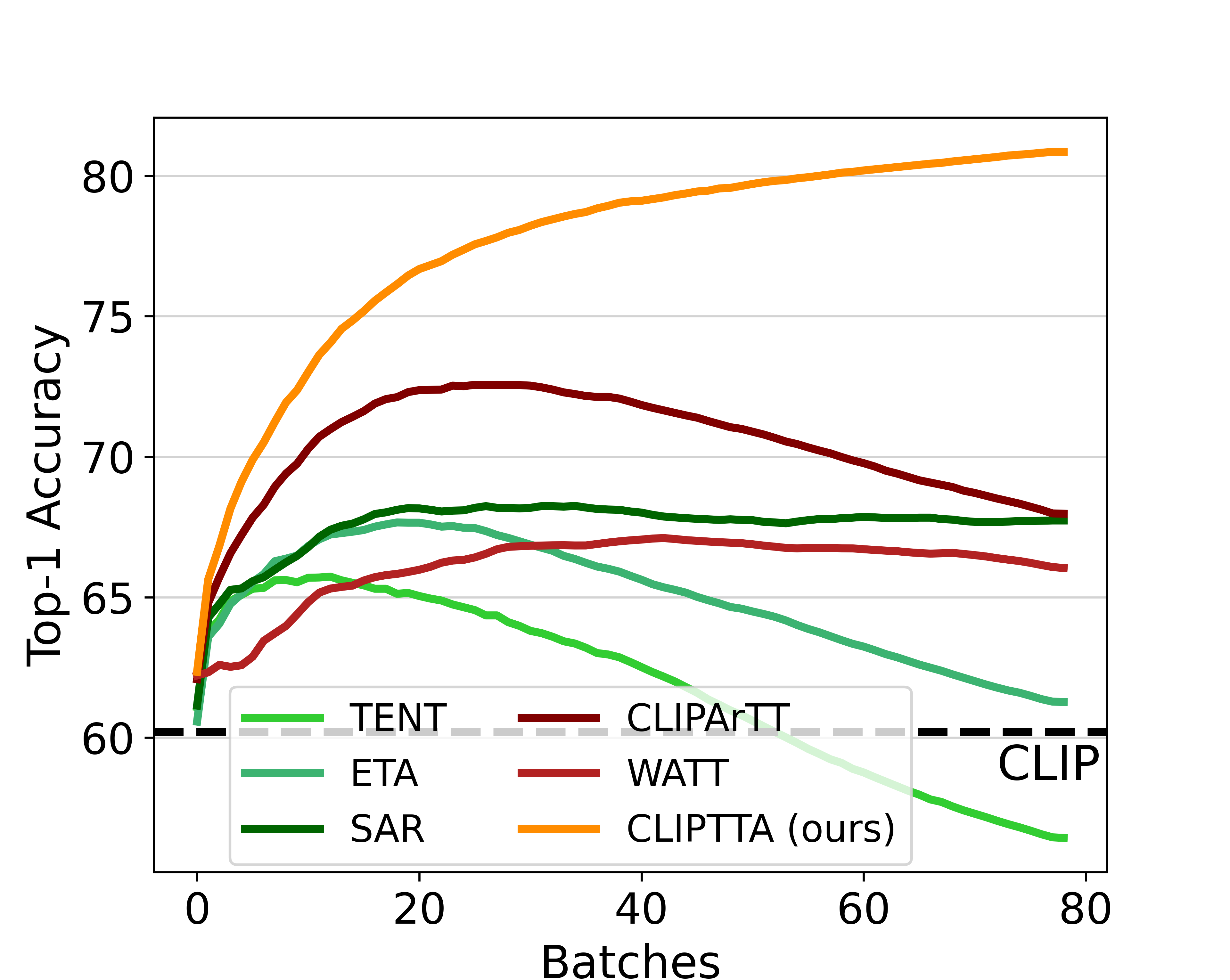}
&
\includegraphics[width=0.33\linewidth, height=0.255\textwidth]{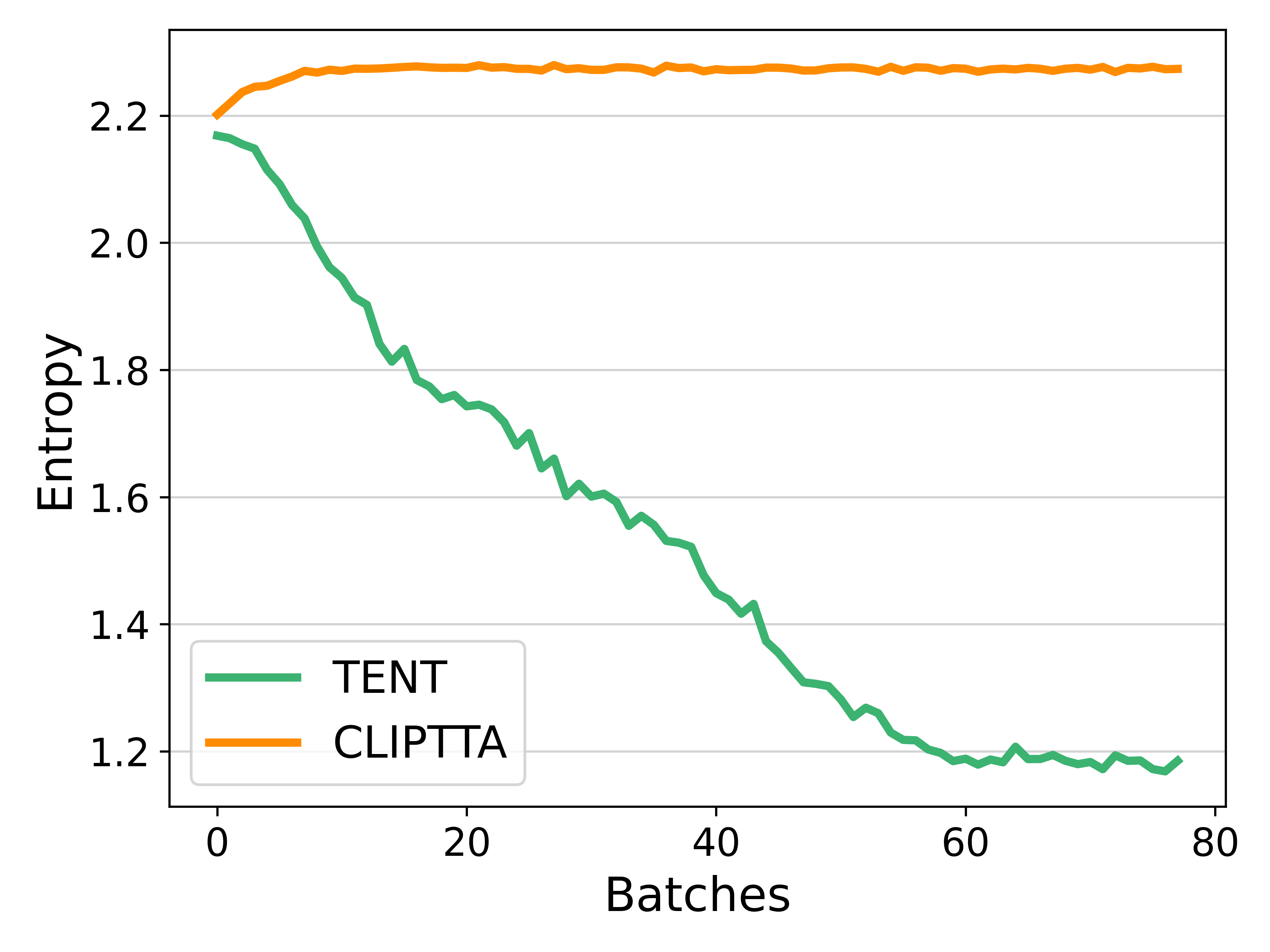}&
\includegraphics[width=0.33\linewidth, height=0.255\textwidth]{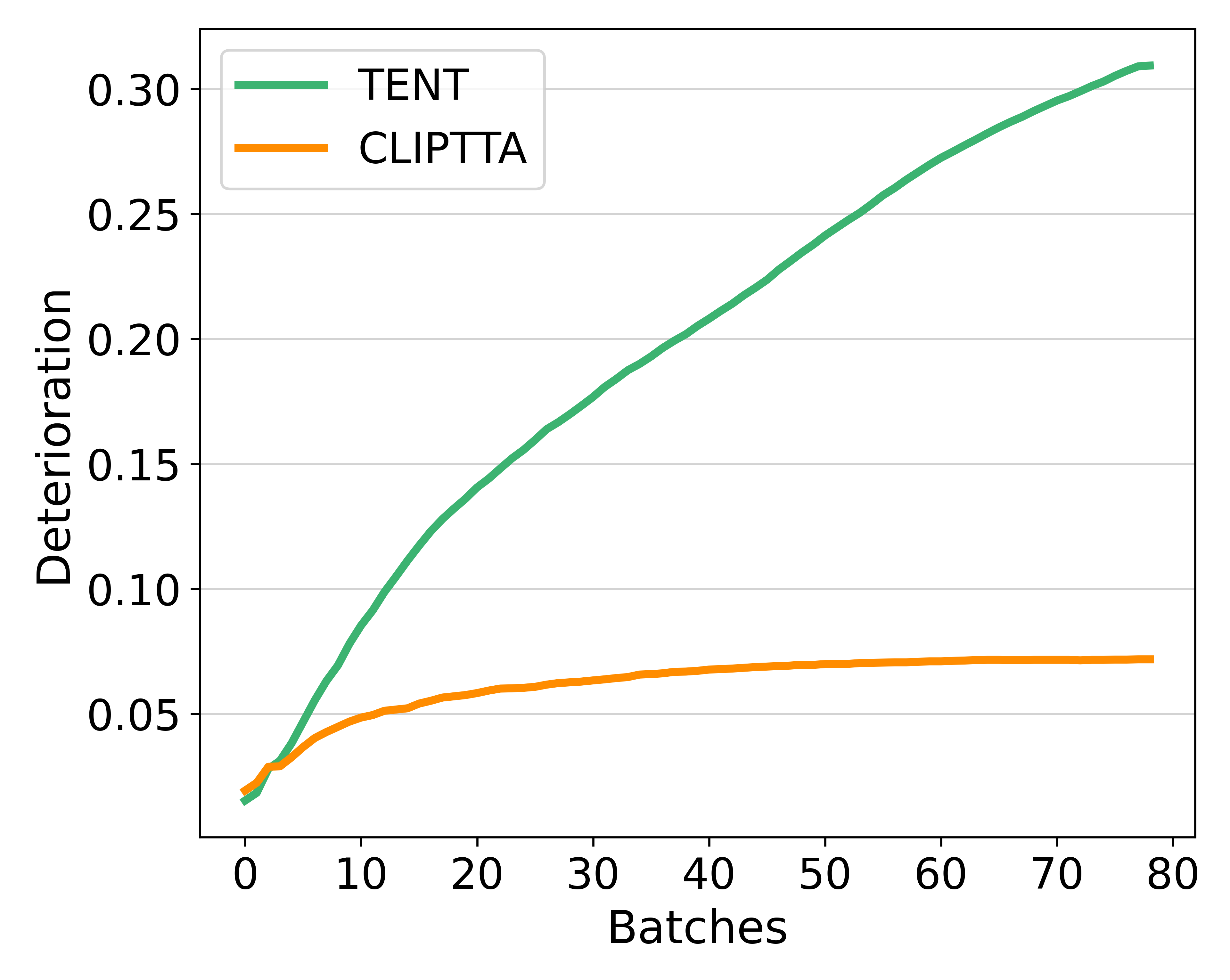}
\\[-0.75ex]
\centering (a)  Accuracy vs batches &\centering (b) Class collapse &  \centering (c) Deterioration ratio
\end{tabularx}}
\vspace{-0.75em}
\caption{\textbf{\CLIPTTA accuracy and robustness on CIFAR-10-C.} (a) In the non-episodic setting, \CLIPTTA steadily improves top-1 accuracy across batches while competing methods degrade. (b) \CLIPTTA maintains high prediction entropy, preserving diversity in predicted classes, whereas TENT shows marked entropy collapse. (c) The deterioration ratio, defined as the fraction of initially correct predictions that become incorrect, increases significantly for TENT but remains low for \CLIPTTA.
}
\label{fig:robustness}
\end{figure}

\paragraph{On \CLIPTTA's robustness.} Figure \ref{fig:robustness} provides empirical insights supporting the gradient analysis in \cref{sec:CLIPTTA-analysis}, by illustrating C\textsc{LIP}TTA’s stability and robustness over batches on CIFAR-10-C. \CLIPTTA is the only method that steadily improves accuracy throughout adaptation while all competing objectives plateau or degrade (Fig.\ref{fig:robustness}a). This stability is closely linked to C\textsc{LIP}TTA’s ability to maintain prediction diversity. As shown in Fig.\ref{fig:robustness}b, prediction entropy remains nearly constant for \CLIPTTA, whereas TENT exhibits a sharp drop in entropy, indicating collapse toward a small subset of classes. This behavior results in harmful label drift. \cref{fig:robustness}c tracks the deterioration ratio, defined as the fraction of initially correct predictions that become incorrect during adaptation. TENT reaches over 30\% deterioration, compared to less than 7\% with \CLIPTTA. These findings confirm the significant stabilizing effect of our batch-aware contrastive loss during adaptation.


%% file: sec/conclusion.tex
\newpage
\section{Conclusion}
\label{sec:conclusion}
\vspace{-0.75em}
This work introduces \CLIPTTA, showing that using a simple soft contrastive loss can be highly beneficial to adapt VLMs in pseudo-label TTA. By a careful analysis of our loss and its gradient, we show that our method brings robustness to the class collapse and pseudo-label drift issues. We also introduce a contrastive outlier exposure loss to tackle the open-set TTA setting. Extensive experiments conducted on a wide range of benchmarks demonstrate that our method significantly outperforms previous baselines on both closed-set and open-set adaptation. Ablation experiments and model analyses strengthen the foundations of our contribution. In our approach, cross-modal interactions are limited to global text-image interactions. Future works then include investigating the link between text and visual adaptation more in depth, and adapting gradient-based TTA for real-time settings, \eg embodied agents.





%% file: sec/supplementary.tex
\clearpage
\setcounter{page}{1}
\setcounter{section}{0}
\renewcommand{\thesection}{\Alph{section}}
\renewcommand{\thesubsection}{\thesection.\arabic{subsection}}

\begin{center}
\textbf{\Large \CLIPTTA: Robust Contrastive Vision-Language Test-Time Adaptation - Appendix}
\end{center}
\vspace{2em}
The appendix is organized as follows. In \cref{supp_seq:theoratical_analysis}, we present a detailed theoretical analysis of the gradients of the $\Ls_{\text{\CLIPTTA}}$ loss. In \cref{supp_seq:toy_quali_analysis}, we provide additional insights into how $\Ls_{\text{\CLIPTTA}}$ helps mitigate collapse and pseudo-labeling errors. Finally, in \cref{supp_seq:exp_details}, we elaborate on the experimental protocol and include additional experimental results.

\section{Theoretical Analysis}
\label{supp_seq:theoratical_analysis}

\subsection{Gradient Analysis}
\label{supp_seq:gradient_analysis}
In this section, we provide a detailed analysis of the gradients of the TENT loss $\mathcal{L}_{\text{TENT}}$, CLIP's contrastive loss $\mathcal{L}_{\text{cont.}}$ (\ie using hard pseudo-captions), our soft contrastive loss $\mathcal{L}_{\text{s-cont.}}$, the regularization loss  $\mathcal{L}_{\text{reg}}$ and the \CLIPTTA loss $\mathcal{L}_{\text{\CLIPTTA}}$. Furthermore, we show how, benefiting from the information of other predictions in the batch, both contrastive losses allow to avoid collapse. Finally, when combined with the regularization loss, \CLIPTTA allows mitigating the effect of pseudo label errors. 

Let's consider a batch of examples $\vx_1, ..., \vx_N$, and let's write $N_k$, the number of predictions assigned to the $k^{th}$ class. To simplify the computations, we place ourselves in the case where only the parameters of the visual encoder are updated.

\paragraph{Gradient of  $\mathcal{L}_{\text{TENT}}$.} We recall that the TENT loss writes as follows:  
\begin{align*}
    \Ls_{\text{TENT}} = -\sum_{k=1}^C q_{ik}\log{q_{ik}}
\end{align*}
with $q_{ik}$ the probability of image $\vx_i$ being classified as class $k$ (see \cref{eq:clip_prob}). The gradient of $\Ls_{\text{TENT}}$ w.r.t.~$\vz_i$ is:
\begin{align}
\begin{split}
    \nabla_{\vz_i}\Ls_{\text{TENT}} = - \sum_{k=1}^C \nabla_{\vz_i} q_{ik}\log{q_{ik}}
                                  &= - \sum_{k=1}^C (1+\log{q_{ik}}) \nabla_{z_i} q_{ik}\\
                                  &= - \sum_{k=1}^C (1+\log{q_{ik}})~q_{ik} \sum_{c=1}^{C} q_{ic}~(\vz_t^k - \vz_t^c)\\
                                  &= - \sum_{k=1}^C ~\big[\sum_{c=1}^C \log{\frac{q_{ik}}{q_{ic}}}~q_{ic}~\big]~ q_{ik} ~ \vz_t^k\\
\end{split}
\label{eq:grad_tent}
\end{align}

From \cref{eq:grad_tent} we can see that the gradient will always push $\vz_i$ in the direction of the predicted class $\hat{k}$ because in that case we have $\log{\frac{q_{i\hat{k}}}{q_{ic}}} > 0, \forall c\neq \hat{k}$. And there is no mechanism allowing to reduce the magnitude of the gradient towards the predicted class even when we are approaching a situation of collapse.

\paragraph{Gradient of  $\mathcal{L}_{\text{cont.}}$} Using the notation introduced in the main paper, let $\hat{\vt}_i$ represent the pseudo-caption associated with the example $\vx_i$ in the batch, and let $p(\hat{\vt}_j | \vx_i)$ denote the probability of $\vx_i$ matching $\hat{\vt}_j$ within the batch. Specifically, we have: 
$$
p(\hat{\vt}_j | \vx_i) = \frac{e^{\tr{\vz_i}\! \widehat{\vz_t}^j}}{\sum_{l=1}^N e^{\tr{\vz_i}\! \widehat{\vz_t}^l}}
$$

The unsymmetrized version of CLIP's contrastive loss writes:

\begin{align*}
    \Ls_{\text{cont.}} &= \sum_{i=1}^N-\log{p(\hat{\vt}_i | \vx_i)}= \sum_{i=1}^N - \tr{\vz_i}\! \widehat{\vz_t}^i + \log{\big( \sum_{j=1}^N e^{\tr{\vz_i}\! \widehat{\vz_t}^j}\big) }= \sum_{i=1}^N - \tr{\vz_i}\! \widehat{\vz_t}^i + \log{\big( \sum_{k=1}^C N_k e^{\tr{\vz_i}\! \vz_t^k}\big) }
\end{align*}

where $\widehat{\vz_t}^j$ is the embedding of the pseudo caption associated with image $\vx_j$ and $\vz_t^k$ is the embedding of class $k$. Let's compute the gradient of $\Ls_{\text{cont.}}$ w.r.t.~$\vz_i$:

\begin{align}
\begin{split}
    \nabla_{z_i}\Ls_{\text{cont.}}= - \widehat{\vz_t}^i + \frac{\nabla_{z_i}\sum_{k=1}^C N_k e^{\tr{\vz_i}\! \vz_t^k}}{\sum_{c=1}^C N_c e^{\tr{\vz_i}\! \vz_t^c}}&= - \widehat{\vz_t}^i + \sum_{k=1}^C \frac{N_k e^{\tr{\vz_i}\! \vz_t^k}}{\sum_{c=1}^C N_ce^{\tr{\vz_i}\! \vz_t^c}}~\vz_t^k\\
    &= - \widehat{\vz_t}^i + \sum_{k=1}^C \frac{N_k e^{\tr{\vz_i}\! \vz_t^k}}{\sum_{c=1}^C N_ce^{\tr{\vz_i}\! \vz_t^c}}\frac{\sum_{c=1}^C e^{\tr{\vz_i}\! \vz_t^c}}{\sum_{c=1}^C e^{\tr{\vz_i}\! \vz_t^c}}~\vz_t^k\\
    &= - \widehat{\vz_t}^i + \sum_{k=1}^C \frac{N_k ~q_{ik}}{\sum_{c=1}^C N_c~q_{ic}}~\vz_t^k\\
    &= - \widehat{\vz_t}^i + \sum_{k=1}^C w_{k,i}~\vz_t^k
\end{split}
\label{eq:grad_clip}
\end{align}

with $w_{k,i} = \frac{N_k ~q_{ik}}{\sum_{c=1}^C N_c~q_{ic}}$.

From \cref{eq:grad_clip}, we observe that CLIP’s contrastive loss consistently drives the visual embedding $\vz_i$ toward the embedding of its predicted class $\widehat{\vz_t}^i$, as $w_{k,i} \leq 1$. However, the gradient’s magnitude is influenced by the proportion of predictions assigned to the same class within the batch. Specifically, as the system approaches a collapse scenario (i.e., $w_{k,i} \rightarrow 1$), the gradient of $\Ls_{\text{cont.}}$ diminishes and eventually vanishes:

\begin{align}
|| \nabla_{z_i}\Ls_{\text{cont.}} || \underset{w_{k,i} \rightarrow 1}{\rightarrow} 0
\end{align}

\paragraph{Gradient of  $\mathcal{L}_{\text{s-cont}}$} The unsymmetrized version of our $\mathcal{L}_{\text{s-cont}}$ loss writes:

\begin{align*}
\mathcal{L}_{\text{s-cont}} = &\sum_{i=1}^N -\sum_{j=1}^N p(\hat{\vt}_j |\vx_i)\log{p(\hat{\vt}_j |\vx_i)}\
\end{align*}

Let's compute the gradient of $\Ls_{\text{s-cont}}$ w.r.t.~$\vz_i$:

\begin{align*}
    \nabla_{z_i}\Ls_{\text{s-cont}} &= -\sum_{j=1}^N \nabla_{z_i}[ p(\hat{\vt}_j |\vx_i)\log{p(\hat{\vt}_j |\vx_i)}] \\
    &= -\sum_{j=1}^N  p(\hat{\vt}_j |\vx_i) \nabla_{z_i}\log{p(\hat{\vt}_j |\vx_i)}  + \log{p(\hat{\vt}_j |\vx_i)} \nabla_{z_i} p(\hat{\vt}_j |\vx_i).
\end{align*}
Using the fact that $\nabla p = p\nabla\log{p}$, we have:
\begin{align*}
    \nabla_{z_i}\Ls_{\text{s-cont}}
    &= -\sum_{j=1}^N  p(\hat{\vt}_j |\vx_i) \nabla_{z_i}\log{p(\hat{\vt}_j |\vx_i)}  + \log{p(\hat{\vt}_j |\vx_i)} p(\hat{\vt}_j |\vx_i) \nabla_{z_i} \log{p(\hat{\vt}_j |\vx_i)}\\
    &= -\sum_{j=1}^N   [1 + \log{p(\hat{\vt}_j |\vx_i)} ]p(\hat{\vt}_j |\vx_i) \nabla_{z_i} \log{p(\hat{\vt}_j |\vx_i)}\\
\end{align*}

Now we can use the fact that $\nabla_{z_i} -\log{p(\hat{\vt}_j |\vx_i)}=-\widehat{\vz_t}^j +\sum_{k=1}^C w_{k,i}~\vz_t^k$ based on the computation $\Ls_{\text{cont.}}$ in \cref{eq:grad_clip}. Therefore, we have:

\begin{align}
    \nabla_{z_i}\Ls_{\text{s-cont}}
    &= \sum_{j=1}^N   \beta_{i,j} [-\widehat{\vz_t}^j +\sum_{k=1}^C w_{k,i}~\vz_t^k]
    \label{eq:grad_scont}
\end{align}
with $\beta_{i,j}=p(\hat{\vt}_j |\vx_i)(1 + \log{p(\hat{\vt}_j |\vx_i)})$.

The gradient of $\Ls_{\text{s-cont}}$ does not solely push the visual embedding toward the predicted class. Instead, it incorporates other predictions within the batch to guide the gradient direction, thereby mitigating the risk of pseudo-labeling errors. However, similar to CLIP’s contrastive loss, the gradient diminishes as we approach a collapse scenario. In the case of collapse, where all examples in the batch are predicted to belong to the same class  $c$ , the following conditions hold: $w_c(\vx_i)=1$ and $w_{k,i}=0, \forall k\neq c$, and $\widehat{\vz_t}^j = \vz_t^c \forall j$. Consequently, the term $[-\widehat{\vz_t}^j +\sum_{k=1}^C w_{k,i}~\vz_t^k]$ cancels out, leading to a null gradient.

\paragraph{Binary classification case.} We derive \cref{eq:grad_binary} in the main paper, starting from \cref{eq:grad_scont}, and assuming that the classification task comprises two classes $K=\{a, b\}$, with $N=N_a + N_b$ as the total batch size. To build on the intuition of the working mechanisms of our soft contrastive loss, we adopt the case where class $a$ is dominant in the batch (\ie, $N_a \gg N_b$). First, we expand on the second sum term inside \cref{eq:grad_scont}, as follows:

\begin{align}
\begin{split}
\sum_{k=1}^C w_{k,i}\vz_i^k = w_{a,i}\vz_t^a + w_{b,i}\vz_t^b  =\frac{N_aq_{ia}}{N_aq_{ia} + Nbq_{ib}}\vz_t^a + \frac{N_bq_{ib}}{N_aq_{ia} + N_b q_{ib}}\vz_t^b = \underbrace{\frac{N_aq_{ia}\vz_t^a + N_b q_{ib}\vz_t^b}{N_aq_{ia} + N_bq_{ib}}}_Q
\end{split}
\end{align}

We notice that we can partition the main sum term in \cref{eq:grad_scont} into two sums that account for the $N_a$ samples predicted as class $a$, and the $N_b$ samples predicted as class $b$:

\begin{align}
    \begin{split}
        \nabla_{z_i}\mathcal{L}_{\text{s-cont}} &= \sum_{j=1}^N \beta_{i,j}[-\hat{\vz}_t^j + Q] = N_a \beta_{ia} [-\vz_t^a + Q] + N_b \beta_{ib}[-\vz_t^b + Q] \\
        &=N_a \beta_{ia} [-\vz_t^a + \frac{N_aq_{ia}}{N_aq_{ia} + N_bq_{ib}}\vz_t^a + \frac{N_b q_{ib}}{N_aq_{ia} + N_bq_{ib}}\vz_t^b] \\
        &+ N_b \beta_{ib}[-\vz_t^b + \frac{N_aq_{ia}}{N_aq_{ia} + N_bq_{ib}}\vz_t^a + \frac{N_b q_{ib}}{N_aq_{ia} + N_bq_{ib}}\vz_t^b]\\
        &=N_a \beta_{ia} [\frac{- N_bq_{ib}}{N_aq_{ia} + N_bq_{ib}}\vz_t^a + \frac{N_b q_{ib}}{N_aq_{ia} + N_bq_{ib}}\vz_t^b] \\
        &+ N_b \beta_{ib}[\frac{-N_aq_{ia}}{N_aq_{ia} + N_bq_{ib}}\vz_t^b + \frac{N_aq_{ia}}{N_aq_{ia} + N_bq_{ib}}\vz_t^a ]\\
        &=  \beta_{ia}q_{ib}\frac{N_aN_b}{N_aq_{ia} + N_bq_{ib}} (\vz_t^{b} -\vz_t^{a}) - \beta_{ib}q_{ia}\frac{N_aN_b}{N_aq_{ia} + N_bq_{ib}} (\vz_t^{b} -\vz_t^{a})\\
        &= [\beta_{i,a} q_{ib} - \beta_{i,b} q_{ia}] \frac{ N_a N_b }{N_a q_{ia} + N_b q_{ib}} (\vz_t^{b} -\vz_t^{a})
    \end{split}
    \label{eq:mid_derivation}
\end{align}

As pointed out, the increasing dominance of class $a$ ($N_b \rightarrow 0)$ reduces the gradient to 0, vanishing the negative effect of class collapse.

\paragraph{Gradient of $\Ls_{\text{reg}}$.} The regularization loss  $\Ls_{\text{reg}}$ writes as:

\begin{equation}
    \mathcal{L}_{\text{reg}} =-\sum_{c=1}^{C} \bar{q}_c \log \bar{q}_c.
\end{equation}

where $\bar{q}_c$ correspond to the average predicted probability for class $c$ inside the batch.

Let's compute the gradient of $\Ls_{reg}$ w.r.t.  ~$z_i$:

\begin{align*}
\nabla_{z_i} \Ls_{reg}  &= \sum_{k=1}^{C}\nabla_{z_i} \overline{\evp}_k \log{\overline{\evp}_k}= \sum_{k=1}^{C} (1 + \log{\overline{\evp}_k})~ \nabla_{z_i} \overline{\evp}_k
\end{align*}

Therefore, we only need to compute $\nabla_{z_i} \overline{\evp}_k$:

\begin{align*}
    \nabla_{z_i} \overline{\evp}_k  &= \nabla_{\rvz_i} \frac{1}{N} \sum_{i=1}^{N}q_{ik} = \frac{1}{N} \nabla_{\rvz_i} q_{ik} = \frac{1}{N} \nabla_{\rvz_i} \frac{e^{\tr{z_i} \vz_t^k}}{\sum_{j=1}^{C}e^{\tr{z_i} \vz_t^j}} = \frac{1}{N} q_{ik} \sum_{j=1}^{C} q_{ij}~[\vz_t^k - \vz_t^j]
\end{align*}

Then we have:

\begin{align}
\begin{split}
\nabla_{z_i} \Ls_{reg} &=  \frac{1}{N} \sum_{k=1}^{C} (1 + \log{\bar{q}_k}) q_{ik} \sum_{j=1}^C q_{ij}(\vz_t^k - \vz_t^j)\\
&=  \frac{1}{N} \sum_{k=1}^{C} [  (1+\log{\bar{q}_k}) q_{ik} \sum_{j\neq k} q_{ij} - q_{ik}\sum_{j\neq k} q_{ij}(1+\log{\bar{q}_j})] ~\vz_t^k\\
  &= \frac{1}{N}  \sum_{k=1}^{C} [\sum_{j=1}^C q_{ij}\log{\frac{\bar{q}_k}{\bar{q}_j}} ]   q_{ik} ~\vz_t^k
\end{split}
\label{eq:grad_reg}
\end{align}

From \cref{eq:grad_reg}, we observe that the gradient is influenced by the ratios $\log{\frac{\bar{q}_k}{\bar{q}_j}}$, driving it towards the classes that are underrepresented in the batch predictions. The use of the regularization loss in conjunction with our soft contrastive loss creates a powerful combined effect, enabling the effective relabeling of misclassified examples, as discussed in \cref{supp_seq:toy_quali_analysis}.

\paragraph{Gradient of $\Ls_{\text{\CLIPTTA}}$.} We recall from the main paper that the final \CLIPTTA loss combines both $\Ls_{\text{s-cont}}$ and $\Ls_{reg}$, thus benefiting both from an enhanced adaptation loss as well a mechanism to combat pseudo-labeling errors (we omit the effect of the memory for simplicity):
\begin{equation}
    \mathcal{L}_{\text{\CLIPTTA}} = \mathcal{L}_{\text{s-cont}}
    + \lambda_{\text{reg}} \mathcal{L}_{\text{reg}},
\end{equation}

Therefore the gradient of $\mathcal{L}_{\text{\CLIPTTA}}$ writes:

\begin{align}
\begin{split}
\nabla_{z_i} \Ls_{\text{\CLIPTTA}} &= \sum_{j=1}^N   \beta_{i,j} [-\widehat{\vz_t}^j +\sum_{k=1}^C w_{k,i}~\vz_t^k] + \lambda_{reg} \frac{1}{N}  \sum_{k=1}^{C} [\sum_{j=1}^C q_{ij}\log{\frac{\bar{q}_k}{\bar{q}_j}} ]   q_{ik} ~\vz_t^k. 
\end{split}
\label{eq:grad_cliptta}
\end{align}

Depending on the composition of the batch, we can see that $\Ls_{\text{\CLIPTTA}}$ will strongly benefit from the contribution of the soft-contrastive loss to provide accurate adaptation, or will be able to correct misclassified examples due to the positive interaction of the combined corrective terms in $\nabla_{z_i} \Ls_{\text{s-cont}}$ and $\nabla_{z_i} \Ls_{reg}$.

\subsection{Analysis of OCE Loss}
\label{supp_seq:oce_analysis}

As discussed in the main paper, our outlier contrastive exposure (OCE) in \cref{eq:oce} of the main paper is a special case of the intra-class variance minimization:

\begin{align}
\begin{split}
    \sigma^2& =\quad p_{\text{id}}\frac{\sum_{i}^{N}w_i(s_i - \mu_{id})^2}{\sum_{i=1}^{N}w_i} + p_{\text{ood}}\frac{\sum_{i}^{N}(1 - w_i)(s_i - \mu_{ood})^2}{\sum_{i=1}^{N}(1 - w_i)}
\end{split}
    \label{eq:ic_var}
\end{align}

\noindent with $p_{\text{id}} = \frac{1}{N} \sum_i w_i$ and $p_{\text{ood}} = \frac{1}{N} \sum_i (1 - w_i)$. This is further condensed into the loss function \cref{eq:intra_class}:

\begin{equation}
    \sigma^2_{\text{intra}} = p_{\text{id}}\mu_{\text{id}}^2 -p_{\text{ood}}\mu_{\text{ood}}^2
    \label{eq:intra_class}
\end{equation}

Here, samples from the same distribution might tend to collapse into a single point. An alternative formulation is inter-class variance maximization, as shown in Eq.~\ref{eq:inter_class}:

\begin{equation}
    \sigma^2_{\text{inter}} = p_{\text{id}}p_{\text{ood}}(\mu_{\text{id}} - \mu_{\text{ood}})^2
    \label{eq:inter_class}
\end{equation}

The impact of the proportions $p_\text{id}$ and $p_\text{ood}$ is twofold. First, when neither ID nor OOD samples are detected, the respective proportion nullifies, and the inter-class variance reaches its minimum. On the contrary, an equilibrium can be reached with both $p_\text{id} = p_\text{ood} = 0.5$, which displays the implicit assumption of equally distributed scores between ID and OOD. We argue that this constraint limits the flexibility of the OOD detection at test time; as the nature of incoming samples is unknown, allowing for a non-uniform distribution in the detection can help filter out less useful samples. Secondly, the product of these probabilities would reduce the scale of the loss, especially compared to the other components of our \CLIPTTA framework, which limits its impact on the adaptation of the model. Hence, a fully contrastive metric can attain the same detection objective by diminishing the latter negative effects:

\begin{equation}
    \sigma^2_{\text{inter}} = (\mu_{\text{id}} - \mu_{\text{ood}})^2
\end{equation}

\section{Discussion on \CLIPTTA's robustness}
\label{supp_seq:toy_quali_analysis}

We further study the properties of \CLIPTTA, to expand the insights on the working mechanisms that assist in its success. Initially, the accuracy across batches (see Fig.~\ref{fig:intro} in the main paper) serves as a straightforward depiction of (a) the general preeminence of \CLIPTTA over other methods, particularly entropy-based techniques, and (b) the collapse effect in methods such as TENT. To elaborate on the underlying advantages of our method, we examine the adaptation process more closely, first in a controlled toy example, then using CIFAR-10-C across all of its corruptions.

\begin{wrapfigure}{r}{0.6\textwidth}
    \centering
    \includegraphics[width=\textwidth]{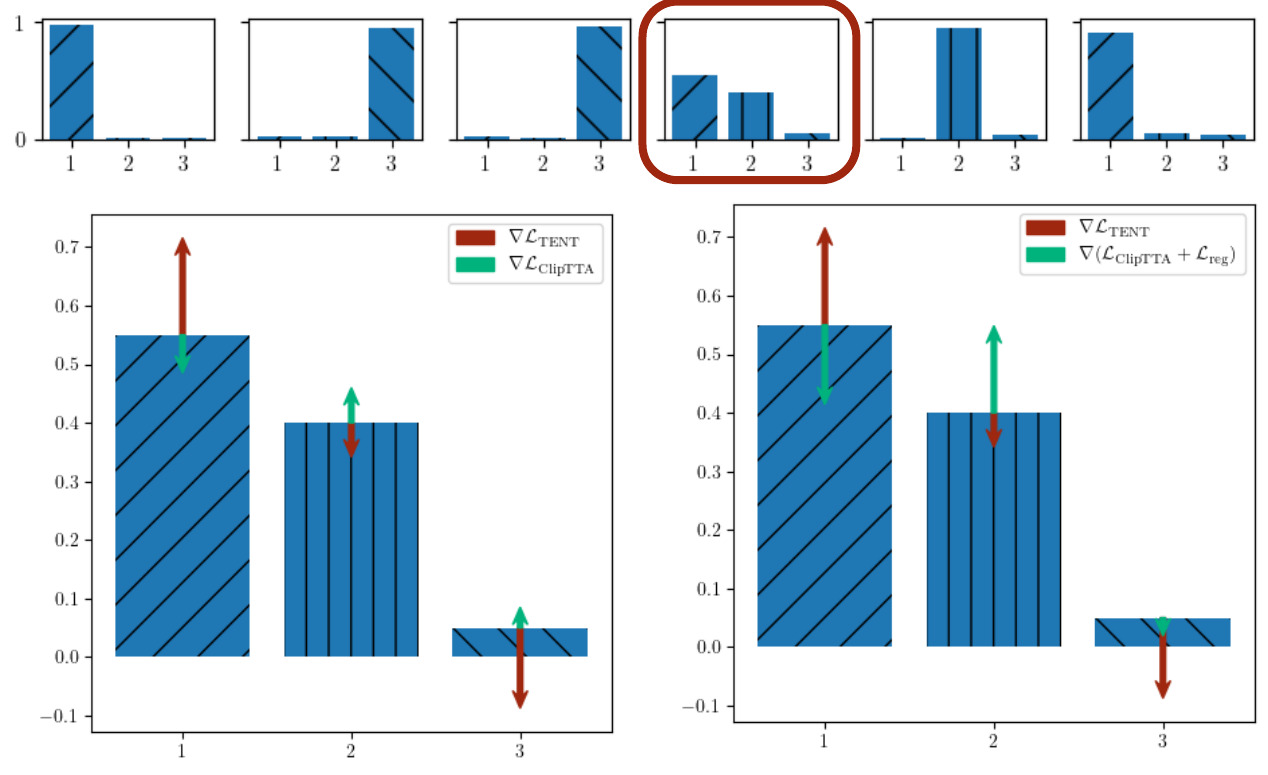}
    \caption{\textbf{Gradient Behavior: TENT vs. \CLIPTTAb} Illustration of gradient directions for TENT, \CLIPTTA, and regularized \CLIPTTA losses on a misclassified sample (circled in red). While TENT (red arrows) reinforces the incorrect prediction to reduce entropy, \CLIPTTA and its regularized version (green arrows) aim to minimize top-2 probability differences, guiding the correction.}
    \label{fig:toy}
\end{wrapfigure}

\mypar{Mitigating pseudo-label errors.}  In \cref{fig:toy}, we present a controlled toy example demonstrating how \CLIPTTA effectively mitigates misclassifications. This example features a batch of six samples in a three-class classification problem. It focuses on the gradient orientations of the TENT, \CLIPTTA, and regularized \CLIPTTA losses for a single misclassified and ambiguous sample. The sample in question exhibits high probabilities for both the predicted and correct labels, indicating low confidence. The gradient of the \CLIPTTA loss is directed toward the correct label, working to minimize the difference between the top two probabilities—a behavior further amplified by the regularized \CLIPTTA loss. In contrast, TENT prioritizes increasing the highest probability, thereby reinforcing the incorrect prediction.


\begin{figure}[!t]
\centering
\begin{tabularx}{\textwidth}{XXX}
 \resizebox{0.32\textwidth}{!}{\input{figures/Fig7_acc.pgf}} & \resizebox{0.3\textwidth}{!}{\input{figures/improvement2.pgf}} & \resizebox{0.3\textwidth}{!}{\input{figures/deterioration2.pgf}} \\
\centering (a) Accuracy & \centering (b) Improvement ratio & \centering (c) Deterioration ratio
\end{tabularx}
\vspace{0.5em}
\caption{\textbf{Improvement \& Deterioration ratios on CIFAR-10-C.} (a) While TENT’s accuracy collapses, \CLIPTTA shows consistent improvement. (b) The improvement ratio quantifies the proportion of misclassified examples correctly relabeled after adaptation. (c) The deterioration ratio captures the proportion of correctly classified examples that become misclassified post-adaptation.}
\label{fig:quantitative_insights}
\vspace{-1em}
\end{figure}
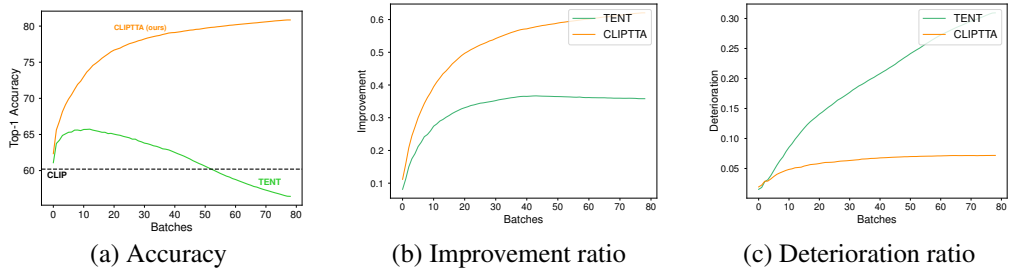

We provide quantitative insights on the CIFAR-10-C dataset in \cref{fig:quantitative_insights}. In \cref{fig:quantitative_insights}-a, we observe the collapse of TENT, while \CLIPTTA maintains robust performance. To further analyze this, we quantify two key metrics: the “improvement ratio” shown in \cref{fig:quantitative_insights}-b, which represents the proportion of misclassified examples that are correctly classified after adaptation, and the “deterioration ratio” shown in \cref{fig:quantitative_insights}-c, which denotes the proportion of correctly classified examples that become misclassified after adaptation. \CLIPTTA outperforms TENT by achieving a higher improvement ratio and a lower deterioration ratio.

\section{Experimental details} 
\label{supp_seq:exp_details}

We provide further information about the experimental setup that was conducted in the main paper. This includes the specifics of the experimental protocol, the baselines and benchmarks that were considered, as well as an extension of the empirical results.

\subsection{Detailed experimental protocol}
\label{supp_seq:exp_details_protocol}

In our experiments, we follow the widely explored \emph{non-episodic} TTA setting \cite{tent,ostta,unient}, in which the model is adapted continually to batches of data, without recovering its original weights. This poses a challenge, as adaptation risks of severely degrading the model, which can aggravate as adaptation goes longer. As some of the considered baselines were originally conceived for an \emph{episodic} setting (\emph{e.g.} CLIPArTT~\cite{clipartt}), some conditioning was applied in order to amplify their performance in this scenario. 

\subsection{Details on baselines and datasets}
\label{supp_seq:exp_details_baselines_datasets}

\paragraph{Benchmarks.} We provide more detailed information about the datasets that compose the different benchmarks used through the main paper. For all the experiments, images of different sizes were reshaped for compatibility with CLIP (\ie, to 224$\times$224).

\paragraph{Natural images.} We employed CIFAR-10 and CIFAR-100~\cite{cifar10}, both containing 10,000 images of size 28$\times$28, and spanning 10 and 100 classes, respectively. We use Imagenet~\cite{imagenet} as a larger-scale dataset, with 1000 classes and 50,000 images in total. 

\paragraph{Corruptions.} Transformed variants of the previous benchmarks are built by applying 15 different corruptions such as \emph{gaussian noise}, \emph{fog}, or \emph{pixelate}. This results in CIFAR-10-C and CIFAR-100-C \cite{cifar10c} and Imagenet-C. Each corruption is utilized in its highest severity level (\emph{e.g., level 5}), yielding the most complex version of each dataset. The number of images in each corrupted set and their size correspond to the previous benchmark, which results in 45 different datasets to evaluate in total.

\paragraph{Fine-grained classification datasets} are a popular choice in zero-shot classification with CLIP, as they span a wide semantic variety in their classes. We utilize Imagenet as well 10 other datasets covering:  Aircraft~\cite{aircraft}, Caltech101~\cite{caltech}, Cars~\cite{cars}, DTD~\cite{dtd}, EuroSat~\cite{eurosat}, Flowers102~\cite{flowers}, Food101~\cite{food}, Pets~\cite{pets}, SUN397~\cite{sun}, and UCF101~\cite{ucf}. The specific details of each dataset are condensed in Table~\ref{tab:fine_grained}.

\begin{table}[]
    \centering
    \begin{tabularx}{\linewidth}{lYlYlYlY}
    \toprule
    Dataset    & Classes & Size & Category \\
    \midrule
    Aircraft   & 100   & 3,333  & Transportation \\
    Caltech101 & 100   & 2,465  & Objects \\
    Cars       & 196   & 8,041  & Transportation \\
    DTD        & 47    & 1,692  & Textures \\
    EuroSat    & 10    & 8,100  & Satellite \\
    Flowers102 & 102   & 2,463  & Flora \\
    Food101    & 101   & 30,300 & Food  \\
    Pets       & 37    & 3,669  & Fauna \\
    SUN397     & 397   & 19,850 & Scenes \\
    UCF101     & 101   & 3,783  & Actions \\
    \bottomrule
    \end{tabularx}
    \caption{\textbf{Detailed information of the fine-grained classification benchmark.}}
    \label{tab:fine_grained}
\end{table}

\paragraph{Domain generalization.} This is a set of datasets popularly use in the context of Domain Adaptation. We use Visda-C~\cite{visda}, which includes 12 common classes and contains two main sets: a set of 152,397 3D renderings and a set of 55,388 of images cropped from MS COCO~\cite{mscoco}. We also incorporate PACS~\cite{pacs}, with seven classes, and OfficeHome~\cite{officehome} with 65 classes, which include images in four different styles, as summarized in Table~\ref{tab:dataset_summary}a). Finally, we include the challenging Imagenet-Domains benchmark, involving four variants of Imagenet: Imagenet-V2~\cite{imagenetv2}, Imagenet-R~\cite{imagenetr}, Imagenet-S~\cite{imagenets}, Imagenet-A~\cite{imageneta}, each of which is detailed in Table~\ref{tab:dataset_summary}b).

\paragraph{Out-of-distribution datasets.} In our open-set TTA setup, each ID dataset in the natural and corrupted image benchmarks is paired with a corresponding OOD dataset. The classification task is performed only on ID samples, while OOD samples are solely used for detection (\ie, recognizing and rejecting unknowns). Thus, OOD class labels are not meaningful in this context. Following prior work~\cite{unient,stamp}, we use SVHN~\cite{svhn} (26,032 street view digit images) as the OOD set for CIFAR-10 and CIFAR-100, and Places365~\cite{places} (1.8M scene images) for ImageNet. In the corrupted setting (\ie, CIFAR-10/100-C and ImageNet-C), we use SVHN-C and Places365-C as OOD sources, matched by corruption type (\eg, JPEG compression) and set to maximum severity.



\begin{table*}[t]
\centering
\begin{tabular}{c c}
    \begin{tabular}{l cc}
    \toprule
    Dataset & Domain  & Size \\
    \midrule    
    \multirow{4}{*}{PACS} & Photo        & 1,670 \\
                          & Cartoon      & 2,344 \\
                          & Sketch       & 3,929 \\
                          & Art painting & 2,048 \\
    \midrule
    \multirow{4}{*}{OfficeHome} & Art        & 965   \\
                                & Clipart    & 2,535 \\
                                & Product    & 2,470 \\
                                & Real world & 1,495 \\
    \bottomrule
    \end{tabular}
    &
    \begin{tabular}{lccc}
    \toprule
    Dataset     & Domain      & Classes & Size \\
    \midrule
    Imagenet    & Natural     & 1,000   & 50,000 \\
    Imagenet-V2 & Natural     & 1,000   & 10,000 \\
    Imagenet-S  & Sketch      & 1,000   & 50,000 \\
    Imagenet-R  & Art         & 200     & 30,000 \\
    Imagenet-A  & Adversarial & 7,500   & 7,500  \\
    \bottomrule \\
    \end{tabular}\\[1em]
(a) PACS and OfficeHome & (b) Imagenet-Domains
\end{tabular}
\caption{\textbf{Detailed dataset statistics.} (a) PACS and OfficeHome. (b) Imagenet-Domains.}
\label{tab:dataset_summary}
\end{table*}

\paragraph{Baselines.} We group baselines into three categories based on their adaptation strategy. The first group includes entropy-based methods for standard classifiers such as TENT~\cite{tent}, ETA~\cite{eta}, SAR~\cite{sar}, RoTTA~\cite{rotta}, OSTTA~\cite{ostta}, SoTTA~\cite{sotta}, STAMP~\cite{stamp}, and UniEnt~\cite{unient}. These methods typically operate by minimizing the conditional entropy of the model’s predictions and require adaptations to work with CLIP’s vision-language outputs. The second group comprises CLIP-specific methods such as CLIPArTT~\cite{clipartt} and WATT~\cite{watt}, which modify the loss or prompt structure to better leverage CLIP’s multimodal nature. The third group includes alternative CLIP-based adaptation approaches: TPT~\cite{tpt}, which performs prompt tuning via entropy minimization, and TDA~\cite{tda}, which operates without gradients using a memory-based episodic scheme. All baselines are implemented following their respective publications. For CLIP-based methods, minimal changes were needed to integrate into our framework. For non-CLIP methods, we use CLIP’s image-to-text similarities (as defined in Eq.\ref{eq:clip_prob}, Sec.\ref{sec:method}) as classification logits. Entropy-based baselines directly apply their loss to these logits. Hyperparameter details are provided below when applicable.

\begin{itemize}
    \item ETA: a similarity threshold of $\epsilon = 1$ and an entropy threshold $\alpha = 0.4$ are used. These are kept for all cases.
    \item SAR: an entropy threshold $\alpha = 0.4$ and an exponential moving average (EMA) weight $m = 0.2$ are used for all cases. The SAM optimizer is employed.
    \item RoTTA: we use a timeliness weight $\lambda_t = 1$ and an uncertainty weight $\lambda_u = 1$, a memory capacity equivalent to the batch size. These are kept for all cases.
    \item TDA: we use the same values used for Imagenet in the original paper. We employ $\alpha_{\text{pos}} = 2.0$, $\beta_{\text{pos}} = 2.0$, $\alpha_{\text{neg}} = 0.117$, $\beta_{\text{neg}} = 1.0$, entropy thresholds $H_o = \{0.2, 0.5\}$, entropy masks $M_o = \{0.03, 1.0\}$, and positive and negative shot capacities of 2 and 3, respectively.
    \item CLIPArTT: we take $K=3$ most probable classes in all datasets, except for $K = 5$ in VisDA-C, which uses a learning rate of $1\times10^{-5}$.
    \item WATT: we use two adaptation iterations per text prompt, and two meta-repetitions are used. A learning rate of $1\times10^{-5}$ is used for VisDA-C.
    \item SoTTA: we use the confidence threshold $\tau = 1/|\mathcal{C}|$, with $\mathcal{C}$ the number of classes. The memory capacity is equal to the batch size. The SAM optimizer is employed.
    \item UniEnt: we use $\lambda_\text{reg} = 1$ and $\lambda_\text{ood} = 1$. 
\end{itemize}

\subsection{Extended experimental results}
\label{supp_seq:exp_details_extended_results}

\begin{wraptable}{r}{0.55\textwidth}
    \caption{\textbf{Impact of updating the text encoder}.}
    \centering
    \begingroup
    \label{tab:visual_text}
    \small
    \begin{tabularx}{0.8\linewidth}{>{\small}l >{\small}Y >{\small}Y }
        \toprule
       \multirow{2}{*}{Dataset} & \CLIPTTA   & \CLIPTTA  \\
         & \scriptsize{(Vision only)} & \scriptsize{(Vision + Text)} \\
        \midrule
        CIFAR-10 & 95.0 &  93.5 ~~\scriptsize{(\textcolor{BrickRed}{-1.5})} \\
        CIFAR-100 & 74.9 & 75.0 ~~\scriptsize{(\textcolor{OliveGreen}{+0.1})}\\
        ImageNet & 69.1 & 69.6 ~~\scriptsize{(\textcolor{OliveGreen}{+0.5})}\\
        ImageNet-V2 & 62.7 & 63.1  ~~\scriptsize{(\textcolor{OliveGreen}{+0.4})} \\
        ImageNet-A & 54.0 & 54.2  ~~\scriptsize{(\textcolor{OliveGreen}{+0.2})}\\
        ImageNet-R & 80.1 & 79.9 ~~\scriptsize{(\textcolor{BrickRed}{-0.2})}\\
        ImageNet-S & 50.8 & 51.2 ~~\scriptsize{(\textcolor{OliveGreen}{+0.4})}\\
        Aircraft & 26.5& 26.9 ~~\scriptsize{(\textcolor{OliveGreen}{+0.4})}\\
        Caltech101 & 94.2 & 94.4 ~~\scriptsize{(\textcolor{OliveGreen}{+0.2})}\\
        Cars & 66.7 & 67.1 ~~\scriptsize{(\textcolor{OliveGreen}{+0.4})}\\ 
        DTD  & 46.5 & 48.1 ~~\scriptsize{(\textcolor{OliveGreen}{+1.6})}\\
        EuroSat  & 80.3& 72.9 ~~\scriptsize{(\textcolor{BrickRed}{-7.4})}\\
        Flowers102  & 71.3& 71.7 ~~\scriptsize{(\textcolor{OliveGreen}{+0.4})} \\
        Food101  & 86.7. & 86.8 ~~\scriptsize{(\textcolor{OliveGreen}{+0.1})} \\
        OxfordPets  & 91.6& 92.4 ~~\scriptsize{(\textcolor{OliveGreen}{+0.8})}\\
        SUN397  & 65.2 & 67.5 ~~\scriptsize{(\textcolor{OliveGreen}{+2.5})}\\
        UCF101  & 69.3 & 70.3 ~~\scriptsize{(\textcolor{OliveGreen}{+1.0})}\\
        \midrule
        Median & 69.2 & 70.3 ~~\scriptsize{(\textcolor{OliveGreen}{+1.1})} \\
        \bottomrule
    \end{tabularx}
    \endgroup
\end{wraptable}

\paragraph{Adapting the text encoder.} \CLIPTTA is evaluated across a diverse set of datasets by adapting not only the visual encoder but also the text encoder, as shown in Table~\ref{tab:visual_text}. Updating the text encoder proves beneficial in many cases, particularly for semantically complex datasets where CLIP’s pre-trained embeddings may lack sufficient specialization. This is evident in datasets focused on fine-grained classification, such as SUN397 and OxfordPets, where incorporating text encoder updates yields notable improvements. However, updating the text encoder can sometimes have detrimental effects, especially on datasets containing general or well-represented concepts, such as EuroSat. Despite being visually challenging, the broad and commonly encountered class labels in such datasets may already be adequately represented in CLIP’s original text embeddings. In these cases, further adaptation of the text encoder may disrupt this alignment, leading to performance degradation. This behavior underscores the importance of selectively adapting the text encoder based on the semantic complexity of the dataset.

Moreover, the results highlight the trade-off between generalization and specialization when jointly adapting both encoders. While semantically complex datasets benefit from increased specialization, datasets with simpler or well-represented class concepts risk losing the robust generalization capabilities inherent to CLIP’s pre-trained representations. This suggests that a targeted or dataset-specific strategy for adapting the text encoder may be more effective in leveraging its potential.

\paragraph{Open-set TTA on corrupted datasets.} Table~\ref{tab:top1_open_set_corrupted} reports results in the challenging open-set setting under corruption shifts. This scenario is challenging because models must adapt to noisy in-distribution samples while maintaining robustness to unseen OOD classes. As previously observed, TENT is highly unstable in these settings, suffering from severe model collapse that is exacerbated by corrupted inputs. Its accuracy drops to 2.1\% on ImageNet-C and 10.6\% on CIFAR-100-C, with poor OOD detection (FPR95 above 95\%), confirming its sensitivity to pseudo-label noise.

In contrast, \CLIPTTA with the OCE loss maintains high performance across all benchmarks, achieving the best overall results on both accuracy and OOD detection.  On average, on the corrupted datasets, it improves over UniEnt by +5.8 points in accuracy and reduces FPR95 by nearly 20 points. These gains demonstrate the benefit of aligning the adaptation objective with CLIP’s pre-training loss while integrating an explicit OOD detection signal. The results confirm that \CLIPTTA is well-suited for open-set test-time adaptation, even under strong distribution shifts such as corruptions.

\vspace{1.0em}
\begin{table*}[!ht]
\setlength\tabcolsep{0pt}
    \caption{\textbf{Open-set TTA results}. Top-1 accuracy with ViT-B/16 backbone on the open-set setting.}    \vspace{-0.5em}
    \label{tab:open_set_details}
    \centering
    \begingroup
   \small
    \begin{tabularx}{\linewidth}{>{\small}l  
    >{\small}Y >{\small}Y>{\small}Y
    c 
    >{\small}Y >{\small}Y>{\small}Y 
    c
    >{\small}Y >{\small}Y>{\small}Y 
    c
    >{\small}Y >{\small}Y>{\small}Y }
    \toprule
       & \multicolumn{3}{c}{CIFAR-10}  & &\multicolumn{3}{c}{CIFAR-100}    & &\multicolumn{3}{c}{ImageNet} & & \multicolumn{3}{c}{Average}\\
       \cmidrule(l{4pt}r{4pt}){2-4}
       \cmidrule(l{4pt}r{4pt}){6-8}
       \cmidrule(l{4pt}r{4pt}){10-12}
       \cmidrule(l{4pt}r{4pt}){14-16}
      & {\scriptsize ACC$\uparrow$}  & {\scriptsize AUC$\uparrow$} & {\scriptsize FPR95$\downarrow$} & & {\scriptsize ACC$\uparrow$} & {\scriptsize AUC$\uparrow$} & {\scriptsize FPR95$\downarrow$}  & & {\scriptsize ACC$\uparrow$} & {\scriptsize AUC$\uparrow$} & {\scriptsize FPR95$\downarrow$~}&  & {\scriptsize ACC$\uparrow$} & {\scriptsize AUC$\uparrow$} & {\scriptsize FPR95$\downarrow$} \\
    \midrule
    CLIP        & 89.3 & 98.5 & 5.2  & ~~~ & 68.1 & 86.8 & 83.5 & ~~~&  66.7 & 90.1 & 43.8 & ~~~ &  74.7 & 91.8 & 44.2\\
    TENT \cite{tent} & 93.0 & 42.3 & 89.3 & &69.1  & 36.2 & 94.8 &  ~~~&  12.4 & 49.9  & 89.4  & ~~~ &   58.2 & 42.8 & 91.2  \\
    \midrule
    OSTTA \cite{ostta}  & 90.9 & 60.5 & 72.8 & &70.9 & 43.3 & 93.8 &  ~~~&  66.9 & 84.9 & 59.2 & ~~~ &  76.2	&62.9 &	75.3\\
    SoTTA \cite{sotta}  & 89.5 & 98.5 & 4.9 & & 68.9 & 88.5 & 76.3 & ~~~& 66.7 & 89.3 & 47.1 & ~~~ &  75.0	& 92.1 & 42.8 \\
    STAMP \cite{stamp}   & 89.9 & 98.6 & 5.5 & & 67.5 & 87.7 & 80.0 & & 29.7& 63.0 & 80.2 & &62.4	&83.1 & 55.2 \\
    UniEnt \cite{unient}  & \underline{94.2} & \textbf{99.9}& \textbf{0.0} & & \underline{72.7} & \underline{97.8} & \underline{8.7} &  ~~~&   65.2		&  95.4		&  17.1 & ~~~ & \underline{77.3}	& \underline{97.7} & \underline{8.6} \\
    \midrule
    \rowcolor{cliptta_color} \CLIPTTA + $\text{\scriptsize{OCE}}$ \scriptsize{(ours)}  ~~ &  \textbf{94.6} & \underline{99.8} &  \underline{0.4}&  &\textbf{74.9} & \textbf{98.4} & \textbf{7.6} &   ~~~&  \textbf{67.6} & \textbf{97.7} & \textbf{9.7} & ~~~ &  \textbf{79.0} & \textbf{98.6} & \textbf{5.9}\\
    \bottomrule
    \end{tabularx}
    \endgroup
\end{table*}

\vspace{1.5em}
\begin{table*}[ht]
\setlength\tabcolsep{0pt}
    \caption{\textbf{Open-set TTA results on Corrupted Datasets}. Top-1 accuracy with ViT-B/16 backbone on the open-set setting.}    \vspace{-0.5em}
    \centering
    \begingroup
   \small
    \label{tab:top1_open_set_corrupted}
    \begin{tabularx}{\linewidth}{>{\small}l  
    >{\small}Y >{\small}Y>{\small}Y
    c 
    >{\small}Y >{\small}Y>{\small}Y 
    c
    >{\small}Y >{\small}Y>{\small}Y 
    c
    >{\small}Y >{\small}Y>{\small}Y }
    \toprule
       & \multicolumn{3}{c}{CIFAR-10-C}  & &\multicolumn{3}{c}{CIFAR-100-C}    & &\multicolumn{3}{c}{Imagenet-C} & & \multicolumn{3}{c}{Average}\\
       \cmidrule(l{4pt}r{4pt}){2-4}
       \cmidrule(l{4pt}r{4pt}){6-8}
       \cmidrule(l{4pt}r{4pt}){10-12}
       \cmidrule(l{4pt}r{4pt}){14-16}
      & {\scriptsize ACC$\uparrow$}  & {\scriptsize AUC$\uparrow$} & {\scriptsize FPR95$\downarrow$} & & {\scriptsize ACC$\uparrow$} & {\scriptsize AUC$\uparrow$} & {\scriptsize FPR95$\downarrow$}  & & {\scriptsize ACC$\uparrow$} & {\scriptsize AUC$\uparrow$} & {\scriptsize FPR95$\downarrow$~}&  & {\scriptsize ACC$\uparrow$} & {\scriptsize AUC$\uparrow$} & {\scriptsize FPR95$\downarrow$} \\
    \midrule
    CLIP        &  60.2 & 88.0 & 58.0 & & 35.2 & 67.0 & 93.8 & & 24.6 & 68.6 & 89.2 & & 40.0 & 74.5 & 80.3  \\
    TENT  \cite{tent} &  26.9 & 50.7 & 91.2 & & 10.6 & 43.1 & 95.0 & & 2.1 & 48.0 & 95.0 & & 13.2 & 47.3 & 93.7 \\
    \midrule
    OSTTA  \cite{ostta}  &  62.7 & 53.4 & 85.2 & & 34.5 & 36.3 & 92.6 & & \underline{31.2} & \underline{75.1} & \underline{79.8} & & 42.8 & 54.9 & 85.9\\
    STAMP   \cite{stamp}  &  60.4 & 88.0 & 57.1 & &34.5 & 66.8 & 93.8& & 9.0 & 53.8 & 92.9& & 34.6 & 69.5 & 81.3\\
    UniEnt  \cite{unient}  &  \underline{78.7} & \underline{98.6} & \textbf{5.7} & & \underline{48.9} & \underline{91.0} & \underline{31.0} & & 23.6 & 44.8 & 90.8 & &\underline{50.4} & \underline{78.1} & \underline{42.5}\\
    \midrule
    \rowcolor{cliptta_color} \CLIPTTA + $\text{\scriptsize{OCE}}$ \scriptsize{(ours)}~~&  \textbf{79.1} & \textbf{98.7} & \underline{6.1} & & \textbf{50.4} & \textbf{96.7} & \textbf{19.2}& & \textbf{39.0} & \textbf{89.0} & \textbf{43.2} & & \textbf{56.2} & \textbf{94.8} & \textbf{22.8}\\
    \bottomrule
    \end{tabularx}
    \endgroup
\end{table*}

\paragraph{Domain shifts benchmarks.} We provide the extended results of the different domain shifts (Table~\ref{tab:top1_corruptions}), including Imagenet-Domains (Table~\ref{tab:imagenet_domains}), VisDA-C (Table~\ref{tab:visda}), OfficeHome (Table~\ref{tab:office}), and PACS (Table~\ref{tab:pacs_supp}). \CLIPTTA achieves the best performance across all of these datasets on average, demonstrating great flexibility across domains. Our method also obtains highly competitive results independently in each sub-dataset. 

\vspace{1.0em}
\begin{table*}[ht]
    \setlength\tabcolsep{0.4pt}
    \caption{\textbf{Detailed results on the Imagenet-Domains benchmark.}}
    \vspace{-0.5em}
    \label{tab:imagenet_domains}
    \centering
    \begingroup
   \small
    \begin{tabularx}{\linewidth}{>{\small}l >{\small}Y >{\small}Y >{\small}Y >{\small}Y >{\small}Y >{\small}Y} 
        \toprule
         & \small{ImageNet} & \small{ImageNet-A} & \small{ImageNet-V2} & \small{ImageNet-R} & \small{ImageNet-S} & \small{Average} \\ 
         \midrule
         CLIP & 66.7 & 47.8 & 60.8 & 74.0 & 47.8 & 59.4 \\
         TPT \scriptsize{(NeurIPS '22)}& 69.0 & \underline{54.8} & 63.5 & 77.1 & 47.9 & 62.4 \\
         TDA \scriptsize{(CVPR '24)} & \underline{69.5} & \textbf{60.1} & \textbf{64.7} & \textbf{80.2} & \underline{50.5} & \textbf{65.0} \\
         \midrule
         TENT  \scriptsize{(ICLR '21)} & 66.5 & 51.3 & 60.2 & 79.4 & 43.7 & 60.2 \\
         ETA \scriptsize{(ICML '22)} & 67.4 & 49.2 & 60.9 & 75.3 & 46.8 & 59.9 \\
         SAR \scriptsize{(ICLR '23)}& 66.7 & 51.5 & 60.5 & 79.6 & 44.6 & 60.6 \\
         RoTTA  \scriptsize{(CVPR '23)}& 68.4 & 51.2 & 62.5 & 78.1 & 47.8 & 61.6 \\
         CLIPArTT \scriptsize{(WACV '25)} & 67.6 & 50.7 &  61.2 & 76.2 & 47.9 & 60.7  \\
         WATT \scriptsize{(NeurIPS '24)} & 69.0 & 51.1 & 62.5 & 78.1 & 48.2 & 61.8  \\
         \midrule
             \rowcolor{cliptta_color}\CLIPTTA (ours) & \textbf{69.6} & 54.0 &  62.7 & \textbf{80.2} & \textbf{50.8} & \underline{63.4} \\
         
         \bottomrule
        \end{tabularx}
    \endgroup
\end{table*}

\vspace{1.0em}
\begin{table*}[!ht]
    \centering
    \begin{tabularx}{\linewidth}{>{\small}l >{\small}Y >{\small}Y >{\small}Y}
    \toprule
    & \small{Synthetic 3D} & \small{MS COCO} & \small{Average} \\
    \midrule
    CLIP & 87.2 & 86.7 & 87.0 \\
    TPT \cite{tpt} & 85.5 & 84.5 & 85.0 \\
    TDA \cite{tda} & 86.6 & 86.5 & 86.5  \\
    \midrule
    TENT \cite{tent} & \textbf{93.2} & 85.3 & \underline{89.3}\\
    ETA \cite{eta} & 91.1 & 85.4 & 88.3  \\
    SAR \cite{sar} & 88.1 & \textbf{87.5} & 87.8  \\
    RoTTA \cite{rotta} & 80.6 & 86.7 & 83.7  \\
    CLIPArTT \cite{clipartt} & 82.2 & 86.0 & 84.1  \\
    WATT \cite{watt} & 88.4 & \underline{87.0} & 87.7 \\
    \midrule 
    \rowcolor{cliptta_color} \CLIPTTA \scriptsize{(ours)}& \underline{92.2} & 86.9 & \textbf{89.6} \\
    \bottomrule
    \end{tabularx}
    \caption{\textbf{Detailed results on the two domains of the Visda-C dataset.}}
    \label{tab:visda}
\end{table*}

\begin{table*}[!ht]
    \centering
    \begin{tabularx}{\linewidth}{>{\small}l >{\small}Y >{\small}Y >{\small}Y >{\small}Y >{\small}Y}
    \toprule
    & \small{Art} & \small{Clipart} & \small{Product} & \small{Real} & \small{Average} \\
    \midrule
    CLIP & 83.2 & 68.0 & 89.1 & 89.8 & 82.5 \\
    TPT \cite{tpt} & 82.5 & 66.3 & 88.5 & 89.2 & 81.7 \\
    TDA \cite{tda} & 83.2 & 68.8 & 89.8 & 90.4 & 83.0 \\
    \midrule
    TENT \cite{tent} & 84.1 & 68.8 & 90.0 & 90.5 & 83.4 \\
    ETA \cite{eta} & 84.3 & 70.8 & \underline{90.4} & \underline{90.7} & \underline{84.1} \\
    SAR \cite{sar} & \textbf{84.4} & \textbf{70.9} & 89.6 & 90.3 & 83.8 \\
    RoTTA \cite{rotta} & 82.9 & 68.0 & 89.1 & 89.8 & 82.5 \\
    CLIPArTT \cite{clipartt} & 82.6 & 68.4 & 87.6 & 89.6 & 82.0 \\
    WATT \cite{watt} & 83.8 & 69.0 & 90.0 & 90.5 & 83.4 \\
    \midrule 
    \rowcolor{cliptta_color}\CLIPTTA \scriptsize{(ours)} & \underline{84.2} & \underline{70.7} & \textbf{91.0} & \textbf{91.0} & \textbf{84.2} \\
    \bottomrule
    \end{tabularx}
    \caption{\textbf{Detailed results on the four domains of the OfficeHome (OH) dataset.}}
    \label{tab:office}
\end{table*}

\begin{table*}[!ht]
    \centering
    \begin{tabularx}{\linewidth}{>{\small}l >{\small}Y>{\small}Y>{\small}Y>{\small}Y>{\small}Y}
    \toprule
    & \small{Photo} & \small{Art} & \small{Cartoon} & \small{Sketch} & \small{Average} \\
    \midrule
    CLIP & 99.9 & 97.4 & 99.1 & 88.1 & 96.1 \\
    TPT \cite{tpt} & 99.5 & 95.3 & 93.9 & 87.2 & 94.0 \\
    TDA \cite{tda} & \textbf{99.9} & 97.5 & 98.9 & 88.1 & 96.1 \\
    \midrule
    TENT \cite{tent} & 99.8 & 98.0 & \underline{99.2} & 89.1 & 96.6 \\
    ETA \cite{eta} & 99.8 & 97.9 & \textbf{99.3} & 89.8 & \underline{96.7} \\
    SAR \cite{sar} & \textbf{99.9} & 97.5 & 99.1 & 88.2 & 96.2 \\
    RoTTA \cite{rotta} & \textbf{99.9} & 93.8 & 98.8 & 88.1 & 95.8 \\
    CLIPArTT \cite{clipartt} & 99.5 & 96.9 & 98.3 & 90.4 & 96.2 \\
    WATT \cite{watt} &\textbf{99.9} & \underline{97.6} & \underline{99.2} & 88.4 & 96.2 \\
    \midrule 
    \rowcolor{cliptta_color}\CLIPTTA \scriptsize{(ours)} & \textbf{99.9} & \textbf{98.0} & \textbf{99.3} & \textbf{92.0} & \textbf{97.5} \\
    \bottomrule
    \end{tabularx}
    \caption{\textbf{Detailed results on the four domains of the PACS dataset.}}
    \label{tab:pacs_supp}
\end{table*}

\begin{table*}[!ht]
\setlength\tabcolsep{0pt}
    \caption{\textbf{Closed-set TTA on coarse-grained datasets}. Top-1 accuracy with ViT-B/16 backbone on coarse-grained datasets (CIFAR-10 and CIFAR-100).}
    \centering
    \label{tab:top1_coarse}
    \begin{tabularx}{\linewidth}{>{\small}l >{\small}Y >{\small}Y>{\small}Y}
    \toprule
    & {CIFAR-10}  & {CIFAR-100}  & {Average} \\    
    \midrule
    CLIP                              &89.3&68.1&78.7 \\
    TPT \cite{tpt}    &89.8&67.4&78.6  \\
    TDA \cite{tda}       &91.4&69.8&80.6  \\
    \midrule
    TENT \cite{tent}      &\underline{94.9}&72.9&83.9 \\
    ETA  \cite{eta}      &94.8&\underline{73.7}&\underline{84.3} \\
    SAR  \cite{sar}      &92.1&73.2&82.7 \\
    RoTTA \cite{rotta}     &89.4&68.5&79.0 \\
    CLIPArTT \cite{clipartt}  &88.4&73.2&80.8   \\
    WATT \cite{watt}   &{92.5}&70.8&81.7 \\
    \midrule
    \rowcolor{cliptta_color} \CLIPTTA \scriptsize{(ours)} &\textbf{95.1}&\textbf{75.3}&\textbf{85.2}\\
    \bottomrule
    \end{tabularx}
\end{table*}

\begin{table*}[!ht]
    \setlength\tabcolsep{1pt}
    \caption{\textbf{Closed-set TTA on fine-grained datasets}. Top-1 accuracy comparison of \CLIPTTA against other TTA methods on a suite of 11 fine-grained datasets.}
    \label{tab:top1_datasets_suite}
    \centering
    \begingroup
    \begin{tabularx}{\linewidth}{>{\small}l >{\small}Y>{\small}Y>{\small}Y>{\small}Y>{\small}Y>{\small}Y>{\small}Y>{\small}Y>{\small}Y>{\small}Y>{\small}Y>{\small}Y>{\small}Y}
        \toprule
          & \rotatebox[origin=c]{70}{\scriptsize{ImageNet}} & \rotatebox[origin=c]{70}{\scriptsize{Aircraft}} & \rotatebox[origin=c]{70}{\scriptsize{Caltech101}} & \rotatebox[origin=c]{70}{\scriptsize{Cars}} & \rotatebox[origin=c]{70}{\scriptsize{DTD}} & \rotatebox[origin=c]{70}{\scriptsize{EuroSAT}} & \rotatebox[origin=c]{70}{\scriptsize{Flowers102}} & \rotatebox[origin=c]{70}{\scriptsize{Food101}} & \rotatebox[origin=c]{70}{\scriptsize{OxfordPets}} & \rotatebox[origin=c]{70}{\scriptsize{SUN397}} & \rotatebox[origin=c]{70}{\scriptsize{UCF101}} & \rotatebox[origin=c]{70}{\scriptsize{Average}} \\
         \midrule
         CLIP~ & 66.7 & 24.8 & 92.2 & 65.5 & 44.1 & 48.3 & 70.7 & 84.8 & 88.4 & 62.3 & 64.7 & 64.7\\ 
         TPT \cite{tpt} & {69.0} & \underline{24.8} & \textbf{94.2} & \underline{66.9} & \textbf{47.8} & 42.4 & 69.0 & 84.7 & 87.8 & 65.5 & 68.0 & 65.6\\ 
         TDA \cite{tda} & \underline{69.5} & 23.9 & \textbf{94.2} & \textbf{67.3} & \underline{47.4} & \underline{58.0} & \textbf{71.4} & 86.1 & {88.6} & \textbf{67.6} & \textbf{70.7} & \underline{67.7}\\
         \midrule
         TENT \cite{tent} & 66.5 & 15.5 & 93.8 & 63.0 & 43.1 & 58.4 & \underline{71.3} & 86.5 & 89.5 & 63.1 & 68.0& 65.3 \\
         ETA \cite{eta} & 67.4 & 24.8 & 93.0 & 65.2 & 44.4 & 47.5 & \textbf{71.4} & 85.9 & 89.2 & 63.6 & 66.6 & 65.4 \\
         SAR  \cite{sar} & 66.7  & 21.9 & 93.9 & 64.0 & 43.9 & 50.2 & 70.9 & \underline{86.5}& 89.6& 63.3& 67.7 & 65.3  \\
         RoTTA \cite{rotta} &  68.4&   22.3 &  94.0 &  58.1 & 45.2 &  24.2 & 70.5 & 81.6 &   87.0 & 64.9 & 66.8 &  62.1 \\
         CLIPArTT \cite{clipartt}  &67.5 & 24.0 & 92.7 & 64.0 & 43.4 & 46.7 & 67.0 & 84.2 & 87.1 & 64.2 & 67.0 & 64.4\\ 
         WATT \cite{watt} & 69.0 & 23.6 & \underline{94.1} & 65.8 & 44.7& 40.0 & \textbf{71.4} & \underline{86.2} & 88.7 & \underline{66.3} & 68.2& 65.3\\ 
         \midrule
         \rowcolor{cliptta_color}\CLIPTTA \scriptsize{(ours)} & \textbf{69.6} & \textbf{26.5} & \textbf{94.2} &  66.7 & 46.5  &   \textbf{80.3} &   \underline{71.3} &  \textbf{86.7}&  \textbf{91.6} & 65.2 &   \underline{69.3} &   \textbf{69.8} \\
         \bottomrule
    \end{tabularx}
    \endgroup
\end{table*}

\paragraph{Semantic datasets.} We report closed-set adaptation results on both coarse- and fine-grained classification tasks in Tables~\ref{tab:top1_coarse} and~\ref{tab:top1_datasets_suite}. On coarse-grained benchmarks (CIFAR-10 and CIFAR-100), \CLIPTTA achieves the highest accuracy on both datasets, with a strong average of 85.2\%, outperforming all TENT-based and CLIP-based methods, including CLIPArTT and WATT. Notably, it improves over TENT by +0.2 points on CIFAR-10 and +2.4 points on CIFAR-100, and remains significantly ahead of zero-shot CLIP (+6.5 points on average). On fine-grained datasets, \CLIPTTA consistently ranks among the top methods, achieving the best average accuracy across the 11 datasets (69.8\%). Despite its simplicity, it performs favorably compared to more complex CLIP-specific methods such as TPT, TDA, and CLIPArTT, which rely on prompt tuning or heuristic loss modifications. \CLIPTTA performs particularly well on datasets like EuroSAT (+22.3 over CLIPArTT) and OxfordPets (+4.5 over TDA) while maintaining competitive results on the others. These findings highlight the robustness of our adaptation objective across both coarse- and fine-grained tasks without requiring task-specific tuning or architectural modifications.

\begin{wraptable}{r}{0.6\textwidth}
    \centering
    \begin{tabularx}{\linewidth}{>{\small}l >{\small}Y >{\small}Y>{\small}Y}
    \toprule
    & {CIFAR-10}  & {CIFAR-100}  & {Imagenet}  \\    
    \midrule
    \CLIPTTA          & $94.9\pm 0.03$ & $75.3\pm 0.07$& $69.1\pm0.01$ \\
    \bottomrule
    \end{tabularx}
    \caption{Accuracy of \CLIPTTA averaged over three random initializations (mean $\pm$ 95\% CI).}
    \label{tab:statistics}
\end{wraptable}

\paragraph{Statistical significance.} We conduct additional runs of \CLIPTTA to assess its sensitivity to random initialization, reporting the mean accuracy and 95\% confidence interval in \cref{tab:statistics}. The results indicate that \CLIPTTA is highly stable, with very low variance across independent runs. The tight confidence intervals (e.g., $\pm 0.01$ on ImageNet) confirm the reliability and reproducibility of the observed performance gains, further supporting the robustness of the method across different datasets.

\newpage
\subsection{Hyperparameter analysis.}
\label{sec:supp_hp}

In this section, we evaluate the sensitivity of \CLIPTTA to its key hyperparameters: the regularization weight $\lambda_{\text{reg}}$, the OOD loss weight $\lambda_{\text{oce}}$, and the adaptation batch size. Results show that \CLIPTTA remains robust across a wide range of values, requiring minimal tuning for strong performance.

\begin{figure}[!t]
    \centering
    \includegraphics[width=0.7\linewidth]{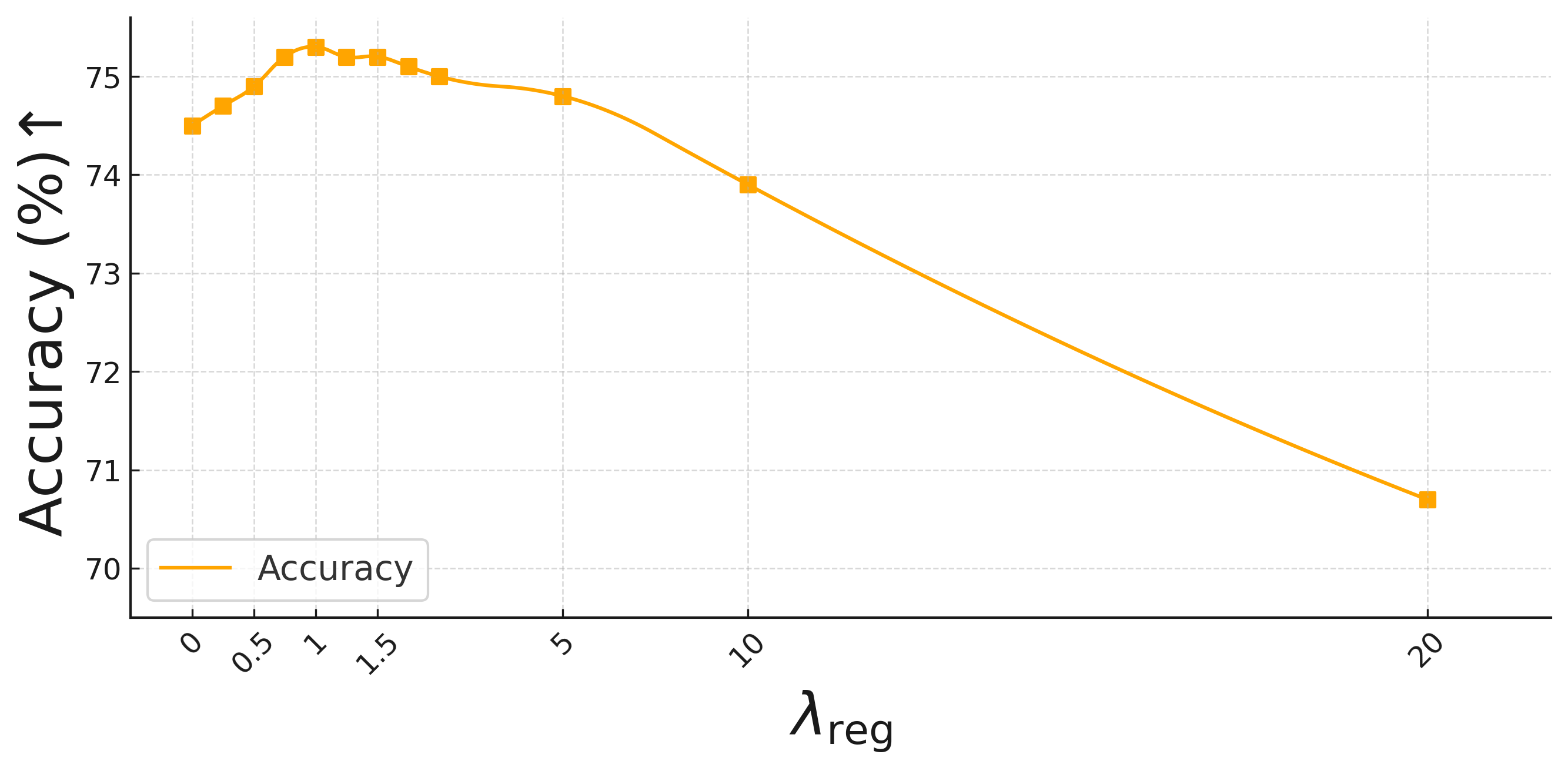}
    \caption{\textbf{Impact of $\lambda_{reg}$ on CIFAR-100.} Effect of $\lambda_{reg}$ on the closed-set accuracy of \CLIPTTA when evaluated on CIFAR-100.}
    \label{fig:hyper_param}
\end{figure}

\paragraph{Effect of $\lambda_{\text{reg}}$.}
Figure~\ref{fig:hyper_param} shows the impact of the regularization weight $\lambda_{\text{reg}}$ on CIFAR-100 accuracy. We observe that \CLIPTTA is remarkably stable for values ranging from 0.5 to 2.0, with accuracy consistently above 75\% in this range. Performance peaks around $\lambda_{\text{reg}} = 1$, which we use as the default. Beyond that, accuracy gradually declines, indicating that overly strong regularization may suppress beneficial updates. Overall, this confirms that \CLIPTTA does not require precise tuning of $\lambda_{\text{reg}}$ to perform well and that a wide range of values yields near-optimal performance.

\begin{wraptable}{r}{0.6\textwidth}
    \setlength\tabcolsep{1pt}
    \caption{\textbf{Impact of $\lambda_\text{oce}$ on Imagenet}. Effect of $\lambda_\text{oce}$ on accuracy and open-set metrics AUROC (AUC) and false positive rate (FPR).}
    \label{tab:lambda_oce}
    \centering
    \begingroup
    \begin{tabularx}{\linewidth}{>{\small}l >{\small}Y>{\small}Y>{\small}Y>{\small}Y>{\small}Y>{\small}Y>{\small}Y>{\small}Y>{\small}Y}
        \toprule
          $\lambda_\text{oce}$ & 0 & 0.25 & 0.5 & 1 & 2 & 5 & 10 & 20 & 100\\
          \midrule
          \textbf{Acc} & 67.6 & 67.6 & 67.6 & 67.6 & 67.5 & 67.3 & 66.4 & 64.5 & 56.6 \\
          \textbf{AUC} & 93.5 & 97.5 & 97.6 & 97.7 & 97.8 & 98.0 & 98.4 & 98.8 & 99.2 \\
          \textbf{FPR} & 25.7 & 10.1 & 9.8  & 9.7  & 8.8  & 7.8  & 6.3  & 4.7  & 2.3  \\
         \bottomrule
    \end{tabularx}
    \endgroup
\end{wraptable}

\paragraph{Effect of $\lambda_{\text{oce}}$.}
Table~\ref{tab:lambda_oce} reports the impact of $\lambda_{\text{oce}}$ on ImageNet in the open-set setting. While accuracy stays stable for small values, OOD detection improves substantially: AUROC increases from 93.5\% (no OCE) to 97.7\% at $\lambda_{\text{oce}} = 1$, and FPR95 drops by 16 points. Performance remains robust in the range [0.25–2], confirming the stability of the OCE loss.


\begin{wraptable}{r}{0.6\textwidth}
    \setlength\tabcolsep{1pt}
    \caption{\textbf{Accuracy on different batch sizes on the CIFAR-10 dataset}. Although \CLIPTTA benefits from larger batches, it remains competitive even in the extreme case of 1 image adaptation.}
    \vspace{1em}
    \label{tab:batch_size}
    \centering
    \begingroup
    \begin{tabularx}{\linewidth}{>{\small}l >{\small}Y>{\small}Y>{\small}Y>{\small}Y>{\small}Y>{\small}Y>{\small}Y>{\small}Y}
        \toprule
          \textbf{Batch size} & 1 & 2 & 8 & 32 & 64 & 128 & 256 & 512 \\
          \midrule
          \textbf{Accuracy} & 93.4 & 94.7 & 94.7 & 94.8 & 95.0 & 95.1 & 95.1 & 95.2 \\
         \bottomrule
    \end{tabularx}
    \endgroup
\end{wraptable}

\paragraph{Effect of batch size.}
Table~\ref{tab:batch_size} presents accuracy on CIFAR-10 for batch sizes ranging from 1 to 512. Accuracy increases with batch size and saturates around 64 samples, showing that \CLIPTTA benefits from richer batch-level statistics. Remarkably, even in the extreme case of a single image per batch, \CLIPTTA remains competitive (93.4\% accuracy). This is made possible by the confident memory, which stores reliable past predictions and enables the use of our soft contrastive loss even when no other images are available in the current batch. As a result, \CLIPTTA is well-suited for deployment in streaming where batch sizes may be small.

%% file: figures/Fig7_acc.pgf
\begingroup%
\makeatletter%
\begin{pgfpicture}%
\pgfpathrectangle{\pgfpointorigin}{\pgfqpoint{6.400000in}{4.800000in}}%
\pgfusepath{use as bounding box, clip}%
\begin{pgfscope}%
\pgfsetbuttcap%
\pgfsetmiterjoin%
\definecolor{currentfill}{rgb}{1.000000,1.000000,1.000000}%
\pgfsetfillcolor{currentfill}%
\pgfsetlinewidth{0.000000pt}%
\definecolor{currentstroke}{rgb}{1.000000,1.000000,1.000000}%
\pgfsetstrokecolor{currentstroke}%
\pgfsetdash{}{0pt}%
\pgfpathmoveto{\pgfqpoint{0.000000in}{0.000000in}}%
\pgfpathlineto{\pgfqpoint{6.400000in}{0.000000in}}%
\pgfpathlineto{\pgfqpoint{6.400000in}{4.800000in}}%
\pgfpathlineto{\pgfqpoint{0.000000in}{4.800000in}}%
\pgfpathlineto{\pgfqpoint{0.000000in}{0.000000in}}%
\pgfpathclose%
\pgfusepath{fill}%
\end{pgfscope}%
\begin{pgfscope}%
\pgfsetbuttcap%
\pgfsetmiterjoin%
\definecolor{currentfill}{rgb}{1.000000,1.000000,1.000000}%
\pgfsetfillcolor{currentfill}%
\pgfsetlinewidth{0.000000pt}%
\definecolor{currentstroke}{rgb}{0.000000,0.000000,0.000000}%
\pgfsetstrokecolor{currentstroke}%
\pgfsetstrokeopacity{0.000000}%
\pgfsetdash{}{0pt}%
\pgfpathmoveto{\pgfqpoint{0.800000in}{0.528000in}}%
\pgfpathlineto{\pgfqpoint{5.760000in}{0.528000in}}%
\pgfpathlineto{\pgfqpoint{5.760000in}{4.224000in}}%
\pgfpathlineto{\pgfqpoint{0.800000in}{4.224000in}}%
\pgfpathlineto{\pgfqpoint{0.800000in}{0.528000in}}%
\pgfpathclose%
\pgfusepath{fill}%
\end{pgfscope}%
\begin{pgfscope}%
\pgfsetbuttcap%
\pgfsetroundjoin%
\definecolor{currentfill}{rgb}{0.000000,0.000000,0.000000}%
\pgfsetfillcolor{currentfill}%
\pgfsetlinewidth{0.803000pt}%
\definecolor{currentstroke}{rgb}{0.000000,0.000000,0.000000}%
\pgfsetstrokecolor{currentstroke}%
\pgfsetdash{}{0pt}%
\pgfsys@defobject{currentmarker}{\pgfqpoint{0.000000in}{-0.048611in}}{\pgfqpoint{0.000000in}{0.000000in}}{%
\pgfpathmoveto{\pgfqpoint{0.000000in}{0.000000in}}%
\pgfpathlineto{\pgfqpoint{0.000000in}{-0.048611in}}%
\pgfusepath{stroke,fill}%
}%
\begin{pgfscope}%
\pgfsys@transformshift{1.025455in}{0.528000in}%
\pgfsys@useobject{currentmarker}{}%
\end{pgfscope}%
\end{pgfscope}%
\begin{pgfscope}%
\definecolor{textcolor}{rgb}{0.000000,0.000000,0.000000}%
\pgfsetstrokecolor{textcolor}%
\pgfsetfillcolor{textcolor}%
\pgftext[x=1.025455in,y=0.430778in,,top]{\color{textcolor}\sffamily\fontsize{14.000000}{16.800000}\selectfont 0}%
\end{pgfscope}%
\begin{pgfscope}%
\pgfsetbuttcap%
\pgfsetroundjoin%
\definecolor{currentfill}{rgb}{0.000000,0.000000,0.000000}%
\pgfsetfillcolor{currentfill}%
\pgfsetlinewidth{0.803000pt}%
\definecolor{currentstroke}{rgb}{0.000000,0.000000,0.000000}%
\pgfsetstrokecolor{currentstroke}%
\pgfsetdash{}{0pt}%
\pgfsys@defobject{currentmarker}{\pgfqpoint{0.000000in}{-0.048611in}}{\pgfqpoint{0.000000in}{0.000000in}}{%
\pgfpathmoveto{\pgfqpoint{0.000000in}{0.000000in}}%
\pgfpathlineto{\pgfqpoint{0.000000in}{-0.048611in}}%
\pgfusepath{stroke,fill}%
}%
\begin{pgfscope}%
\pgfsys@transformshift{1.603543in}{0.528000in}%
\pgfsys@useobject{currentmarker}{}%
\end{pgfscope}%
\end{pgfscope}%
\begin{pgfscope}%
\definecolor{textcolor}{rgb}{0.000000,0.000000,0.000000}%
\pgfsetstrokecolor{textcolor}%
\pgfsetfillcolor{textcolor}%
\pgftext[x=1.603543in,y=0.430778in,,top]{\color{textcolor}\sffamily\fontsize{14.000000}{16.800000}\selectfont 10}%
\end{pgfscope}%
\begin{pgfscope}%
\pgfsetbuttcap%
\pgfsetroundjoin%
\definecolor{currentfill}{rgb}{0.000000,0.000000,0.000000}%
\pgfsetfillcolor{currentfill}%
\pgfsetlinewidth{0.803000pt}%
\definecolor{currentstroke}{rgb}{0.000000,0.000000,0.000000}%
\pgfsetstrokecolor{currentstroke}%
\pgfsetdash{}{0pt}%
\pgfsys@defobject{currentmarker}{\pgfqpoint{0.000000in}{-0.048611in}}{\pgfqpoint{0.000000in}{0.000000in}}{%
\pgfpathmoveto{\pgfqpoint{0.000000in}{0.000000in}}%
\pgfpathlineto{\pgfqpoint{0.000000in}{-0.048611in}}%
\pgfusepath{stroke,fill}%
}%
\begin{pgfscope}%
\pgfsys@transformshift{2.181632in}{0.528000in}%
\pgfsys@useobject{currentmarker}{}%
\end{pgfscope}%
\end{pgfscope}%
\begin{pgfscope}%
\definecolor{textcolor}{rgb}{0.000000,0.000000,0.000000}%
\pgfsetstrokecolor{textcolor}%
\pgfsetfillcolor{textcolor}%
\pgftext[x=2.181632in,y=0.430778in,,top]{\color{textcolor}\sffamily\fontsize{14.000000}{16.800000}\selectfont 20}%
\end{pgfscope}%
\begin{pgfscope}%
\pgfsetbuttcap%
\pgfsetroundjoin%
\definecolor{currentfill}{rgb}{0.000000,0.000000,0.000000}%
\pgfsetfillcolor{currentfill}%
\pgfsetlinewidth{0.803000pt}%
\definecolor{currentstroke}{rgb}{0.000000,0.000000,0.000000}%
\pgfsetstrokecolor{currentstroke}%
\pgfsetdash{}{0pt}%
\pgfsys@defobject{currentmarker}{\pgfqpoint{0.000000in}{-0.048611in}}{\pgfqpoint{0.000000in}{0.000000in}}{%
\pgfpathmoveto{\pgfqpoint{0.000000in}{0.000000in}}%
\pgfpathlineto{\pgfqpoint{0.000000in}{-0.048611in}}%
\pgfusepath{stroke,fill}%
}%
\begin{pgfscope}%
\pgfsys@transformshift{2.759720in}{0.528000in}%
\pgfsys@useobject{currentmarker}{}%
\end{pgfscope}%
\end{pgfscope}%
\begin{pgfscope}%
\definecolor{textcolor}{rgb}{0.000000,0.000000,0.000000}%
\pgfsetstrokecolor{textcolor}%
\pgfsetfillcolor{textcolor}%
\pgftext[x=2.759720in,y=0.430778in,,top]{\color{textcolor}\sffamily\fontsize{14.000000}{16.800000}\selectfont 30}%
\end{pgfscope}%
\begin{pgfscope}%
\pgfsetbuttcap%
\pgfsetroundjoin%
\definecolor{currentfill}{rgb}{0.000000,0.000000,0.000000}%
\pgfsetfillcolor{currentfill}%
\pgfsetlinewidth{0.803000pt}%
\definecolor{currentstroke}{rgb}{0.000000,0.000000,0.000000}%
\pgfsetstrokecolor{currentstroke}%
\pgfsetdash{}{0pt}%
\pgfsys@defobject{currentmarker}{\pgfqpoint{0.000000in}{-0.048611in}}{\pgfqpoint{0.000000in}{0.000000in}}{%
\pgfpathmoveto{\pgfqpoint{0.000000in}{0.000000in}}%
\pgfpathlineto{\pgfqpoint{0.000000in}{-0.048611in}}%
\pgfusepath{stroke,fill}%
}%
\begin{pgfscope}%
\pgfsys@transformshift{3.337809in}{0.528000in}%
\pgfsys@useobject{currentmarker}{}%
\end{pgfscope}%
\end{pgfscope}%
\begin{pgfscope}%
\definecolor{textcolor}{rgb}{0.000000,0.000000,0.000000}%
\pgfsetstrokecolor{textcolor}%
\pgfsetfillcolor{textcolor}%
\pgftext[x=3.337809in,y=0.430778in,,top]{\color{textcolor}\sffamily\fontsize{14.000000}{16.800000}\selectfont 40}%
\end{pgfscope}%
\begin{pgfscope}%
\pgfsetbuttcap%
\pgfsetroundjoin%
\definecolor{currentfill}{rgb}{0.000000,0.000000,0.000000}%
\pgfsetfillcolor{currentfill}%
\pgfsetlinewidth{0.803000pt}%
\definecolor{currentstroke}{rgb}{0.000000,0.000000,0.000000}%
\pgfsetstrokecolor{currentstroke}%
\pgfsetdash{}{0pt}%
\pgfsys@defobject{currentmarker}{\pgfqpoint{0.000000in}{-0.048611in}}{\pgfqpoint{0.000000in}{0.000000in}}{%
\pgfpathmoveto{\pgfqpoint{0.000000in}{0.000000in}}%
\pgfpathlineto{\pgfqpoint{0.000000in}{-0.048611in}}%
\pgfusepath{stroke,fill}%
}%
\begin{pgfscope}%
\pgfsys@transformshift{3.915897in}{0.528000in}%
\pgfsys@useobject{currentmarker}{}%
\end{pgfscope}%
\end{pgfscope}%
\begin{pgfscope}%
\definecolor{textcolor}{rgb}{0.000000,0.000000,0.000000}%
\pgfsetstrokecolor{textcolor}%
\pgfsetfillcolor{textcolor}%
\pgftext[x=3.915897in,y=0.430778in,,top]{\color{textcolor}\sffamily\fontsize{14.000000}{16.800000}\selectfont 50}%
\end{pgfscope}%
\begin{pgfscope}%
\pgfsetbuttcap%
\pgfsetroundjoin%
\definecolor{currentfill}{rgb}{0.000000,0.000000,0.000000}%
\pgfsetfillcolor{currentfill}%
\pgfsetlinewidth{0.803000pt}%
\definecolor{currentstroke}{rgb}{0.000000,0.000000,0.000000}%
\pgfsetstrokecolor{currentstroke}%
\pgfsetdash{}{0pt}%
\pgfsys@defobject{currentmarker}{\pgfqpoint{0.000000in}{-0.048611in}}{\pgfqpoint{0.000000in}{0.000000in}}{%
\pgfpathmoveto{\pgfqpoint{0.000000in}{0.000000in}}%
\pgfpathlineto{\pgfqpoint{0.000000in}{-0.048611in}}%
\pgfusepath{stroke,fill}%
}%
\begin{pgfscope}%
\pgfsys@transformshift{4.493986in}{0.528000in}%
\pgfsys@useobject{currentmarker}{}%
\end{pgfscope}%
\end{pgfscope}%
\begin{pgfscope}%
\definecolor{textcolor}{rgb}{0.000000,0.000000,0.000000}%
\pgfsetstrokecolor{textcolor}%
\pgfsetfillcolor{textcolor}%
\pgftext[x=4.493986in,y=0.430778in,,top]{\color{textcolor}\sffamily\fontsize{14.000000}{16.800000}\selectfont 60}%
\end{pgfscope}%
\begin{pgfscope}%
\pgfsetbuttcap%
\pgfsetroundjoin%
\definecolor{currentfill}{rgb}{0.000000,0.000000,0.000000}%
\pgfsetfillcolor{currentfill}%
\pgfsetlinewidth{0.803000pt}%
\definecolor{currentstroke}{rgb}{0.000000,0.000000,0.000000}%
\pgfsetstrokecolor{currentstroke}%
\pgfsetdash{}{0pt}%
\pgfsys@defobject{currentmarker}{\pgfqpoint{0.000000in}{-0.048611in}}{\pgfqpoint{0.000000in}{0.000000in}}{%
\pgfpathmoveto{\pgfqpoint{0.000000in}{0.000000in}}%
\pgfpathlineto{\pgfqpoint{0.000000in}{-0.048611in}}%
\pgfusepath{stroke,fill}%
}%
\begin{pgfscope}%
\pgfsys@transformshift{5.072075in}{0.528000in}%
\pgfsys@useobject{currentmarker}{}%
\end{pgfscope}%
\end{pgfscope}%
\begin{pgfscope}%
\definecolor{textcolor}{rgb}{0.000000,0.000000,0.000000}%
\pgfsetstrokecolor{textcolor}%
\pgfsetfillcolor{textcolor}%
\pgftext[x=5.072075in,y=0.430778in,,top]{\color{textcolor}\sffamily\fontsize{14.000000}{16.800000}\selectfont 70}%
\end{pgfscope}%
\begin{pgfscope}%
\pgfsetbuttcap%
\pgfsetroundjoin%
\definecolor{currentfill}{rgb}{0.000000,0.000000,0.000000}%
\pgfsetfillcolor{currentfill}%
\pgfsetlinewidth{0.803000pt}%
\definecolor{currentstroke}{rgb}{0.000000,0.000000,0.000000}%
\pgfsetstrokecolor{currentstroke}%
\pgfsetdash{}{0pt}%
\pgfsys@defobject{currentmarker}{\pgfqpoint{0.000000in}{-0.048611in}}{\pgfqpoint{0.000000in}{0.000000in}}{%
\pgfpathmoveto{\pgfqpoint{0.000000in}{0.000000in}}%
\pgfpathlineto{\pgfqpoint{0.000000in}{-0.048611in}}%
\pgfusepath{stroke,fill}%
}%
\begin{pgfscope}%
\pgfsys@transformshift{5.650163in}{0.528000in}%
\pgfsys@useobject{currentmarker}{}%
\end{pgfscope}%
\end{pgfscope}%
\begin{pgfscope}%
\definecolor{textcolor}{rgb}{0.000000,0.000000,0.000000}%
\pgfsetstrokecolor{textcolor}%
\pgfsetfillcolor{textcolor}%
\pgftext[x=5.650163in,y=0.430778in,,top]{\color{textcolor}\sffamily\fontsize{14.000000}{16.800000}\selectfont 80}%
\end{pgfscope}%
\begin{pgfscope}%
\definecolor{textcolor}{rgb}{0.000000,0.000000,0.000000}%
\pgfsetstrokecolor{textcolor}%
\pgfsetfillcolor{textcolor}%
\pgftext[x=3.280000in,y=0.187044in,,top]{\color{textcolor}\sffamily\fontsize{16.000000}{19.200000}\selectfont Batches}%
\end{pgfscope}%
\begin{pgfscope}%
\pgfsetbuttcap%
\pgfsetroundjoin%
\definecolor{currentfill}{rgb}{0.000000,0.000000,0.000000}%
\pgfsetfillcolor{currentfill}%
\pgfsetlinewidth{0.803000pt}%
\definecolor{currentstroke}{rgb}{0.000000,0.000000,0.000000}%
\pgfsetstrokecolor{currentstroke}%
\pgfsetdash{}{0pt}%
\pgfsys@defobject{currentmarker}{\pgfqpoint{-0.048611in}{0.000000in}}{\pgfqpoint{-0.000000in}{0.000000in}}{%
\pgfpathmoveto{\pgfqpoint{-0.000000in}{0.000000in}}%
\pgfpathlineto{\pgfqpoint{-0.048611in}{0.000000in}}%
\pgfusepath{stroke,fill}%
}%
\begin{pgfscope}%
\pgfsys@transformshift{0.800000in}{1.187204in}%
\pgfsys@useobject{currentmarker}{}%
\end{pgfscope}%
\end{pgfscope}%
\begin{pgfscope}%
\definecolor{textcolor}{rgb}{0.000000,0.000000,0.000000}%
\pgfsetstrokecolor{textcolor}%
\pgfsetfillcolor{textcolor}%
\pgftext[x=0.455355in, y=1.113338in, left, base]{\color{textcolor}\sffamily\fontsize{14.000000}{16.800000}\selectfont 60}%
\end{pgfscope}%
\begin{pgfscope}%
\pgfsetbuttcap%
\pgfsetroundjoin%
\definecolor{currentfill}{rgb}{0.000000,0.000000,0.000000}%
\pgfsetfillcolor{currentfill}%
\pgfsetlinewidth{0.803000pt}%
\definecolor{currentstroke}{rgb}{0.000000,0.000000,0.000000}%
\pgfsetstrokecolor{currentstroke}%
\pgfsetdash{}{0pt}%
\pgfsys@defobject{currentmarker}{\pgfqpoint{-0.048611in}{0.000000in}}{\pgfqpoint{-0.000000in}{0.000000in}}{%
\pgfpathmoveto{\pgfqpoint{-0.000000in}{0.000000in}}%
\pgfpathlineto{\pgfqpoint{-0.048611in}{0.000000in}}%
\pgfusepath{stroke,fill}%
}%
\begin{pgfscope}%
\pgfsys@transformshift{0.800000in}{1.875165in}%
\pgfsys@useobject{currentmarker}{}%
\end{pgfscope}%
\end{pgfscope}%
\begin{pgfscope}%
\definecolor{textcolor}{rgb}{0.000000,0.000000,0.000000}%
\pgfsetstrokecolor{textcolor}%
\pgfsetfillcolor{textcolor}%
\pgftext[x=0.455355in, y=1.801299in, left, base]{\color{textcolor}\sffamily\fontsize{14.000000}{16.800000}\selectfont 65}%
\end{pgfscope}%
\begin{pgfscope}%
\pgfsetbuttcap%
\pgfsetroundjoin%
\definecolor{currentfill}{rgb}{0.000000,0.000000,0.000000}%
\pgfsetfillcolor{currentfill}%
\pgfsetlinewidth{0.803000pt}%
\definecolor{currentstroke}{rgb}{0.000000,0.000000,0.000000}%
\pgfsetstrokecolor{currentstroke}%
\pgfsetdash{}{0pt}%
\pgfsys@defobject{currentmarker}{\pgfqpoint{-0.048611in}{0.000000in}}{\pgfqpoint{-0.000000in}{0.000000in}}{%
\pgfpathmoveto{\pgfqpoint{-0.000000in}{0.000000in}}%
\pgfpathlineto{\pgfqpoint{-0.048611in}{0.000000in}}%
\pgfusepath{stroke,fill}%
}%
\begin{pgfscope}%
\pgfsys@transformshift{0.800000in}{2.563125in}%
\pgfsys@useobject{currentmarker}{}%
\end{pgfscope}%
\end{pgfscope}%
\begin{pgfscope}%
\definecolor{textcolor}{rgb}{0.000000,0.000000,0.000000}%
\pgfsetstrokecolor{textcolor}%
\pgfsetfillcolor{textcolor}%
\pgftext[x=0.455355in, y=2.489259in, left, base]{\color{textcolor}\sffamily\fontsize{14.000000}{16.800000}\selectfont 70}%
\end{pgfscope}%
\begin{pgfscope}%
\pgfsetbuttcap%
\pgfsetroundjoin%
\definecolor{currentfill}{rgb}{0.000000,0.000000,0.000000}%
\pgfsetfillcolor{currentfill}%
\pgfsetlinewidth{0.803000pt}%
\definecolor{currentstroke}{rgb}{0.000000,0.000000,0.000000}%
\pgfsetstrokecolor{currentstroke}%
\pgfsetdash{}{0pt}%
\pgfsys@defobject{currentmarker}{\pgfqpoint{-0.048611in}{0.000000in}}{\pgfqpoint{-0.000000in}{0.000000in}}{%
\pgfpathmoveto{\pgfqpoint{-0.000000in}{0.000000in}}%
\pgfpathlineto{\pgfqpoint{-0.048611in}{0.000000in}}%
\pgfusepath{stroke,fill}%
}%
\begin{pgfscope}%
\pgfsys@transformshift{0.800000in}{3.251086in}%
\pgfsys@useobject{currentmarker}{}%
\end{pgfscope}%
\end{pgfscope}%
\begin{pgfscope}%
\definecolor{textcolor}{rgb}{0.000000,0.000000,0.000000}%
\pgfsetstrokecolor{textcolor}%
\pgfsetfillcolor{textcolor}%
\pgftext[x=0.455355in, y=3.177220in, left, base]{\color{textcolor}\sffamily\fontsize{14.000000}{16.800000}\selectfont 75}%
\end{pgfscope}%
\begin{pgfscope}%
\pgfsetbuttcap%
\pgfsetroundjoin%
\definecolor{currentfill}{rgb}{0.000000,0.000000,0.000000}%
\pgfsetfillcolor{currentfill}%
\pgfsetlinewidth{0.803000pt}%
\definecolor{currentstroke}{rgb}{0.000000,0.000000,0.000000}%
\pgfsetstrokecolor{currentstroke}%
\pgfsetdash{}{0pt}%
\pgfsys@defobject{currentmarker}{\pgfqpoint{-0.048611in}{0.000000in}}{\pgfqpoint{-0.000000in}{0.000000in}}{%
\pgfpathmoveto{\pgfqpoint{-0.000000in}{0.000000in}}%
\pgfpathlineto{\pgfqpoint{-0.048611in}{0.000000in}}%
\pgfusepath{stroke,fill}%
}%
\begin{pgfscope}%
\pgfsys@transformshift{0.800000in}{3.939047in}%
\pgfsys@useobject{currentmarker}{}%
\end{pgfscope}%
\end{pgfscope}%
\begin{pgfscope}%
\definecolor{textcolor}{rgb}{0.000000,0.000000,0.000000}%
\pgfsetstrokecolor{textcolor}%
\pgfsetfillcolor{textcolor}%
\pgftext[x=0.455355in, y=3.865181in, left, base]{\color{textcolor}\sffamily\fontsize{14.000000}{16.800000}\selectfont 80}%
\end{pgfscope}%
\begin{pgfscope}%
\definecolor{textcolor}{rgb}{0.000000,0.000000,0.000000}%
\pgfsetstrokecolor{textcolor}%
\pgfsetfillcolor{textcolor}%
\pgftext[x=0.399799in,y=2.376000in,,bottom,rotate=90.000000]{\color{textcolor}\sffamily\fontsize{16.000000}{19.200000}\selectfont Top-1 Accuracy}%
\end{pgfscope}%
\begin{pgfscope}%
\pgfpathrectangle{\pgfqpoint{0.800000in}{0.528000in}}{\pgfqpoint{4.960000in}{3.696000in}}%
\pgfusepath{clip}%
\pgfsetrectcap%
\pgfsetroundjoin%
\pgfsetlinewidth{1.505625pt}%
\definecolor{currentstroke}{rgb}{0.196078,0.803922,0.196078}%
\pgfsetstrokecolor{currentstroke}%
\pgfsetdash{}{0pt}%
\pgfpathmoveto{\pgfqpoint{1.025455in}{1.337179in}}%
\pgfpathlineto{\pgfqpoint{1.083263in}{1.703174in}}%
\pgfpathlineto{\pgfqpoint{1.141072in}{1.765091in}}%
\pgfpathlineto{\pgfqpoint{1.198881in}{1.855902in}}%
\pgfpathlineto{\pgfqpoint{1.256690in}{1.886172in}}%
\pgfpathlineto{\pgfqpoint{1.314499in}{1.916442in}}%
\pgfpathlineto{\pgfqpoint{1.372308in}{1.920570in}}%
\pgfpathlineto{\pgfqpoint{1.430117in}{1.957720in}}%
\pgfpathlineto{\pgfqpoint{1.487925in}{1.959096in}}%
\pgfpathlineto{\pgfqpoint{1.545734in}{1.948088in}}%
\pgfpathlineto{\pgfqpoint{1.603543in}{1.970103in}}%
\pgfpathlineto{\pgfqpoint{1.661352in}{1.971479in}}%
\pgfpathlineto{\pgfqpoint{1.719161in}{1.975607in}}%
\pgfpathlineto{\pgfqpoint{1.776970in}{1.957720in}}%
\pgfpathlineto{\pgfqpoint{1.834779in}{1.945337in}}%
\pgfpathlineto{\pgfqpoint{1.892587in}{1.932953in}}%
\pgfpathlineto{\pgfqpoint{1.950396in}{1.916442in}}%
\pgfpathlineto{\pgfqpoint{2.008205in}{1.916442in}}%
\pgfpathlineto{\pgfqpoint{2.066014in}{1.891676in}}%
\pgfpathlineto{\pgfqpoint{2.123823in}{1.895803in}}%
\pgfpathlineto{\pgfqpoint{2.181632in}{1.880668in}}%
\pgfpathlineto{\pgfqpoint{2.239441in}{1.868285in}}%
\pgfpathlineto{\pgfqpoint{2.297249in}{1.858654in}}%
\pgfpathlineto{\pgfqpoint{2.355058in}{1.839391in}}%
\pgfpathlineto{\pgfqpoint{2.412867in}{1.825631in}}%
\pgfpathlineto{\pgfqpoint{2.470676in}{1.811872in}}%
\pgfpathlineto{\pgfqpoint{2.528485in}{1.785730in}}%
\pgfpathlineto{\pgfqpoint{2.586294in}{1.785730in}}%
\pgfpathlineto{\pgfqpoint{2.644103in}{1.752708in}}%
\pgfpathlineto{\pgfqpoint{2.701911in}{1.734821in}}%
\pgfpathlineto{\pgfqpoint{2.759720in}{1.710054in}}%
\pgfpathlineto{\pgfqpoint{2.817529in}{1.699047in}}%
\pgfpathlineto{\pgfqpoint{2.875338in}{1.681160in}}%
\pgfpathlineto{\pgfqpoint{2.933147in}{1.659145in}}%
\pgfpathlineto{\pgfqpoint{2.990956in}{1.648138in}}%
\pgfpathlineto{\pgfqpoint{3.048765in}{1.627499in}}%
\pgfpathlineto{\pgfqpoint{3.106573in}{1.601356in}}%
\pgfpathlineto{\pgfqpoint{3.164382in}{1.594477in}}%
\pgfpathlineto{\pgfqpoint{3.222191in}{1.580717in}}%
\pgfpathlineto{\pgfqpoint{3.280000in}{1.557327in}}%
\pgfpathlineto{\pgfqpoint{3.337809in}{1.532560in}}%
\pgfpathlineto{\pgfqpoint{3.395618in}{1.507794in}}%
\pgfpathlineto{\pgfqpoint{3.453427in}{1.485779in}}%
\pgfpathlineto{\pgfqpoint{3.511235in}{1.462388in}}%
\pgfpathlineto{\pgfqpoint{3.569044in}{1.434870in}}%
\pgfpathlineto{\pgfqpoint{3.626853in}{1.407351in}}%
\pgfpathlineto{\pgfqpoint{3.684662in}{1.374329in}}%
\pgfpathlineto{\pgfqpoint{3.742471in}{1.348187in}}%
\pgfpathlineto{\pgfqpoint{3.800280in}{1.320668in}}%
\pgfpathlineto{\pgfqpoint{3.858089in}{1.297278in}}%
\pgfpathlineto{\pgfqpoint{3.915897in}{1.269759in}}%
\pgfpathlineto{\pgfqpoint{3.973706in}{1.242241in}}%
\pgfpathlineto{\pgfqpoint{4.031515in}{1.214722in}}%
\pgfpathlineto{\pgfqpoint{4.089324in}{1.188580in}}%
\pgfpathlineto{\pgfqpoint{4.147133in}{1.161061in}}%
\pgfpathlineto{\pgfqpoint{4.204942in}{1.132167in}}%
\pgfpathlineto{\pgfqpoint{4.262751in}{1.107400in}}%
\pgfpathlineto{\pgfqpoint{4.320559in}{1.081258in}}%
\pgfpathlineto{\pgfqpoint{4.378368in}{1.063371in}}%
\pgfpathlineto{\pgfqpoint{4.436177in}{1.034477in}}%
\pgfpathlineto{\pgfqpoint{4.493986in}{1.016590in}}%
\pgfpathlineto{\pgfqpoint{4.551795in}{0.993199in}}%
\pgfpathlineto{\pgfqpoint{4.609604in}{0.971184in}}%
\pgfpathlineto{\pgfqpoint{4.667413in}{0.949170in}}%
\pgfpathlineto{\pgfqpoint{4.725221in}{0.927155in}}%
\pgfpathlineto{\pgfqpoint{4.783030in}{0.907892in}}%
\pgfpathlineto{\pgfqpoint{4.840839in}{0.884501in}}%
\pgfpathlineto{\pgfqpoint{4.898648in}{0.872118in}}%
\pgfpathlineto{\pgfqpoint{4.956457in}{0.850103in}}%
\pgfpathlineto{\pgfqpoint{5.014266in}{0.830840in}}%
\pgfpathlineto{\pgfqpoint{5.072075in}{0.814329in}}%
\pgfpathlineto{\pgfqpoint{5.129883in}{0.797818in}}%
\pgfpathlineto{\pgfqpoint{5.187692in}{0.779931in}}%
\pgfpathlineto{\pgfqpoint{5.245501in}{0.763420in}}%
\pgfpathlineto{\pgfqpoint{5.303310in}{0.748285in}}%
\pgfpathlineto{\pgfqpoint{5.361119in}{0.731774in}}%
\pgfpathlineto{\pgfqpoint{5.418928in}{0.713887in}}%
\pgfpathlineto{\pgfqpoint{5.476737in}{0.698752in}}%
\pgfpathlineto{\pgfqpoint{5.534545in}{0.696000in}}%
\pgfusepath{stroke}%
\end{pgfscope}%
\begin{pgfscope}%
\pgfpathrectangle{\pgfqpoint{0.800000in}{0.528000in}}{\pgfqpoint{4.960000in}{3.696000in}}%
\pgfusepath{clip}%
\pgfsetrectcap%
\pgfsetroundjoin%
\pgfsetlinewidth{1.505625pt}%
\definecolor{currentstroke}{rgb}{1.000000,0.549020,0.000000}%
\pgfsetstrokecolor{currentstroke}%
\pgfsetdash{}{0pt}%
\pgfpathmoveto{\pgfqpoint{1.025455in}{1.509170in}}%
\pgfpathlineto{\pgfqpoint{1.083263in}{1.961848in}}%
\pgfpathlineto{\pgfqpoint{1.141072in}{2.125582in}}%
\pgfpathlineto{\pgfqpoint{1.198881in}{2.308580in}}%
\pgfpathlineto{\pgfqpoint{1.256690in}{2.440668in}}%
\pgfpathlineto{\pgfqpoint{1.314499in}{2.547990in}}%
\pgfpathlineto{\pgfqpoint{1.372308in}{2.633297in}}%
\pgfpathlineto{\pgfqpoint{1.430117in}{2.733740in}}%
\pgfpathlineto{\pgfqpoint{1.487925in}{2.828678in}}%
\pgfpathlineto{\pgfqpoint{1.545734in}{2.889219in}}%
\pgfpathlineto{\pgfqpoint{1.603543in}{2.977278in}}%
\pgfpathlineto{\pgfqpoint{1.661352in}{3.062585in}}%
\pgfpathlineto{\pgfqpoint{1.719161in}{3.121749in}}%
\pgfpathlineto{\pgfqpoint{1.776970in}{3.189170in}}%
\pgfpathlineto{\pgfqpoint{1.834779in}{3.230447in}}%
\pgfpathlineto{\pgfqpoint{1.892587in}{3.275853in}}%
\pgfpathlineto{\pgfqpoint{1.950396in}{3.326762in}}%
\pgfpathlineto{\pgfqpoint{2.008205in}{3.368039in}}%
\pgfpathlineto{\pgfqpoint{2.066014in}{3.407941in}}%
\pgfpathlineto{\pgfqpoint{2.123823in}{3.450595in}}%
\pgfpathlineto{\pgfqpoint{2.181632in}{3.482241in}}%
\pgfpathlineto{\pgfqpoint{2.239441in}{3.501504in}}%
\pgfpathlineto{\pgfqpoint{2.297249in}{3.520767in}}%
\pgfpathlineto{\pgfqpoint{2.355058in}{3.552413in}}%
\pgfpathlineto{\pgfqpoint{2.412867in}{3.577179in}}%
\pgfpathlineto{\pgfqpoint{2.470676in}{3.603322in}}%
\pgfpathlineto{\pgfqpoint{2.528485in}{3.619833in}}%
\pgfpathlineto{\pgfqpoint{2.586294in}{3.637720in}}%
\pgfpathlineto{\pgfqpoint{2.644103in}{3.659735in}}%
\pgfpathlineto{\pgfqpoint{2.701911in}{3.673494in}}%
\pgfpathlineto{\pgfqpoint{2.759720in}{3.694133in}}%
\pgfpathlineto{\pgfqpoint{2.817529in}{3.712020in}}%
\pgfpathlineto{\pgfqpoint{2.875338in}{3.725779in}}%
\pgfpathlineto{\pgfqpoint{2.933147in}{3.739538in}}%
\pgfpathlineto{\pgfqpoint{2.990956in}{3.751921in}}%
\pgfpathlineto{\pgfqpoint{3.048765in}{3.761553in}}%
\pgfpathlineto{\pgfqpoint{3.106573in}{3.779440in}}%
\pgfpathlineto{\pgfqpoint{3.164382in}{3.791823in}}%
\pgfpathlineto{\pgfqpoint{3.222191in}{3.806958in}}%
\pgfpathlineto{\pgfqpoint{3.280000in}{3.813838in}}%
\pgfpathlineto{\pgfqpoint{3.337809in}{3.816590in}}%
\pgfpathlineto{\pgfqpoint{3.395618in}{3.824845in}}%
\pgfpathlineto{\pgfqpoint{3.453427in}{3.833101in}}%
\pgfpathlineto{\pgfqpoint{3.511235in}{3.844108in}}%
\pgfpathlineto{\pgfqpoint{3.569044in}{3.852364in}}%
\pgfpathlineto{\pgfqpoint{3.626853in}{3.861995in}}%
\pgfpathlineto{\pgfqpoint{3.684662in}{3.866123in}}%
\pgfpathlineto{\pgfqpoint{3.742471in}{3.877130in}}%
\pgfpathlineto{\pgfqpoint{3.800280in}{3.879882in}}%
\pgfpathlineto{\pgfqpoint{3.858089in}{3.889514in}}%
\pgfpathlineto{\pgfqpoint{3.915897in}{3.899145in}}%
\pgfpathlineto{\pgfqpoint{3.973706in}{3.907400in}}%
\pgfpathlineto{\pgfqpoint{4.031515in}{3.914280in}}%
\pgfpathlineto{\pgfqpoint{4.089324in}{3.918408in}}%
\pgfpathlineto{\pgfqpoint{4.147133in}{3.926663in}}%
\pgfpathlineto{\pgfqpoint{4.204942in}{3.932167in}}%
\pgfpathlineto{\pgfqpoint{4.262751in}{3.939047in}}%
\pgfpathlineto{\pgfqpoint{4.320559in}{3.945926in}}%
\pgfpathlineto{\pgfqpoint{4.378368in}{3.954182in}}%
\pgfpathlineto{\pgfqpoint{4.436177in}{3.958310in}}%
\pgfpathlineto{\pgfqpoint{4.493986in}{3.965189in}}%
\pgfpathlineto{\pgfqpoint{4.551795in}{3.970693in}}%
\pgfpathlineto{\pgfqpoint{4.609604in}{3.976197in}}%
\pgfpathlineto{\pgfqpoint{4.667413in}{3.981700in}}%
\pgfpathlineto{\pgfqpoint{4.725221in}{3.987204in}}%
\pgfpathlineto{\pgfqpoint{4.783030in}{3.992708in}}%
\pgfpathlineto{\pgfqpoint{4.840839in}{3.998211in}}%
\pgfpathlineto{\pgfqpoint{4.898648in}{4.002339in}}%
\pgfpathlineto{\pgfqpoint{4.956457in}{4.009219in}}%
\pgfpathlineto{\pgfqpoint{5.014266in}{4.014722in}}%
\pgfpathlineto{\pgfqpoint{5.072075in}{4.020226in}}%
\pgfpathlineto{\pgfqpoint{5.129883in}{4.025730in}}%
\pgfpathlineto{\pgfqpoint{5.187692in}{4.031233in}}%
\pgfpathlineto{\pgfqpoint{5.245501in}{4.038113in}}%
\pgfpathlineto{\pgfqpoint{5.303310in}{4.042241in}}%
\pgfpathlineto{\pgfqpoint{5.361119in}{4.046369in}}%
\pgfpathlineto{\pgfqpoint{5.418928in}{4.051872in}}%
\pgfpathlineto{\pgfqpoint{5.476737in}{4.056000in}}%
\pgfpathlineto{\pgfqpoint{5.534545in}{4.056000in}}%
\pgfusepath{stroke}%
\end{pgfscope}%
\begin{pgfscope}%
\pgfpathrectangle{\pgfqpoint{0.800000in}{0.528000in}}{\pgfqpoint{4.960000in}{3.696000in}}%
\pgfusepath{clip}%
\pgfsetbuttcap%
\pgfsetroundjoin%
\pgfsetlinewidth{1.505625pt}%
\definecolor{currentstroke}{rgb}{0.000000,0.000000,0.000000}%
\pgfsetstrokecolor{currentstroke}%
\pgfsetdash{{5.550000pt}{2.400000pt}}{0.000000pt}%
\pgfpathmoveto{\pgfqpoint{0.800000in}{1.214722in}}%
\pgfpathlineto{\pgfqpoint{5.760000in}{1.214722in}}%
\pgfusepath{stroke}%
\end{pgfscope}%
\begin{pgfscope}%
\pgfsetrectcap%
\pgfsetmiterjoin%
\pgfsetlinewidth{0.803000pt}%
\definecolor{currentstroke}{rgb}{0.000000,0.000000,0.000000}%
\pgfsetstrokecolor{currentstroke}%
\pgfsetdash{}{0pt}%
\pgfpathmoveto{\pgfqpoint{0.800000in}{0.528000in}}%
\pgfpathlineto{\pgfqpoint{0.800000in}{4.224000in}}%
\pgfusepath{stroke}%
\end{pgfscope}%
\begin{pgfscope}%
\pgfsetrectcap%
\pgfsetmiterjoin%
\pgfsetlinewidth{0.803000pt}%
\definecolor{currentstroke}{rgb}{0.000000,0.000000,0.000000}%
\pgfsetstrokecolor{currentstroke}%
\pgfsetdash{}{0pt}%
\pgfpathmoveto{\pgfqpoint{5.760000in}{0.528000in}}%
\pgfpathlineto{\pgfqpoint{5.760000in}{4.224000in}}%
\pgfusepath{stroke}%
\end{pgfscope}%
\begin{pgfscope}%
\pgfsetrectcap%
\pgfsetmiterjoin%
\pgfsetlinewidth{0.803000pt}%
\definecolor{currentstroke}{rgb}{0.000000,0.000000,0.000000}%
\pgfsetstrokecolor{currentstroke}%
\pgfsetdash{}{0pt}%
\pgfpathmoveto{\pgfqpoint{0.800000in}{0.528000in}}%
\pgfpathlineto{\pgfqpoint{5.760000in}{0.528000in}}%
\pgfusepath{stroke}%
\end{pgfscope}%
\begin{pgfscope}%
\pgfsetrectcap%
\pgfsetmiterjoin%
\pgfsetlinewidth{0.803000pt}%
\definecolor{currentstroke}{rgb}{0.000000,0.000000,0.000000}%
\pgfsetstrokecolor{currentstroke}%
\pgfsetdash{}{0pt}%
\pgfpathmoveto{\pgfqpoint{0.800000in}{4.224000in}}%
\pgfpathlineto{\pgfqpoint{5.760000in}{4.224000in}}%
\pgfusepath{stroke}%
\end{pgfscope}%
\begin{pgfscope}%
\definecolor{textcolor}{rgb}{0.000000,0.000000,0.000000}%
\pgfsetstrokecolor{textcolor}%
\pgfsetfillcolor{textcolor}%
\pgftext[x=0.909837in,y=1.049612in,left,base]{\color{textcolor}\sffamily\fontsize{12.000000}{14.400000}\bfseries\selectfont CLIP}%
\end{pgfscope}%
\begin{pgfscope}%
\definecolor{textcolor}{rgb}{0.196078,0.803922,0.196078}%
\pgfsetstrokecolor{textcolor}%
\pgfsetfillcolor{textcolor}%
\pgftext[x=4.927552in,y=0.912020in,left,base]{\color{textcolor}\sffamily\fontsize{12.000000}{14.400000}\bfseries\selectfont TENT}%
\end{pgfscope}%
\begin{pgfscope}%
\definecolor{textcolor}{rgb}{1.000000,0.549020,0.000000}%
\pgfsetstrokecolor{textcolor}%
\pgfsetfillcolor{textcolor}%
\pgftext[x=2.181632in,y=3.870251in,left,base]{\color{textcolor}\sffamily\fontsize{10.000000}{12.000000}\bfseries\selectfont CLIPTTA (ours)}%
\end{pgfscope}%
\end{pgfpicture}%
\makeatother%
\endgroup%

%% file: figures/improvement2.pgf
\begingroup%
\makeatletter%
\begin{pgfpicture}%
\pgfpathrectangle{\pgfpointorigin}{\pgfqpoint{6.400000in}{4.800000in}}%
\pgfusepath{use as bounding box, clip}%
\begin{pgfscope}%
\pgfsetbuttcap%
\pgfsetmiterjoin%
\definecolor{currentfill}{rgb}{1.000000,1.000000,1.000000}%
\pgfsetfillcolor{currentfill}%
\pgfsetlinewidth{0.000000pt}%
\definecolor{currentstroke}{rgb}{1.000000,1.000000,1.000000}%
\pgfsetstrokecolor{currentstroke}%
\pgfsetdash{}{0pt}%
\pgfpathmoveto{\pgfqpoint{0.000000in}{0.000000in}}%
\pgfpathlineto{\pgfqpoint{6.400000in}{0.000000in}}%
\pgfpathlineto{\pgfqpoint{6.400000in}{4.800000in}}%
\pgfpathlineto{\pgfqpoint{0.000000in}{4.800000in}}%
\pgfpathlineto{\pgfqpoint{0.000000in}{0.000000in}}%
\pgfpathclose%
\pgfusepath{fill}%
\end{pgfscope}%
\begin{pgfscope}%
\pgfsetbuttcap%
\pgfsetmiterjoin%
\definecolor{currentfill}{rgb}{1.000000,1.000000,1.000000}%
\pgfsetfillcolor{currentfill}%
\pgfsetlinewidth{0.000000pt}%
\definecolor{currentstroke}{rgb}{0.000000,0.000000,0.000000}%
\pgfsetstrokecolor{currentstroke}%
\pgfsetstrokeopacity{0.000000}%
\pgfsetdash{}{0pt}%
\pgfpathmoveto{\pgfqpoint{0.827070in}{0.706016in}}%
\pgfpathlineto{\pgfqpoint{6.236125in}{0.706016in}}%
\pgfpathlineto{\pgfqpoint{6.236125in}{4.650000in}}%
\pgfpathlineto{\pgfqpoint{0.827070in}{4.650000in}}%
\pgfpathlineto{\pgfqpoint{0.827070in}{0.706016in}}%
\pgfpathclose%
\pgfusepath{fill}%
\end{pgfscope}%
\begin{pgfscope}%
\pgfsetbuttcap%
\pgfsetroundjoin%
\definecolor{currentfill}{rgb}{0.000000,0.000000,0.000000}%
\pgfsetfillcolor{currentfill}%
\pgfsetlinewidth{0.803000pt}%
\definecolor{currentstroke}{rgb}{0.000000,0.000000,0.000000}%
\pgfsetstrokecolor{currentstroke}%
\pgfsetdash{}{0pt}%
\pgfsys@defobject{currentmarker}{\pgfqpoint{0.000000in}{-0.048611in}}{\pgfqpoint{0.000000in}{0.000000in}}{%
\pgfpathmoveto{\pgfqpoint{0.000000in}{0.000000in}}%
\pgfpathlineto{\pgfqpoint{0.000000in}{-0.048611in}}%
\pgfusepath{stroke,fill}%
}%
\begin{pgfscope}%
\pgfsys@transformshift{1.072936in}{0.706016in}%
\pgfsys@useobject{currentmarker}{}%
\end{pgfscope}%
\end{pgfscope}%
\begin{pgfscope}%
\definecolor{textcolor}{rgb}{0.000000,0.000000,0.000000}%
\pgfsetstrokecolor{textcolor}%
\pgfsetfillcolor{textcolor}%
\pgftext[x=1.072936in,y=0.608794in,,top]{\color{textcolor}\sffamily\fontsize{14.000000}{16.800000}\selectfont 0}%
\end{pgfscope}%
\begin{pgfscope}%
\pgfsetbuttcap%
\pgfsetroundjoin%
\definecolor{currentfill}{rgb}{0.000000,0.000000,0.000000}%
\pgfsetfillcolor{currentfill}%
\pgfsetlinewidth{0.803000pt}%
\definecolor{currentstroke}{rgb}{0.000000,0.000000,0.000000}%
\pgfsetstrokecolor{currentstroke}%
\pgfsetdash{}{0pt}%
\pgfsys@defobject{currentmarker}{\pgfqpoint{0.000000in}{-0.048611in}}{\pgfqpoint{0.000000in}{0.000000in}}{%
\pgfpathmoveto{\pgfqpoint{0.000000in}{0.000000in}}%
\pgfpathlineto{\pgfqpoint{0.000000in}{-0.048611in}}%
\pgfusepath{stroke,fill}%
}%
\begin{pgfscope}%
\pgfsys@transformshift{1.703362in}{0.706016in}%
\pgfsys@useobject{currentmarker}{}%
\end{pgfscope}%
\end{pgfscope}%
\begin{pgfscope}%
\definecolor{textcolor}{rgb}{0.000000,0.000000,0.000000}%
\pgfsetstrokecolor{textcolor}%
\pgfsetfillcolor{textcolor}%
\pgftext[x=1.703362in,y=0.608794in,,top]{\color{textcolor}\sffamily\fontsize{14.000000}{16.800000}\selectfont 10}%
\end{pgfscope}%
\begin{pgfscope}%
\pgfsetbuttcap%
\pgfsetroundjoin%
\definecolor{currentfill}{rgb}{0.000000,0.000000,0.000000}%
\pgfsetfillcolor{currentfill}%
\pgfsetlinewidth{0.803000pt}%
\definecolor{currentstroke}{rgb}{0.000000,0.000000,0.000000}%
\pgfsetstrokecolor{currentstroke}%
\pgfsetdash{}{0pt}%
\pgfsys@defobject{currentmarker}{\pgfqpoint{0.000000in}{-0.048611in}}{\pgfqpoint{0.000000in}{0.000000in}}{%
\pgfpathmoveto{\pgfqpoint{0.000000in}{0.000000in}}%
\pgfpathlineto{\pgfqpoint{0.000000in}{-0.048611in}}%
\pgfusepath{stroke,fill}%
}%
\begin{pgfscope}%
\pgfsys@transformshift{2.333788in}{0.706016in}%
\pgfsys@useobject{currentmarker}{}%
\end{pgfscope}%
\end{pgfscope}%
\begin{pgfscope}%
\definecolor{textcolor}{rgb}{0.000000,0.000000,0.000000}%
\pgfsetstrokecolor{textcolor}%
\pgfsetfillcolor{textcolor}%
\pgftext[x=2.333788in,y=0.608794in,,top]{\color{textcolor}\sffamily\fontsize{14.000000}{16.800000}\selectfont 20}%
\end{pgfscope}%
\begin{pgfscope}%
\pgfsetbuttcap%
\pgfsetroundjoin%
\definecolor{currentfill}{rgb}{0.000000,0.000000,0.000000}%
\pgfsetfillcolor{currentfill}%
\pgfsetlinewidth{0.803000pt}%
\definecolor{currentstroke}{rgb}{0.000000,0.000000,0.000000}%
\pgfsetstrokecolor{currentstroke}%
\pgfsetdash{}{0pt}%
\pgfsys@defobject{currentmarker}{\pgfqpoint{0.000000in}{-0.048611in}}{\pgfqpoint{0.000000in}{0.000000in}}{%
\pgfpathmoveto{\pgfqpoint{0.000000in}{0.000000in}}%
\pgfpathlineto{\pgfqpoint{0.000000in}{-0.048611in}}%
\pgfusepath{stroke,fill}%
}%
\begin{pgfscope}%
\pgfsys@transformshift{2.964214in}{0.706016in}%
\pgfsys@useobject{currentmarker}{}%
\end{pgfscope}%
\end{pgfscope}%
\begin{pgfscope}%
\definecolor{textcolor}{rgb}{0.000000,0.000000,0.000000}%
\pgfsetstrokecolor{textcolor}%
\pgfsetfillcolor{textcolor}%
\pgftext[x=2.964214in,y=0.608794in,,top]{\color{textcolor}\sffamily\fontsize{14.000000}{16.800000}\selectfont 30}%
\end{pgfscope}%
\begin{pgfscope}%
\pgfsetbuttcap%
\pgfsetroundjoin%
\definecolor{currentfill}{rgb}{0.000000,0.000000,0.000000}%
\pgfsetfillcolor{currentfill}%
\pgfsetlinewidth{0.803000pt}%
\definecolor{currentstroke}{rgb}{0.000000,0.000000,0.000000}%
\pgfsetstrokecolor{currentstroke}%
\pgfsetdash{}{0pt}%
\pgfsys@defobject{currentmarker}{\pgfqpoint{0.000000in}{-0.048611in}}{\pgfqpoint{0.000000in}{0.000000in}}{%
\pgfpathmoveto{\pgfqpoint{0.000000in}{0.000000in}}%
\pgfpathlineto{\pgfqpoint{0.000000in}{-0.048611in}}%
\pgfusepath{stroke,fill}%
}%
\begin{pgfscope}%
\pgfsys@transformshift{3.594640in}{0.706016in}%
\pgfsys@useobject{currentmarker}{}%
\end{pgfscope}%
\end{pgfscope}%
\begin{pgfscope}%
\definecolor{textcolor}{rgb}{0.000000,0.000000,0.000000}%
\pgfsetstrokecolor{textcolor}%
\pgfsetfillcolor{textcolor}%
\pgftext[x=3.594640in,y=0.608794in,,top]{\color{textcolor}\sffamily\fontsize{14.000000}{16.800000}\selectfont 40}%
\end{pgfscope}%
\begin{pgfscope}%
\pgfsetbuttcap%
\pgfsetroundjoin%
\definecolor{currentfill}{rgb}{0.000000,0.000000,0.000000}%
\pgfsetfillcolor{currentfill}%
\pgfsetlinewidth{0.803000pt}%
\definecolor{currentstroke}{rgb}{0.000000,0.000000,0.000000}%
\pgfsetstrokecolor{currentstroke}%
\pgfsetdash{}{0pt}%
\pgfsys@defobject{currentmarker}{\pgfqpoint{0.000000in}{-0.048611in}}{\pgfqpoint{0.000000in}{0.000000in}}{%
\pgfpathmoveto{\pgfqpoint{0.000000in}{0.000000in}}%
\pgfpathlineto{\pgfqpoint{0.000000in}{-0.048611in}}%
\pgfusepath{stroke,fill}%
}%
\begin{pgfscope}%
\pgfsys@transformshift{4.225066in}{0.706016in}%
\pgfsys@useobject{currentmarker}{}%
\end{pgfscope}%
\end{pgfscope}%
\begin{pgfscope}%
\definecolor{textcolor}{rgb}{0.000000,0.000000,0.000000}%
\pgfsetstrokecolor{textcolor}%
\pgfsetfillcolor{textcolor}%
\pgftext[x=4.225066in,y=0.608794in,,top]{\color{textcolor}\sffamily\fontsize{14.000000}{16.800000}\selectfont 50}%
\end{pgfscope}%
\begin{pgfscope}%
\pgfsetbuttcap%
\pgfsetroundjoin%
\definecolor{currentfill}{rgb}{0.000000,0.000000,0.000000}%
\pgfsetfillcolor{currentfill}%
\pgfsetlinewidth{0.803000pt}%
\definecolor{currentstroke}{rgb}{0.000000,0.000000,0.000000}%
\pgfsetstrokecolor{currentstroke}%
\pgfsetdash{}{0pt}%
\pgfsys@defobject{currentmarker}{\pgfqpoint{0.000000in}{-0.048611in}}{\pgfqpoint{0.000000in}{0.000000in}}{%
\pgfpathmoveto{\pgfqpoint{0.000000in}{0.000000in}}%
\pgfpathlineto{\pgfqpoint{0.000000in}{-0.048611in}}%
\pgfusepath{stroke,fill}%
}%
\begin{pgfscope}%
\pgfsys@transformshift{4.855492in}{0.706016in}%
\pgfsys@useobject{currentmarker}{}%
\end{pgfscope}%
\end{pgfscope}%
\begin{pgfscope}%
\definecolor{textcolor}{rgb}{0.000000,0.000000,0.000000}%
\pgfsetstrokecolor{textcolor}%
\pgfsetfillcolor{textcolor}%
\pgftext[x=4.855492in,y=0.608794in,,top]{\color{textcolor}\sffamily\fontsize{14.000000}{16.800000}\selectfont 60}%
\end{pgfscope}%
\begin{pgfscope}%
\pgfsetbuttcap%
\pgfsetroundjoin%
\definecolor{currentfill}{rgb}{0.000000,0.000000,0.000000}%
\pgfsetfillcolor{currentfill}%
\pgfsetlinewidth{0.803000pt}%
\definecolor{currentstroke}{rgb}{0.000000,0.000000,0.000000}%
\pgfsetstrokecolor{currentstroke}%
\pgfsetdash{}{0pt}%
\pgfsys@defobject{currentmarker}{\pgfqpoint{0.000000in}{-0.048611in}}{\pgfqpoint{0.000000in}{0.000000in}}{%
\pgfpathmoveto{\pgfqpoint{0.000000in}{0.000000in}}%
\pgfpathlineto{\pgfqpoint{0.000000in}{-0.048611in}}%
\pgfusepath{stroke,fill}%
}%
\begin{pgfscope}%
\pgfsys@transformshift{5.485918in}{0.706016in}%
\pgfsys@useobject{currentmarker}{}%
\end{pgfscope}%
\end{pgfscope}%
\begin{pgfscope}%
\definecolor{textcolor}{rgb}{0.000000,0.000000,0.000000}%
\pgfsetstrokecolor{textcolor}%
\pgfsetfillcolor{textcolor}%
\pgftext[x=5.485918in,y=0.608794in,,top]{\color{textcolor}\sffamily\fontsize{14.000000}{16.800000}\selectfont 70}%
\end{pgfscope}%
\begin{pgfscope}%
\pgfsetbuttcap%
\pgfsetroundjoin%
\definecolor{currentfill}{rgb}{0.000000,0.000000,0.000000}%
\pgfsetfillcolor{currentfill}%
\pgfsetlinewidth{0.803000pt}%
\definecolor{currentstroke}{rgb}{0.000000,0.000000,0.000000}%
\pgfsetstrokecolor{currentstroke}%
\pgfsetdash{}{0pt}%
\pgfsys@defobject{currentmarker}{\pgfqpoint{0.000000in}{-0.048611in}}{\pgfqpoint{0.000000in}{0.000000in}}{%
\pgfpathmoveto{\pgfqpoint{0.000000in}{0.000000in}}%
\pgfpathlineto{\pgfqpoint{0.000000in}{-0.048611in}}%
\pgfusepath{stroke,fill}%
}%
\begin{pgfscope}%
\pgfsys@transformshift{6.116344in}{0.706016in}%
\pgfsys@useobject{currentmarker}{}%
\end{pgfscope}%
\end{pgfscope}%
\begin{pgfscope}%
\definecolor{textcolor}{rgb}{0.000000,0.000000,0.000000}%
\pgfsetstrokecolor{textcolor}%
\pgfsetfillcolor{textcolor}%
\pgftext[x=6.116344in,y=0.608794in,,top]{\color{textcolor}\sffamily\fontsize{14.000000}{16.800000}\selectfont 80}%
\end{pgfscope}%
\begin{pgfscope}%
\definecolor{textcolor}{rgb}{0.000000,0.000000,0.000000}%
\pgfsetstrokecolor{textcolor}%
\pgfsetfillcolor{textcolor}%
\pgftext[x=3.531597in,y=0.365061in,,top]{\color{textcolor}\sffamily\fontsize{16.000000}{19.200000}\selectfont Batches}%
\end{pgfscope}%
\begin{pgfscope}%
\pgfsetbuttcap%
\pgfsetroundjoin%
\definecolor{currentfill}{rgb}{0.000000,0.000000,0.000000}%
\pgfsetfillcolor{currentfill}%
\pgfsetlinewidth{0.803000pt}%
\definecolor{currentstroke}{rgb}{0.000000,0.000000,0.000000}%
\pgfsetstrokecolor{currentstroke}%
\pgfsetdash{}{0pt}%
\pgfsys@defobject{currentmarker}{\pgfqpoint{-0.048611in}{0.000000in}}{\pgfqpoint{-0.000000in}{0.000000in}}{%
\pgfpathmoveto{\pgfqpoint{-0.000000in}{0.000000in}}%
\pgfpathlineto{\pgfqpoint{-0.048611in}{0.000000in}}%
\pgfusepath{stroke,fill}%
}%
\begin{pgfscope}%
\pgfsys@transformshift{0.827070in}{1.010277in}%
\pgfsys@useobject{currentmarker}{}%
\end{pgfscope}%
\end{pgfscope}%
\begin{pgfscope}%
\definecolor{textcolor}{rgb}{0.000000,0.000000,0.000000}%
\pgfsetstrokecolor{textcolor}%
\pgfsetfillcolor{textcolor}%
\pgftext[x=0.420616in, y=0.936411in, left, base]{\color{textcolor}\sffamily\fontsize{14.000000}{16.800000}\selectfont 0.1}%
\end{pgfscope}%
\begin{pgfscope}%
\pgfsetbuttcap%
\pgfsetroundjoin%
\definecolor{currentfill}{rgb}{0.000000,0.000000,0.000000}%
\pgfsetfillcolor{currentfill}%
\pgfsetlinewidth{0.803000pt}%
\definecolor{currentstroke}{rgb}{0.000000,0.000000,0.000000}%
\pgfsetstrokecolor{currentstroke}%
\pgfsetdash{}{0pt}%
\pgfsys@defobject{currentmarker}{\pgfqpoint{-0.048611in}{0.000000in}}{\pgfqpoint{-0.000000in}{0.000000in}}{%
\pgfpathmoveto{\pgfqpoint{-0.000000in}{0.000000in}}%
\pgfpathlineto{\pgfqpoint{-0.048611in}{0.000000in}}%
\pgfusepath{stroke,fill}%
}%
\begin{pgfscope}%
\pgfsys@transformshift{0.827070in}{1.675109in}%
\pgfsys@useobject{currentmarker}{}%
\end{pgfscope}%
\end{pgfscope}%
\begin{pgfscope}%
\definecolor{textcolor}{rgb}{0.000000,0.000000,0.000000}%
\pgfsetstrokecolor{textcolor}%
\pgfsetfillcolor{textcolor}%
\pgftext[x=0.420616in, y=1.601243in, left, base]{\color{textcolor}\sffamily\fontsize{14.000000}{16.800000}\selectfont 0.2}%
\end{pgfscope}%
\begin{pgfscope}%
\pgfsetbuttcap%
\pgfsetroundjoin%
\definecolor{currentfill}{rgb}{0.000000,0.000000,0.000000}%
\pgfsetfillcolor{currentfill}%
\pgfsetlinewidth{0.803000pt}%
\definecolor{currentstroke}{rgb}{0.000000,0.000000,0.000000}%
\pgfsetstrokecolor{currentstroke}%
\pgfsetdash{}{0pt}%
\pgfsys@defobject{currentmarker}{\pgfqpoint{-0.048611in}{0.000000in}}{\pgfqpoint{-0.000000in}{0.000000in}}{%
\pgfpathmoveto{\pgfqpoint{-0.000000in}{0.000000in}}%
\pgfpathlineto{\pgfqpoint{-0.048611in}{0.000000in}}%
\pgfusepath{stroke,fill}%
}%
\begin{pgfscope}%
\pgfsys@transformshift{0.827070in}{2.339941in}%
\pgfsys@useobject{currentmarker}{}%
\end{pgfscope}%
\end{pgfscope}%
\begin{pgfscope}%
\definecolor{textcolor}{rgb}{0.000000,0.000000,0.000000}%
\pgfsetstrokecolor{textcolor}%
\pgfsetfillcolor{textcolor}%
\pgftext[x=0.420616in, y=2.266075in, left, base]{\color{textcolor}\sffamily\fontsize{14.000000}{16.800000}\selectfont 0.3}%
\end{pgfscope}%
\begin{pgfscope}%
\pgfsetbuttcap%
\pgfsetroundjoin%
\definecolor{currentfill}{rgb}{0.000000,0.000000,0.000000}%
\pgfsetfillcolor{currentfill}%
\pgfsetlinewidth{0.803000pt}%
\definecolor{currentstroke}{rgb}{0.000000,0.000000,0.000000}%
\pgfsetstrokecolor{currentstroke}%
\pgfsetdash{}{0pt}%
\pgfsys@defobject{currentmarker}{\pgfqpoint{-0.048611in}{0.000000in}}{\pgfqpoint{-0.000000in}{0.000000in}}{%
\pgfpathmoveto{\pgfqpoint{-0.000000in}{0.000000in}}%
\pgfpathlineto{\pgfqpoint{-0.048611in}{0.000000in}}%
\pgfusepath{stroke,fill}%
}%
\begin{pgfscope}%
\pgfsys@transformshift{0.827070in}{3.004773in}%
\pgfsys@useobject{currentmarker}{}%
\end{pgfscope}%
\end{pgfscope}%
\begin{pgfscope}%
\definecolor{textcolor}{rgb}{0.000000,0.000000,0.000000}%
\pgfsetstrokecolor{textcolor}%
\pgfsetfillcolor{textcolor}%
\pgftext[x=0.420616in, y=2.930907in, left, base]{\color{textcolor}\sffamily\fontsize{14.000000}{16.800000}\selectfont 0.4}%
\end{pgfscope}%
\begin{pgfscope}%
\pgfsetbuttcap%
\pgfsetroundjoin%
\definecolor{currentfill}{rgb}{0.000000,0.000000,0.000000}%
\pgfsetfillcolor{currentfill}%
\pgfsetlinewidth{0.803000pt}%
\definecolor{currentstroke}{rgb}{0.000000,0.000000,0.000000}%
\pgfsetstrokecolor{currentstroke}%
\pgfsetdash{}{0pt}%
\pgfsys@defobject{currentmarker}{\pgfqpoint{-0.048611in}{0.000000in}}{\pgfqpoint{-0.000000in}{0.000000in}}{%
\pgfpathmoveto{\pgfqpoint{-0.000000in}{0.000000in}}%
\pgfpathlineto{\pgfqpoint{-0.048611in}{0.000000in}}%
\pgfusepath{stroke,fill}%
}%
\begin{pgfscope}%
\pgfsys@transformshift{0.827070in}{3.669605in}%
\pgfsys@useobject{currentmarker}{}%
\end{pgfscope}%
\end{pgfscope}%
\begin{pgfscope}%
\definecolor{textcolor}{rgb}{0.000000,0.000000,0.000000}%
\pgfsetstrokecolor{textcolor}%
\pgfsetfillcolor{textcolor}%
\pgftext[x=0.420616in, y=3.595739in, left, base]{\color{textcolor}\sffamily\fontsize{14.000000}{16.800000}\selectfont 0.5}%
\end{pgfscope}%
\begin{pgfscope}%
\pgfsetbuttcap%
\pgfsetroundjoin%
\definecolor{currentfill}{rgb}{0.000000,0.000000,0.000000}%
\pgfsetfillcolor{currentfill}%
\pgfsetlinewidth{0.803000pt}%
\definecolor{currentstroke}{rgb}{0.000000,0.000000,0.000000}%
\pgfsetstrokecolor{currentstroke}%
\pgfsetdash{}{0pt}%
\pgfsys@defobject{currentmarker}{\pgfqpoint{-0.048611in}{0.000000in}}{\pgfqpoint{-0.000000in}{0.000000in}}{%
\pgfpathmoveto{\pgfqpoint{-0.000000in}{0.000000in}}%
\pgfpathlineto{\pgfqpoint{-0.048611in}{0.000000in}}%
\pgfusepath{stroke,fill}%
}%
\begin{pgfscope}%
\pgfsys@transformshift{0.827070in}{4.334437in}%
\pgfsys@useobject{currentmarker}{}%
\end{pgfscope}%
\end{pgfscope}%
\begin{pgfscope}%
\definecolor{textcolor}{rgb}{0.000000,0.000000,0.000000}%
\pgfsetstrokecolor{textcolor}%
\pgfsetfillcolor{textcolor}%
\pgftext[x=0.420616in, y=4.260571in, left, base]{\color{textcolor}\sffamily\fontsize{14.000000}{16.800000}\selectfont 0.6}%
\end{pgfscope}%
\begin{pgfscope}%
\definecolor{textcolor}{rgb}{0.000000,0.000000,0.000000}%
\pgfsetstrokecolor{textcolor}%
\pgfsetfillcolor{textcolor}%
\pgftext[x=0.365061in,y=2.678008in,,bottom,rotate=90.000000]{\color{textcolor}\sffamily\fontsize{16.000000}{19.200000}\selectfont Improvement}%
\end{pgfscope}%
\begin{pgfscope}%
\pgfpathrectangle{\pgfqpoint{0.827070in}{0.706016in}}{\pgfqpoint{5.409056in}{3.943984in}}%
\pgfusepath{clip}%
\pgfsetrectcap%
\pgfsetroundjoin%
\pgfsetlinewidth{1.505625pt}%
\definecolor{currentstroke}{rgb}{0.235294,0.701961,0.443137}%
\pgfsetstrokecolor{currentstroke}%
\pgfsetdash{}{0pt}%
\pgfpathmoveto{\pgfqpoint{1.072936in}{0.885288in}}%
\pgfpathlineto{\pgfqpoint{1.135978in}{1.091386in}}%
\pgfpathlineto{\pgfqpoint{1.199021in}{1.341363in}}%
\pgfpathlineto{\pgfqpoint{1.262064in}{1.506242in}}%
\pgfpathlineto{\pgfqpoint{1.325106in}{1.610620in}}%
\pgfpathlineto{\pgfqpoint{1.388149in}{1.753559in}}%
\pgfpathlineto{\pgfqpoint{1.451191in}{1.841982in}}%
\pgfpathlineto{\pgfqpoint{1.514234in}{1.953674in}}%
\pgfpathlineto{\pgfqpoint{1.577277in}{1.998882in}}%
\pgfpathlineto{\pgfqpoint{1.640319in}{2.073343in}}%
\pgfpathlineto{\pgfqpoint{1.703362in}{2.167750in}}%
\pgfpathlineto{\pgfqpoint{1.766404in}{2.212958in}}%
\pgfpathlineto{\pgfqpoint{1.829447in}{2.272128in}}%
\pgfpathlineto{\pgfqpoint{1.892490in}{2.304705in}}%
\pgfpathlineto{\pgfqpoint{1.955532in}{2.349249in}}%
\pgfpathlineto{\pgfqpoint{2.018575in}{2.385150in}}%
\pgfpathlineto{\pgfqpoint{2.081617in}{2.427034in}}%
\pgfpathlineto{\pgfqpoint{2.144660in}{2.460276in}}%
\pgfpathlineto{\pgfqpoint{2.207703in}{2.490858in}}%
\pgfpathlineto{\pgfqpoint{2.270745in}{2.521440in}}%
\pgfpathlineto{\pgfqpoint{2.333788in}{2.543380in}}%
\pgfpathlineto{\pgfqpoint{2.396831in}{2.563325in}}%
\pgfpathlineto{\pgfqpoint{2.459873in}{2.587923in}}%
\pgfpathlineto{\pgfqpoint{2.522916in}{2.601220in}}%
\pgfpathlineto{\pgfqpoint{2.585958in}{2.618506in}}%
\pgfpathlineto{\pgfqpoint{2.649001in}{2.635127in}}%
\pgfpathlineto{\pgfqpoint{2.712044in}{2.642440in}}%
\pgfpathlineto{\pgfqpoint{2.775086in}{2.653742in}}%
\pgfpathlineto{\pgfqpoint{2.838129in}{2.664379in}}%
\pgfpathlineto{\pgfqpoint{2.901171in}{2.675016in}}%
\pgfpathlineto{\pgfqpoint{2.964214in}{2.688313in}}%
\pgfpathlineto{\pgfqpoint{3.027257in}{2.704269in}}%
\pgfpathlineto{\pgfqpoint{3.090299in}{2.714906in}}%
\pgfpathlineto{\pgfqpoint{3.153342in}{2.724214in}}%
\pgfpathlineto{\pgfqpoint{3.216384in}{2.735516in}}%
\pgfpathlineto{\pgfqpoint{3.279427in}{2.746153in}}%
\pgfpathlineto{\pgfqpoint{3.342470in}{2.752802in}}%
\pgfpathlineto{\pgfqpoint{3.405512in}{2.763439in}}%
\pgfpathlineto{\pgfqpoint{3.468555in}{2.771417in}}%
\pgfpathlineto{\pgfqpoint{3.531597in}{2.774741in}}%
\pgfpathlineto{\pgfqpoint{3.594640in}{2.773412in}}%
\pgfpathlineto{\pgfqpoint{3.657683in}{2.777401in}}%
\pgfpathlineto{\pgfqpoint{3.720725in}{2.782054in}}%
\pgfpathlineto{\pgfqpoint{3.783768in}{2.784049in}}%
\pgfpathlineto{\pgfqpoint{3.846811in}{2.780725in}}%
\pgfpathlineto{\pgfqpoint{3.909853in}{2.778065in}}%
\pgfpathlineto{\pgfqpoint{3.972896in}{2.776736in}}%
\pgfpathlineto{\pgfqpoint{4.035938in}{2.775406in}}%
\pgfpathlineto{\pgfqpoint{4.098981in}{2.773412in}}%
\pgfpathlineto{\pgfqpoint{4.162024in}{2.770752in}}%
\pgfpathlineto{\pgfqpoint{4.225066in}{2.770087in}}%
\pgfpathlineto{\pgfqpoint{4.288109in}{2.768758in}}%
\pgfpathlineto{\pgfqpoint{4.351151in}{2.766098in}}%
\pgfpathlineto{\pgfqpoint{4.414194in}{2.764104in}}%
\pgfpathlineto{\pgfqpoint{4.477237in}{2.762109in}}%
\pgfpathlineto{\pgfqpoint{4.540279in}{2.758120in}}%
\pgfpathlineto{\pgfqpoint{4.603322in}{2.756126in}}%
\pgfpathlineto{\pgfqpoint{4.666364in}{2.760115in}}%
\pgfpathlineto{\pgfqpoint{4.729407in}{2.752137in}}%
\pgfpathlineto{\pgfqpoint{4.792450in}{2.748813in}}%
\pgfpathlineto{\pgfqpoint{4.855492in}{2.750142in}}%
\pgfpathlineto{\pgfqpoint{4.918535in}{2.747483in}}%
\pgfpathlineto{\pgfqpoint{4.981577in}{2.746818in}}%
\pgfpathlineto{\pgfqpoint{5.044620in}{2.745489in}}%
\pgfpathlineto{\pgfqpoint{5.107663in}{2.745489in}}%
\pgfpathlineto{\pgfqpoint{5.170705in}{2.745489in}}%
\pgfpathlineto{\pgfqpoint{5.233748in}{2.741500in}}%
\pgfpathlineto{\pgfqpoint{5.296791in}{2.740835in}}%
\pgfpathlineto{\pgfqpoint{5.359833in}{2.736846in}}%
\pgfpathlineto{\pgfqpoint{5.422876in}{2.736846in}}%
\pgfpathlineto{\pgfqpoint{5.485918in}{2.736846in}}%
\pgfpathlineto{\pgfqpoint{5.548961in}{2.734851in}}%
\pgfpathlineto{\pgfqpoint{5.612004in}{2.734851in}}%
\pgfpathlineto{\pgfqpoint{5.675046in}{2.736181in}}%
\pgfpathlineto{\pgfqpoint{5.738089in}{2.734851in}}%
\pgfpathlineto{\pgfqpoint{5.801131in}{2.732857in}}%
\pgfpathlineto{\pgfqpoint{5.864174in}{2.727538in}}%
\pgfpathlineto{\pgfqpoint{5.927217in}{2.726873in}}%
\pgfpathlineto{\pgfqpoint{5.990259in}{2.726873in}}%
\pgfusepath{stroke}%
\end{pgfscope}%
\begin{pgfscope}%
\pgfpathrectangle{\pgfqpoint{0.827070in}{0.706016in}}{\pgfqpoint{5.409056in}{3.943984in}}%
\pgfusepath{clip}%
\pgfsetrectcap%
\pgfsetroundjoin%
\pgfsetlinewidth{1.505625pt}%
\definecolor{currentstroke}{rgb}{1.000000,0.549020,0.000000}%
\pgfsetstrokecolor{currentstroke}%
\pgfsetdash{}{0pt}%
\pgfpathmoveto{\pgfqpoint{1.072936in}{1.088062in}}%
\pgfpathlineto{\pgfqpoint{1.135978in}{1.395215in}}%
\pgfpathlineto{\pgfqpoint{1.199021in}{1.730290in}}%
\pgfpathlineto{\pgfqpoint{1.262064in}{1.972289in}}%
\pgfpathlineto{\pgfqpoint{1.325106in}{2.153123in}}%
\pgfpathlineto{\pgfqpoint{1.388149in}{2.339941in}}%
\pgfpathlineto{\pgfqpoint{1.451191in}{2.487534in}}%
\pgfpathlineto{\pgfqpoint{1.514234in}{2.629143in}}%
\pgfpathlineto{\pgfqpoint{1.577277in}{2.749478in}}%
\pgfpathlineto{\pgfqpoint{1.640319in}{2.849867in}}%
\pgfpathlineto{\pgfqpoint{1.703362in}{2.968872in}}%
\pgfpathlineto{\pgfqpoint{1.766404in}{3.065273in}}%
\pgfpathlineto{\pgfqpoint{1.829447in}{3.146382in}}%
\pgfpathlineto{\pgfqpoint{1.892490in}{3.231481in}}%
\pgfpathlineto{\pgfqpoint{1.955532in}{3.289986in}}%
\pgfpathlineto{\pgfqpoint{2.018575in}{3.359794in}}%
\pgfpathlineto{\pgfqpoint{2.081617in}{3.433590in}}%
\pgfpathlineto{\pgfqpoint{2.144660in}{3.489436in}}%
\pgfpathlineto{\pgfqpoint{2.207703in}{3.547276in}}%
\pgfpathlineto{\pgfqpoint{2.270745in}{3.603122in}}%
\pgfpathlineto{\pgfqpoint{2.333788in}{3.648996in}}%
\pgfpathlineto{\pgfqpoint{2.396831in}{3.684232in}}%
\pgfpathlineto{\pgfqpoint{2.459873in}{3.722792in}}%
\pgfpathlineto{\pgfqpoint{2.522916in}{3.754039in}}%
\pgfpathlineto{\pgfqpoint{2.585958in}{3.787281in}}%
\pgfpathlineto{\pgfqpoint{2.649001in}{3.818528in}}%
\pgfpathlineto{\pgfqpoint{2.712044in}{3.843126in}}%
\pgfpathlineto{\pgfqpoint{2.775086in}{3.869055in}}%
\pgfpathlineto{\pgfqpoint{2.838129in}{3.900302in}}%
\pgfpathlineto{\pgfqpoint{2.901171in}{3.918917in}}%
\pgfpathlineto{\pgfqpoint{2.964214in}{3.949500in}}%
\pgfpathlineto{\pgfqpoint{3.027257in}{3.975428in}}%
\pgfpathlineto{\pgfqpoint{3.090299in}{3.999362in}}%
\pgfpathlineto{\pgfqpoint{3.153342in}{4.023961in}}%
\pgfpathlineto{\pgfqpoint{3.216384in}{4.047230in}}%
\pgfpathlineto{\pgfqpoint{3.279427in}{4.071164in}}%
\pgfpathlineto{\pgfqpoint{3.342470in}{4.089779in}}%
\pgfpathlineto{\pgfqpoint{3.405512in}{4.111054in}}%
\pgfpathlineto{\pgfqpoint{3.468555in}{4.128339in}}%
\pgfpathlineto{\pgfqpoint{3.531597in}{4.139642in}}%
\pgfpathlineto{\pgfqpoint{3.594640in}{4.145625in}}%
\pgfpathlineto{\pgfqpoint{3.657683in}{4.161581in}}%
\pgfpathlineto{\pgfqpoint{3.720725in}{4.174213in}}%
\pgfpathlineto{\pgfqpoint{3.783768in}{4.188839in}}%
\pgfpathlineto{\pgfqpoint{3.846811in}{4.200141in}}%
\pgfpathlineto{\pgfqpoint{3.909853in}{4.213438in}}%
\pgfpathlineto{\pgfqpoint{3.972896in}{4.220751in}}%
\pgfpathlineto{\pgfqpoint{4.035938in}{4.232718in}}%
\pgfpathlineto{\pgfqpoint{4.098981in}{4.241361in}}%
\pgfpathlineto{\pgfqpoint{4.162024in}{4.250004in}}%
\pgfpathlineto{\pgfqpoint{4.225066in}{4.263300in}}%
\pgfpathlineto{\pgfqpoint{4.288109in}{4.274603in}}%
\pgfpathlineto{\pgfqpoint{4.351151in}{4.281251in}}%
\pgfpathlineto{\pgfqpoint{4.414194in}{4.290559in}}%
\pgfpathlineto{\pgfqpoint{4.477237in}{4.299201in}}%
\pgfpathlineto{\pgfqpoint{4.540279in}{4.306514in}}%
\pgfpathlineto{\pgfqpoint{4.603322in}{4.315157in}}%
\pgfpathlineto{\pgfqpoint{4.666364in}{4.323135in}}%
\pgfpathlineto{\pgfqpoint{4.729407in}{4.333108in}}%
\pgfpathlineto{\pgfqpoint{4.792450in}{4.339756in}}%
\pgfpathlineto{\pgfqpoint{4.855492in}{4.345740in}}%
\pgfpathlineto{\pgfqpoint{4.918535in}{4.353053in}}%
\pgfpathlineto{\pgfqpoint{4.981577in}{4.362360in}}%
\pgfpathlineto{\pgfqpoint{5.044620in}{4.369674in}}%
\pgfpathlineto{\pgfqpoint{5.107663in}{4.379646in}}%
\pgfpathlineto{\pgfqpoint{5.170705in}{4.386959in}}%
\pgfpathlineto{\pgfqpoint{5.233748in}{4.390948in}}%
\pgfpathlineto{\pgfqpoint{5.296791in}{4.398261in}}%
\pgfpathlineto{\pgfqpoint{5.359833in}{4.406904in}}%
\pgfpathlineto{\pgfqpoint{5.422876in}{4.414217in}}%
\pgfpathlineto{\pgfqpoint{5.485918in}{4.422195in}}%
\pgfpathlineto{\pgfqpoint{5.548961in}{4.428844in}}%
\pgfpathlineto{\pgfqpoint{5.612004in}{4.437486in}}%
\pgfpathlineto{\pgfqpoint{5.675046in}{4.446794in}}%
\pgfpathlineto{\pgfqpoint{5.738089in}{4.453442in}}%
\pgfpathlineto{\pgfqpoint{5.801131in}{4.459426in}}%
\pgfpathlineto{\pgfqpoint{5.864174in}{4.464080in}}%
\pgfpathlineto{\pgfqpoint{5.927217in}{4.470728in}}%
\pgfpathlineto{\pgfqpoint{5.990259in}{4.470728in}}%
\pgfusepath{stroke}%
\end{pgfscope}%
\begin{pgfscope}%
\pgfsetrectcap%
\pgfsetmiterjoin%
\pgfsetlinewidth{0.803000pt}%
\definecolor{currentstroke}{rgb}{0.000000,0.000000,0.000000}%
\pgfsetstrokecolor{currentstroke}%
\pgfsetdash{}{0pt}%
\pgfpathmoveto{\pgfqpoint{0.827070in}{0.706016in}}%
\pgfpathlineto{\pgfqpoint{0.827070in}{4.650000in}}%
\pgfusepath{stroke}%
\end{pgfscope}%
\begin{pgfscope}%
\pgfsetrectcap%
\pgfsetmiterjoin%
\pgfsetlinewidth{0.803000pt}%
\definecolor{currentstroke}{rgb}{0.000000,0.000000,0.000000}%
\pgfsetstrokecolor{currentstroke}%
\pgfsetdash{}{0pt}%
\pgfpathmoveto{\pgfqpoint{6.236125in}{0.706016in}}%
\pgfpathlineto{\pgfqpoint{6.236125in}{4.650000in}}%
\pgfusepath{stroke}%
\end{pgfscope}%
\begin{pgfscope}%
\pgfsetrectcap%
\pgfsetmiterjoin%
\pgfsetlinewidth{0.803000pt}%
\definecolor{currentstroke}{rgb}{0.000000,0.000000,0.000000}%
\pgfsetstrokecolor{currentstroke}%
\pgfsetdash{}{0pt}%
\pgfpathmoveto{\pgfqpoint{0.827070in}{0.706016in}}%
\pgfpathlineto{\pgfqpoint{6.236125in}{0.706016in}}%
\pgfusepath{stroke}%
\end{pgfscope}%
\begin{pgfscope}%
\pgfsetrectcap%
\pgfsetmiterjoin%
\pgfsetlinewidth{0.803000pt}%
\definecolor{currentstroke}{rgb}{0.000000,0.000000,0.000000}%
\pgfsetstrokecolor{currentstroke}%
\pgfsetdash{}{0pt}%
\pgfpathmoveto{\pgfqpoint{0.827070in}{4.650000in}}%
\pgfpathlineto{\pgfqpoint{6.236125in}{4.650000in}}%
\pgfusepath{stroke}%
\end{pgfscope}%
\begin{pgfscope}%
\pgfsetbuttcap%
\pgfsetmiterjoin%
\definecolor{currentfill}{rgb}{1.000000,1.000000,1.000000}%
\pgfsetfillcolor{currentfill}%
\pgfsetfillopacity{0.800000}%
\pgfsetlinewidth{1.003750pt}%
\definecolor{currentstroke}{rgb}{0.800000,0.800000,0.800000}%
\pgfsetstrokecolor{currentstroke}%
\pgfsetstrokeopacity{0.800000}%
\pgfsetdash{}{0pt}%
\pgfpathmoveto{\pgfqpoint{4.488599in}{3.819879in}}%
\pgfpathlineto{\pgfqpoint{6.080570in}{3.819879in}}%
\pgfpathquadraticcurveto{\pgfqpoint{6.125014in}{3.819879in}}{\pgfqpoint{6.125014in}{3.864323in}}%
\pgfpathlineto{\pgfqpoint{6.125014in}{4.494444in}}%
\pgfpathquadraticcurveto{\pgfqpoint{6.125014in}{4.538889in}}{\pgfqpoint{6.080570in}{4.538889in}}%
\pgfpathlineto{\pgfqpoint{4.488599in}{4.538889in}}%
\pgfpathquadraticcurveto{\pgfqpoint{4.444155in}{4.538889in}}{\pgfqpoint{4.444155in}{4.494444in}}%
\pgfpathlineto{\pgfqpoint{4.444155in}{3.864323in}}%
\pgfpathquadraticcurveto{\pgfqpoint{4.444155in}{3.819879in}}{\pgfqpoint{4.488599in}{3.819879in}}%
\pgfpathlineto{\pgfqpoint{4.488599in}{3.819879in}}%
\pgfpathclose%
\pgfusepath{stroke,fill}%
\end{pgfscope}%
\begin{pgfscope}%
\pgfsetrectcap%
\pgfsetroundjoin%
\pgfsetlinewidth{1.505625pt}%
\definecolor{currentstroke}{rgb}{0.235294,0.701961,0.443137}%
\pgfsetstrokecolor{currentstroke}%
\pgfsetdash{}{0pt}%
\pgfpathmoveto{\pgfqpoint{4.533044in}{4.358941in}}%
\pgfpathlineto{\pgfqpoint{4.755266in}{4.358941in}}%
\pgfpathlineto{\pgfqpoint{4.977488in}{4.358941in}}%
\pgfusepath{stroke}%
\end{pgfscope}%
\begin{pgfscope}%
\definecolor{textcolor}{rgb}{0.000000,0.000000,0.000000}%
\pgfsetstrokecolor{textcolor}%
\pgfsetfillcolor{textcolor}%
\pgftext[x=5.155266in,y=4.281163in,left,base]{\color{textcolor}\sffamily\fontsize{16.000000}{19.200000}\selectfont TENT}%
\end{pgfscope}%
\begin{pgfscope}%
\pgfsetrectcap%
\pgfsetroundjoin%
\pgfsetlinewidth{1.505625pt}%
\definecolor{currentstroke}{rgb}{1.000000,0.549020,0.000000}%
\pgfsetstrokecolor{currentstroke}%
\pgfsetdash{}{0pt}%
\pgfpathmoveto{\pgfqpoint{4.533044in}{4.032769in}}%
\pgfpathlineto{\pgfqpoint{4.755266in}{4.032769in}}%
\pgfpathlineto{\pgfqpoint{4.977488in}{4.032769in}}%
\pgfusepath{stroke}%
\end{pgfscope}%
\begin{pgfscope}%
\definecolor{textcolor}{rgb}{0.000000,0.000000,0.000000}%
\pgfsetstrokecolor{textcolor}%
\pgfsetfillcolor{textcolor}%
\pgftext[x=5.155266in,y=3.954992in,left,base]{\color{textcolor}\sffamily\fontsize{16.000000}{19.200000}\selectfont CLIPTTA}%
\end{pgfscope}%
\end{pgfpicture}%
\makeatother%
\endgroup%

%% file: figures/deterioration2.pgf
\begingroup%
\makeatletter%
\begin{pgfpicture}%
\pgfpathrectangle{\pgfpointorigin}{\pgfqpoint{6.400000in}{4.800000in}}%
\pgfusepath{use as bounding box, clip}%
\begin{pgfscope}%
\pgfsetbuttcap%
\pgfsetmiterjoin%
\definecolor{currentfill}{rgb}{1.000000,1.000000,1.000000}%
\pgfsetfillcolor{currentfill}%
\pgfsetlinewidth{0.000000pt}%
\definecolor{currentstroke}{rgb}{1.000000,1.000000,1.000000}%
\pgfsetstrokecolor{currentstroke}%
\pgfsetdash{}{0pt}%
\pgfpathmoveto{\pgfqpoint{0.000000in}{0.000000in}}%
\pgfpathlineto{\pgfqpoint{6.400000in}{0.000000in}}%
\pgfpathlineto{\pgfqpoint{6.400000in}{4.800000in}}%
\pgfpathlineto{\pgfqpoint{0.000000in}{4.800000in}}%
\pgfpathlineto{\pgfqpoint{0.000000in}{0.000000in}}%
\pgfpathclose%
\pgfusepath{fill}%
\end{pgfscope}%
\begin{pgfscope}%
\pgfsetbuttcap%
\pgfsetmiterjoin%
\definecolor{currentfill}{rgb}{1.000000,1.000000,1.000000}%
\pgfsetfillcolor{currentfill}%
\pgfsetlinewidth{0.000000pt}%
\definecolor{currentstroke}{rgb}{0.000000,0.000000,0.000000}%
\pgfsetstrokecolor{currentstroke}%
\pgfsetstrokeopacity{0.000000}%
\pgfsetdash{}{0pt}%
\pgfpathmoveto{\pgfqpoint{0.950781in}{0.706016in}}%
\pgfpathlineto{\pgfqpoint{6.236125in}{0.706016in}}%
\pgfpathlineto{\pgfqpoint{6.236125in}{4.650000in}}%
\pgfpathlineto{\pgfqpoint{0.950781in}{4.650000in}}%
\pgfpathlineto{\pgfqpoint{0.950781in}{0.706016in}}%
\pgfpathclose%
\pgfusepath{fill}%
\end{pgfscope}%
\begin{pgfscope}%
\pgfsetbuttcap%
\pgfsetroundjoin%
\definecolor{currentfill}{rgb}{0.000000,0.000000,0.000000}%
\pgfsetfillcolor{currentfill}%
\pgfsetlinewidth{0.803000pt}%
\definecolor{currentstroke}{rgb}{0.000000,0.000000,0.000000}%
\pgfsetstrokecolor{currentstroke}%
\pgfsetdash{}{0pt}%
\pgfsys@defobject{currentmarker}{\pgfqpoint{0.000000in}{-0.048611in}}{\pgfqpoint{0.000000in}{0.000000in}}{%
\pgfpathmoveto{\pgfqpoint{0.000000in}{0.000000in}}%
\pgfpathlineto{\pgfqpoint{0.000000in}{-0.048611in}}%
\pgfusepath{stroke,fill}%
}%
\begin{pgfscope}%
\pgfsys@transformshift{1.191024in}{0.706016in}%
\pgfsys@useobject{currentmarker}{}%
\end{pgfscope}%
\end{pgfscope}%
\begin{pgfscope}%
\definecolor{textcolor}{rgb}{0.000000,0.000000,0.000000}%
\pgfsetstrokecolor{textcolor}%
\pgfsetfillcolor{textcolor}%
\pgftext[x=1.191024in,y=0.608794in,,top]{\color{textcolor}\sffamily\fontsize{14.000000}{16.800000}\selectfont 0}%
\end{pgfscope}%
\begin{pgfscope}%
\pgfsetbuttcap%
\pgfsetroundjoin%
\definecolor{currentfill}{rgb}{0.000000,0.000000,0.000000}%
\pgfsetfillcolor{currentfill}%
\pgfsetlinewidth{0.803000pt}%
\definecolor{currentstroke}{rgb}{0.000000,0.000000,0.000000}%
\pgfsetstrokecolor{currentstroke}%
\pgfsetdash{}{0pt}%
\pgfsys@defobject{currentmarker}{\pgfqpoint{0.000000in}{-0.048611in}}{\pgfqpoint{0.000000in}{0.000000in}}{%
\pgfpathmoveto{\pgfqpoint{0.000000in}{0.000000in}}%
\pgfpathlineto{\pgfqpoint{0.000000in}{-0.048611in}}%
\pgfusepath{stroke,fill}%
}%
\begin{pgfscope}%
\pgfsys@transformshift{1.807032in}{0.706016in}%
\pgfsys@useobject{currentmarker}{}%
\end{pgfscope}%
\end{pgfscope}%
\begin{pgfscope}%
\definecolor{textcolor}{rgb}{0.000000,0.000000,0.000000}%
\pgfsetstrokecolor{textcolor}%
\pgfsetfillcolor{textcolor}%
\pgftext[x=1.807032in,y=0.608794in,,top]{\color{textcolor}\sffamily\fontsize{14.000000}{16.800000}\selectfont 10}%
\end{pgfscope}%
\begin{pgfscope}%
\pgfsetbuttcap%
\pgfsetroundjoin%
\definecolor{currentfill}{rgb}{0.000000,0.000000,0.000000}%
\pgfsetfillcolor{currentfill}%
\pgfsetlinewidth{0.803000pt}%
\definecolor{currentstroke}{rgb}{0.000000,0.000000,0.000000}%
\pgfsetstrokecolor{currentstroke}%
\pgfsetdash{}{0pt}%
\pgfsys@defobject{currentmarker}{\pgfqpoint{0.000000in}{-0.048611in}}{\pgfqpoint{0.000000in}{0.000000in}}{%
\pgfpathmoveto{\pgfqpoint{0.000000in}{0.000000in}}%
\pgfpathlineto{\pgfqpoint{0.000000in}{-0.048611in}}%
\pgfusepath{stroke,fill}%
}%
\begin{pgfscope}%
\pgfsys@transformshift{2.423039in}{0.706016in}%
\pgfsys@useobject{currentmarker}{}%
\end{pgfscope}%
\end{pgfscope}%
\begin{pgfscope}%
\definecolor{textcolor}{rgb}{0.000000,0.000000,0.000000}%
\pgfsetstrokecolor{textcolor}%
\pgfsetfillcolor{textcolor}%
\pgftext[x=2.423039in,y=0.608794in,,top]{\color{textcolor}\sffamily\fontsize{14.000000}{16.800000}\selectfont 20}%
\end{pgfscope}%
\begin{pgfscope}%
\pgfsetbuttcap%
\pgfsetroundjoin%
\definecolor{currentfill}{rgb}{0.000000,0.000000,0.000000}%
\pgfsetfillcolor{currentfill}%
\pgfsetlinewidth{0.803000pt}%
\definecolor{currentstroke}{rgb}{0.000000,0.000000,0.000000}%
\pgfsetstrokecolor{currentstroke}%
\pgfsetdash{}{0pt}%
\pgfsys@defobject{currentmarker}{\pgfqpoint{0.000000in}{-0.048611in}}{\pgfqpoint{0.000000in}{0.000000in}}{%
\pgfpathmoveto{\pgfqpoint{0.000000in}{0.000000in}}%
\pgfpathlineto{\pgfqpoint{0.000000in}{-0.048611in}}%
\pgfusepath{stroke,fill}%
}%
\begin{pgfscope}%
\pgfsys@transformshift{3.039047in}{0.706016in}%
\pgfsys@useobject{currentmarker}{}%
\end{pgfscope}%
\end{pgfscope}%
\begin{pgfscope}%
\definecolor{textcolor}{rgb}{0.000000,0.000000,0.000000}%
\pgfsetstrokecolor{textcolor}%
\pgfsetfillcolor{textcolor}%
\pgftext[x=3.039047in,y=0.608794in,,top]{\color{textcolor}\sffamily\fontsize{14.000000}{16.800000}\selectfont 30}%
\end{pgfscope}%
\begin{pgfscope}%
\pgfsetbuttcap%
\pgfsetroundjoin%
\definecolor{currentfill}{rgb}{0.000000,0.000000,0.000000}%
\pgfsetfillcolor{currentfill}%
\pgfsetlinewidth{0.803000pt}%
\definecolor{currentstroke}{rgb}{0.000000,0.000000,0.000000}%
\pgfsetstrokecolor{currentstroke}%
\pgfsetdash{}{0pt}%
\pgfsys@defobject{currentmarker}{\pgfqpoint{0.000000in}{-0.048611in}}{\pgfqpoint{0.000000in}{0.000000in}}{%
\pgfpathmoveto{\pgfqpoint{0.000000in}{0.000000in}}%
\pgfpathlineto{\pgfqpoint{0.000000in}{-0.048611in}}%
\pgfusepath{stroke,fill}%
}%
\begin{pgfscope}%
\pgfsys@transformshift{3.655054in}{0.706016in}%
\pgfsys@useobject{currentmarker}{}%
\end{pgfscope}%
\end{pgfscope}%
\begin{pgfscope}%
\definecolor{textcolor}{rgb}{0.000000,0.000000,0.000000}%
\pgfsetstrokecolor{textcolor}%
\pgfsetfillcolor{textcolor}%
\pgftext[x=3.655054in,y=0.608794in,,top]{\color{textcolor}\sffamily\fontsize{14.000000}{16.800000}\selectfont 40}%
\end{pgfscope}%
\begin{pgfscope}%
\pgfsetbuttcap%
\pgfsetroundjoin%
\definecolor{currentfill}{rgb}{0.000000,0.000000,0.000000}%
\pgfsetfillcolor{currentfill}%
\pgfsetlinewidth{0.803000pt}%
\definecolor{currentstroke}{rgb}{0.000000,0.000000,0.000000}%
\pgfsetstrokecolor{currentstroke}%
\pgfsetdash{}{0pt}%
\pgfsys@defobject{currentmarker}{\pgfqpoint{0.000000in}{-0.048611in}}{\pgfqpoint{0.000000in}{0.000000in}}{%
\pgfpathmoveto{\pgfqpoint{0.000000in}{0.000000in}}%
\pgfpathlineto{\pgfqpoint{0.000000in}{-0.048611in}}%
\pgfusepath{stroke,fill}%
}%
\begin{pgfscope}%
\pgfsys@transformshift{4.271061in}{0.706016in}%
\pgfsys@useobject{currentmarker}{}%
\end{pgfscope}%
\end{pgfscope}%
\begin{pgfscope}%
\definecolor{textcolor}{rgb}{0.000000,0.000000,0.000000}%
\pgfsetstrokecolor{textcolor}%
\pgfsetfillcolor{textcolor}%
\pgftext[x=4.271061in,y=0.608794in,,top]{\color{textcolor}\sffamily\fontsize{14.000000}{16.800000}\selectfont 50}%
\end{pgfscope}%
\begin{pgfscope}%
\pgfsetbuttcap%
\pgfsetroundjoin%
\definecolor{currentfill}{rgb}{0.000000,0.000000,0.000000}%
\pgfsetfillcolor{currentfill}%
\pgfsetlinewidth{0.803000pt}%
\definecolor{currentstroke}{rgb}{0.000000,0.000000,0.000000}%
\pgfsetstrokecolor{currentstroke}%
\pgfsetdash{}{0pt}%
\pgfsys@defobject{currentmarker}{\pgfqpoint{0.000000in}{-0.048611in}}{\pgfqpoint{0.000000in}{0.000000in}}{%
\pgfpathmoveto{\pgfqpoint{0.000000in}{0.000000in}}%
\pgfpathlineto{\pgfqpoint{0.000000in}{-0.048611in}}%
\pgfusepath{stroke,fill}%
}%
\begin{pgfscope}%
\pgfsys@transformshift{4.887069in}{0.706016in}%
\pgfsys@useobject{currentmarker}{}%
\end{pgfscope}%
\end{pgfscope}%
\begin{pgfscope}%
\definecolor{textcolor}{rgb}{0.000000,0.000000,0.000000}%
\pgfsetstrokecolor{textcolor}%
\pgfsetfillcolor{textcolor}%
\pgftext[x=4.887069in,y=0.608794in,,top]{\color{textcolor}\sffamily\fontsize{14.000000}{16.800000}\selectfont 60}%
\end{pgfscope}%
\begin{pgfscope}%
\pgfsetbuttcap%
\pgfsetroundjoin%
\definecolor{currentfill}{rgb}{0.000000,0.000000,0.000000}%
\pgfsetfillcolor{currentfill}%
\pgfsetlinewidth{0.803000pt}%
\definecolor{currentstroke}{rgb}{0.000000,0.000000,0.000000}%
\pgfsetstrokecolor{currentstroke}%
\pgfsetdash{}{0pt}%
\pgfsys@defobject{currentmarker}{\pgfqpoint{0.000000in}{-0.048611in}}{\pgfqpoint{0.000000in}{0.000000in}}{%
\pgfpathmoveto{\pgfqpoint{0.000000in}{0.000000in}}%
\pgfpathlineto{\pgfqpoint{0.000000in}{-0.048611in}}%
\pgfusepath{stroke,fill}%
}%
\begin{pgfscope}%
\pgfsys@transformshift{5.503076in}{0.706016in}%
\pgfsys@useobject{currentmarker}{}%
\end{pgfscope}%
\end{pgfscope}%
\begin{pgfscope}%
\definecolor{textcolor}{rgb}{0.000000,0.000000,0.000000}%
\pgfsetstrokecolor{textcolor}%
\pgfsetfillcolor{textcolor}%
\pgftext[x=5.503076in,y=0.608794in,,top]{\color{textcolor}\sffamily\fontsize{14.000000}{16.800000}\selectfont 70}%
\end{pgfscope}%
\begin{pgfscope}%
\pgfsetbuttcap%
\pgfsetroundjoin%
\definecolor{currentfill}{rgb}{0.000000,0.000000,0.000000}%
\pgfsetfillcolor{currentfill}%
\pgfsetlinewidth{0.803000pt}%
\definecolor{currentstroke}{rgb}{0.000000,0.000000,0.000000}%
\pgfsetstrokecolor{currentstroke}%
\pgfsetdash{}{0pt}%
\pgfsys@defobject{currentmarker}{\pgfqpoint{0.000000in}{-0.048611in}}{\pgfqpoint{0.000000in}{0.000000in}}{%
\pgfpathmoveto{\pgfqpoint{0.000000in}{0.000000in}}%
\pgfpathlineto{\pgfqpoint{0.000000in}{-0.048611in}}%
\pgfusepath{stroke,fill}%
}%
\begin{pgfscope}%
\pgfsys@transformshift{6.119084in}{0.706016in}%
\pgfsys@useobject{currentmarker}{}%
\end{pgfscope}%
\end{pgfscope}%
\begin{pgfscope}%
\definecolor{textcolor}{rgb}{0.000000,0.000000,0.000000}%
\pgfsetstrokecolor{textcolor}%
\pgfsetfillcolor{textcolor}%
\pgftext[x=6.119084in,y=0.608794in,,top]{\color{textcolor}\sffamily\fontsize{14.000000}{16.800000}\selectfont 80}%
\end{pgfscope}%
\begin{pgfscope}%
\definecolor{textcolor}{rgb}{0.000000,0.000000,0.000000}%
\pgfsetstrokecolor{textcolor}%
\pgfsetfillcolor{textcolor}%
\pgftext[x=3.593453in,y=0.365061in,,top]{\color{textcolor}\sffamily\fontsize{16.000000}{19.200000}\selectfont Batches}%
\end{pgfscope}%
\begin{pgfscope}%
\pgfsetbuttcap%
\pgfsetroundjoin%
\definecolor{currentfill}{rgb}{0.000000,0.000000,0.000000}%
\pgfsetfillcolor{currentfill}%
\pgfsetlinewidth{0.803000pt}%
\definecolor{currentstroke}{rgb}{0.000000,0.000000,0.000000}%
\pgfsetstrokecolor{currentstroke}%
\pgfsetdash{}{0pt}%
\pgfsys@defobject{currentmarker}{\pgfqpoint{-0.048611in}{0.000000in}}{\pgfqpoint{-0.000000in}{0.000000in}}{%
\pgfpathmoveto{\pgfqpoint{-0.000000in}{0.000000in}}%
\pgfpathlineto{\pgfqpoint{-0.048611in}{0.000000in}}%
\pgfusepath{stroke,fill}%
}%
\begin{pgfscope}%
\pgfsys@transformshift{0.950781in}{1.308324in}%
\pgfsys@useobject{currentmarker}{}%
\end{pgfscope}%
\end{pgfscope}%
\begin{pgfscope}%
\definecolor{textcolor}{rgb}{0.000000,0.000000,0.000000}%
\pgfsetstrokecolor{textcolor}%
\pgfsetfillcolor{textcolor}%
\pgftext[x=0.420616in, y=1.234458in, left, base]{\color{textcolor}\sffamily\fontsize{14.000000}{16.800000}\selectfont 0.05}%
\end{pgfscope}%
\begin{pgfscope}%
\pgfsetbuttcap%
\pgfsetroundjoin%
\definecolor{currentfill}{rgb}{0.000000,0.000000,0.000000}%
\pgfsetfillcolor{currentfill}%
\pgfsetlinewidth{0.803000pt}%
\definecolor{currentstroke}{rgb}{0.000000,0.000000,0.000000}%
\pgfsetstrokecolor{currentstroke}%
\pgfsetdash{}{0pt}%
\pgfsys@defobject{currentmarker}{\pgfqpoint{-0.048611in}{0.000000in}}{\pgfqpoint{-0.000000in}{0.000000in}}{%
\pgfpathmoveto{\pgfqpoint{-0.000000in}{0.000000in}}%
\pgfpathlineto{\pgfqpoint{-0.048611in}{0.000000in}}%
\pgfusepath{stroke,fill}%
}%
\begin{pgfscope}%
\pgfsys@transformshift{0.950781in}{1.917885in}%
\pgfsys@useobject{currentmarker}{}%
\end{pgfscope}%
\end{pgfscope}%
\begin{pgfscope}%
\definecolor{textcolor}{rgb}{0.000000,0.000000,0.000000}%
\pgfsetstrokecolor{textcolor}%
\pgfsetfillcolor{textcolor}%
\pgftext[x=0.420616in, y=1.844019in, left, base]{\color{textcolor}\sffamily\fontsize{14.000000}{16.800000}\selectfont 0.10}%
\end{pgfscope}%
\begin{pgfscope}%
\pgfsetbuttcap%
\pgfsetroundjoin%
\definecolor{currentfill}{rgb}{0.000000,0.000000,0.000000}%
\pgfsetfillcolor{currentfill}%
\pgfsetlinewidth{0.803000pt}%
\definecolor{currentstroke}{rgb}{0.000000,0.000000,0.000000}%
\pgfsetstrokecolor{currentstroke}%
\pgfsetdash{}{0pt}%
\pgfsys@defobject{currentmarker}{\pgfqpoint{-0.048611in}{0.000000in}}{\pgfqpoint{-0.000000in}{0.000000in}}{%
\pgfpathmoveto{\pgfqpoint{-0.000000in}{0.000000in}}%
\pgfpathlineto{\pgfqpoint{-0.048611in}{0.000000in}}%
\pgfusepath{stroke,fill}%
}%
\begin{pgfscope}%
\pgfsys@transformshift{0.950781in}{2.527447in}%
\pgfsys@useobject{currentmarker}{}%
\end{pgfscope}%
\end{pgfscope}%
\begin{pgfscope}%
\definecolor{textcolor}{rgb}{0.000000,0.000000,0.000000}%
\pgfsetstrokecolor{textcolor}%
\pgfsetfillcolor{textcolor}%
\pgftext[x=0.420616in, y=2.453580in, left, base]{\color{textcolor}\sffamily\fontsize{14.000000}{16.800000}\selectfont 0.15}%
\end{pgfscope}%
\begin{pgfscope}%
\pgfsetbuttcap%
\pgfsetroundjoin%
\definecolor{currentfill}{rgb}{0.000000,0.000000,0.000000}%
\pgfsetfillcolor{currentfill}%
\pgfsetlinewidth{0.803000pt}%
\definecolor{currentstroke}{rgb}{0.000000,0.000000,0.000000}%
\pgfsetstrokecolor{currentstroke}%
\pgfsetdash{}{0pt}%
\pgfsys@defobject{currentmarker}{\pgfqpoint{-0.048611in}{0.000000in}}{\pgfqpoint{-0.000000in}{0.000000in}}{%
\pgfpathmoveto{\pgfqpoint{-0.000000in}{0.000000in}}%
\pgfpathlineto{\pgfqpoint{-0.048611in}{0.000000in}}%
\pgfusepath{stroke,fill}%
}%
\begin{pgfscope}%
\pgfsys@transformshift{0.950781in}{3.137008in}%
\pgfsys@useobject{currentmarker}{}%
\end{pgfscope}%
\end{pgfscope}%
\begin{pgfscope}%
\definecolor{textcolor}{rgb}{0.000000,0.000000,0.000000}%
\pgfsetstrokecolor{textcolor}%
\pgfsetfillcolor{textcolor}%
\pgftext[x=0.420616in, y=3.063142in, left, base]{\color{textcolor}\sffamily\fontsize{14.000000}{16.800000}\selectfont 0.20}%
\end{pgfscope}%
\begin{pgfscope}%
\pgfsetbuttcap%
\pgfsetroundjoin%
\definecolor{currentfill}{rgb}{0.000000,0.000000,0.000000}%
\pgfsetfillcolor{currentfill}%
\pgfsetlinewidth{0.803000pt}%
\definecolor{currentstroke}{rgb}{0.000000,0.000000,0.000000}%
\pgfsetstrokecolor{currentstroke}%
\pgfsetdash{}{0pt}%
\pgfsys@defobject{currentmarker}{\pgfqpoint{-0.048611in}{0.000000in}}{\pgfqpoint{-0.000000in}{0.000000in}}{%
\pgfpathmoveto{\pgfqpoint{-0.000000in}{0.000000in}}%
\pgfpathlineto{\pgfqpoint{-0.048611in}{0.000000in}}%
\pgfusepath{stroke,fill}%
}%
\begin{pgfscope}%
\pgfsys@transformshift{0.950781in}{3.746569in}%
\pgfsys@useobject{currentmarker}{}%
\end{pgfscope}%
\end{pgfscope}%
\begin{pgfscope}%
\definecolor{textcolor}{rgb}{0.000000,0.000000,0.000000}%
\pgfsetstrokecolor{textcolor}%
\pgfsetfillcolor{textcolor}%
\pgftext[x=0.420616in, y=3.672703in, left, base]{\color{textcolor}\sffamily\fontsize{14.000000}{16.800000}\selectfont 0.25}%
\end{pgfscope}%
\begin{pgfscope}%
\pgfsetbuttcap%
\pgfsetroundjoin%
\definecolor{currentfill}{rgb}{0.000000,0.000000,0.000000}%
\pgfsetfillcolor{currentfill}%
\pgfsetlinewidth{0.803000pt}%
\definecolor{currentstroke}{rgb}{0.000000,0.000000,0.000000}%
\pgfsetstrokecolor{currentstroke}%
\pgfsetdash{}{0pt}%
\pgfsys@defobject{currentmarker}{\pgfqpoint{-0.048611in}{0.000000in}}{\pgfqpoint{-0.000000in}{0.000000in}}{%
\pgfpathmoveto{\pgfqpoint{-0.000000in}{0.000000in}}%
\pgfpathlineto{\pgfqpoint{-0.048611in}{0.000000in}}%
\pgfusepath{stroke,fill}%
}%
\begin{pgfscope}%
\pgfsys@transformshift{0.950781in}{4.356130in}%
\pgfsys@useobject{currentmarker}{}%
\end{pgfscope}%
\end{pgfscope}%
\begin{pgfscope}%
\definecolor{textcolor}{rgb}{0.000000,0.000000,0.000000}%
\pgfsetstrokecolor{textcolor}%
\pgfsetfillcolor{textcolor}%
\pgftext[x=0.420616in, y=4.282264in, left, base]{\color{textcolor}\sffamily\fontsize{14.000000}{16.800000}\selectfont 0.30}%
\end{pgfscope}%
\begin{pgfscope}%
\definecolor{textcolor}{rgb}{0.000000,0.000000,0.000000}%
\pgfsetstrokecolor{textcolor}%
\pgfsetfillcolor{textcolor}%
\pgftext[x=0.365061in,y=2.678008in,,bottom,rotate=90.000000]{\color{textcolor}\sffamily\fontsize{16.000000}{19.200000}\selectfont Deterioration}%
\end{pgfscope}%
\begin{pgfscope}%
\pgfpathrectangle{\pgfqpoint{0.950781in}{0.706016in}}{\pgfqpoint{5.285344in}{3.943984in}}%
\pgfusepath{clip}%
\pgfsetrectcap%
\pgfsetroundjoin%
\pgfsetlinewidth{1.505625pt}%
\definecolor{currentstroke}{rgb}{0.235294,0.701961,0.443137}%
\pgfsetstrokecolor{currentstroke}%
\pgfsetdash{}{0pt}%
\pgfpathmoveto{\pgfqpoint{1.191024in}{0.885288in}}%
\pgfpathlineto{\pgfqpoint{1.252625in}{0.923081in}}%
\pgfpathlineto{\pgfqpoint{1.314226in}{1.040117in}}%
\pgfpathlineto{\pgfqpoint{1.375826in}{1.079129in}}%
\pgfpathlineto{\pgfqpoint{1.437427in}{1.164467in}}%
\pgfpathlineto{\pgfqpoint{1.499028in}{1.269312in}}%
\pgfpathlineto{\pgfqpoint{1.560629in}{1.374157in}}%
\pgfpathlineto{\pgfqpoint{1.622229in}{1.468029in}}%
\pgfpathlineto{\pgfqpoint{1.683830in}{1.544834in}}%
\pgfpathlineto{\pgfqpoint{1.745431in}{1.650897in}}%
\pgfpathlineto{\pgfqpoint{1.807032in}{1.739893in}}%
\pgfpathlineto{\pgfqpoint{1.868632in}{1.813041in}}%
\pgfpathlineto{\pgfqpoint{1.930233in}{1.902037in}}%
\pgfpathlineto{\pgfqpoint{1.991834in}{1.977622in}}%
\pgfpathlineto{\pgfqpoint{2.053435in}{2.055646in}}%
\pgfpathlineto{\pgfqpoint{2.115035in}{2.128793in}}%
\pgfpathlineto{\pgfqpoint{2.176636in}{2.198283in}}%
\pgfpathlineto{\pgfqpoint{2.238237in}{2.258020in}}%
\pgfpathlineto{\pgfqpoint{2.299838in}{2.309224in}}%
\pgfpathlineto{\pgfqpoint{2.361438in}{2.357989in}}%
\pgfpathlineto{\pgfqpoint{2.423039in}{2.412849in}}%
\pgfpathlineto{\pgfqpoint{2.484640in}{2.455518in}}%
\pgfpathlineto{\pgfqpoint{2.546241in}{2.505502in}}%
\pgfpathlineto{\pgfqpoint{2.607841in}{2.554267in}}%
\pgfpathlineto{\pgfqpoint{2.669442in}{2.595717in}}%
\pgfpathlineto{\pgfqpoint{2.731043in}{2.644482in}}%
\pgfpathlineto{\pgfqpoint{2.792644in}{2.696905in}}%
\pgfpathlineto{\pgfqpoint{2.854244in}{2.731040in}}%
\pgfpathlineto{\pgfqpoint{2.915845in}{2.770052in}}%
\pgfpathlineto{\pgfqpoint{2.977446in}{2.811502in}}%
\pgfpathlineto{\pgfqpoint{3.039047in}{2.854171in}}%
\pgfpathlineto{\pgfqpoint{3.100647in}{2.902936in}}%
\pgfpathlineto{\pgfqpoint{3.162248in}{2.941948in}}%
\pgfpathlineto{\pgfqpoint{3.223849in}{2.984618in}}%
\pgfpathlineto{\pgfqpoint{3.285449in}{3.015096in}}%
\pgfpathlineto{\pgfqpoint{3.347050in}{3.051669in}}%
\pgfpathlineto{\pgfqpoint{3.408651in}{3.093119in}}%
\pgfpathlineto{\pgfqpoint{3.470252in}{3.128474in}}%
\pgfpathlineto{\pgfqpoint{3.531852in}{3.160171in}}%
\pgfpathlineto{\pgfqpoint{3.593453in}{3.200402in}}%
\pgfpathlineto{\pgfqpoint{3.655054in}{3.235757in}}%
\pgfpathlineto{\pgfqpoint{3.716655in}{3.273550in}}%
\pgfpathlineto{\pgfqpoint{3.778255in}{3.308904in}}%
\pgfpathlineto{\pgfqpoint{3.839856in}{3.350354in}}%
\pgfpathlineto{\pgfqpoint{3.901457in}{3.386928in}}%
\pgfpathlineto{\pgfqpoint{3.963058in}{3.425940in}}%
\pgfpathlineto{\pgfqpoint{4.024658in}{3.474705in}}%
\pgfpathlineto{\pgfqpoint{4.086259in}{3.516155in}}%
\pgfpathlineto{\pgfqpoint{4.147860in}{3.558824in}}%
\pgfpathlineto{\pgfqpoint{4.209461in}{3.597836in}}%
\pgfpathlineto{\pgfqpoint{4.271061in}{3.641725in}}%
\pgfpathlineto{\pgfqpoint{4.332662in}{3.679517in}}%
\pgfpathlineto{\pgfqpoint{4.394263in}{3.717310in}}%
\pgfpathlineto{\pgfqpoint{4.455864in}{3.752665in}}%
\pgfpathlineto{\pgfqpoint{4.517464in}{3.794115in}}%
\pgfpathlineto{\pgfqpoint{4.579065in}{3.838003in}}%
\pgfpathlineto{\pgfqpoint{4.640666in}{3.873358in}}%
\pgfpathlineto{\pgfqpoint{4.702267in}{3.913589in}}%
\pgfpathlineto{\pgfqpoint{4.763867in}{3.950163in}}%
\pgfpathlineto{\pgfqpoint{4.825468in}{3.986736in}}%
\pgfpathlineto{\pgfqpoint{4.887069in}{4.020872in}}%
\pgfpathlineto{\pgfqpoint{4.948670in}{4.050131in}}%
\pgfpathlineto{\pgfqpoint{5.010270in}{4.080609in}}%
\pgfpathlineto{\pgfqpoint{5.071871in}{4.109868in}}%
\pgfpathlineto{\pgfqpoint{5.133472in}{4.140346in}}%
\pgfpathlineto{\pgfqpoint{5.195073in}{4.169605in}}%
\pgfpathlineto{\pgfqpoint{5.256673in}{4.196425in}}%
\pgfpathlineto{\pgfqpoint{5.318274in}{4.220808in}}%
\pgfpathlineto{\pgfqpoint{5.379875in}{4.248848in}}%
\pgfpathlineto{\pgfqpoint{5.441476in}{4.274449in}}%
\pgfpathlineto{\pgfqpoint{5.503076in}{4.300051in}}%
\pgfpathlineto{\pgfqpoint{5.564677in}{4.320776in}}%
\pgfpathlineto{\pgfqpoint{5.626278in}{4.345158in}}%
\pgfpathlineto{\pgfqpoint{5.687879in}{4.370760in}}%
\pgfpathlineto{\pgfqpoint{5.749479in}{4.392704in}}%
\pgfpathlineto{\pgfqpoint{5.811080in}{4.420744in}}%
\pgfpathlineto{\pgfqpoint{5.872681in}{4.445126in}}%
\pgfpathlineto{\pgfqpoint{5.934282in}{4.467071in}}%
\pgfpathlineto{\pgfqpoint{5.995882in}{4.470728in}}%
\pgfusepath{stroke}%
\end{pgfscope}%
\begin{pgfscope}%
\pgfpathrectangle{\pgfqpoint{0.950781in}{0.706016in}}{\pgfqpoint{5.285344in}{3.943984in}}%
\pgfusepath{clip}%
\pgfsetrectcap%
\pgfsetroundjoin%
\pgfsetlinewidth{1.505625pt}%
\definecolor{currentstroke}{rgb}{1.000000,0.549020,0.000000}%
\pgfsetstrokecolor{currentstroke}%
\pgfsetdash{}{0pt}%
\pgfpathmoveto{\pgfqpoint{1.191024in}{0.934053in}}%
\pgfpathlineto{\pgfqpoint{1.252625in}{0.971846in}}%
\pgfpathlineto{\pgfqpoint{1.314226in}{1.049870in}}%
\pgfpathlineto{\pgfqpoint{1.375826in}{1.052308in}}%
\pgfpathlineto{\pgfqpoint{1.437427in}{1.096197in}}%
\pgfpathlineto{\pgfqpoint{1.499028in}{1.146181in}}%
\pgfpathlineto{\pgfqpoint{1.560629in}{1.190069in}}%
\pgfpathlineto{\pgfqpoint{1.622229in}{1.219328in}}%
\pgfpathlineto{\pgfqpoint{1.683830in}{1.244930in}}%
\pgfpathlineto{\pgfqpoint{1.745431in}{1.270531in}}%
\pgfpathlineto{\pgfqpoint{1.807032in}{1.290037in}}%
\pgfpathlineto{\pgfqpoint{1.868632in}{1.302228in}}%
\pgfpathlineto{\pgfqpoint{1.930233in}{1.322953in}}%
\pgfpathlineto{\pgfqpoint{1.991834in}{1.329049in}}%
\pgfpathlineto{\pgfqpoint{2.053435in}{1.335145in}}%
\pgfpathlineto{\pgfqpoint{2.115035in}{1.358308in}}%
\pgfpathlineto{\pgfqpoint{2.176636in}{1.371718in}}%
\pgfpathlineto{\pgfqpoint{2.238237in}{1.387567in}}%
\pgfpathlineto{\pgfqpoint{2.299838in}{1.393663in}}%
\pgfpathlineto{\pgfqpoint{2.361438in}{1.399758in}}%
\pgfpathlineto{\pgfqpoint{2.423039in}{1.409511in}}%
\pgfpathlineto{\pgfqpoint{2.484640in}{1.421702in}}%
\pgfpathlineto{\pgfqpoint{2.546241in}{1.431455in}}%
\pgfpathlineto{\pgfqpoint{2.607841in}{1.432674in}}%
\pgfpathlineto{\pgfqpoint{2.669442in}{1.435113in}}%
\pgfpathlineto{\pgfqpoint{2.731043in}{1.439989in}}%
\pgfpathlineto{\pgfqpoint{2.792644in}{1.450596in}}%
\pgfpathlineto{\pgfqpoint{2.854244in}{1.458276in}}%
\pgfpathlineto{\pgfqpoint{2.915845in}{1.461933in}}%
\pgfpathlineto{\pgfqpoint{2.977446in}{1.465591in}}%
\pgfpathlineto{\pgfqpoint{3.039047in}{1.471321in}}%
\pgfpathlineto{\pgfqpoint{3.100647in}{1.476563in}}%
\pgfpathlineto{\pgfqpoint{3.162248in}{1.482658in}}%
\pgfpathlineto{\pgfqpoint{3.223849in}{1.487535in}}%
\pgfpathlineto{\pgfqpoint{3.285449in}{1.499726in}}%
\pgfpathlineto{\pgfqpoint{3.347050in}{1.502164in}}%
\pgfpathlineto{\pgfqpoint{3.408651in}{1.505822in}}%
\pgfpathlineto{\pgfqpoint{3.470252in}{1.513137in}}%
\pgfpathlineto{\pgfqpoint{3.531852in}{1.514356in}}%
\pgfpathlineto{\pgfqpoint{3.593453in}{1.518013in}}%
\pgfpathlineto{\pgfqpoint{3.655054in}{1.524109in}}%
\pgfpathlineto{\pgfqpoint{3.716655in}{1.526547in}}%
\pgfpathlineto{\pgfqpoint{3.778255in}{1.528985in}}%
\pgfpathlineto{\pgfqpoint{3.839856in}{1.532642in}}%
\pgfpathlineto{\pgfqpoint{3.901457in}{1.536300in}}%
\pgfpathlineto{\pgfqpoint{3.963058in}{1.538738in}}%
\pgfpathlineto{\pgfqpoint{4.024658in}{1.541176in}}%
\pgfpathlineto{\pgfqpoint{4.086259in}{1.543615in}}%
\pgfpathlineto{\pgfqpoint{4.147860in}{1.547272in}}%
\pgfpathlineto{\pgfqpoint{4.209461in}{1.547272in}}%
\pgfpathlineto{\pgfqpoint{4.271061in}{1.550929in}}%
\pgfpathlineto{\pgfqpoint{4.332662in}{1.552148in}}%
\pgfpathlineto{\pgfqpoint{4.394263in}{1.552148in}}%
\pgfpathlineto{\pgfqpoint{4.455864in}{1.555806in}}%
\pgfpathlineto{\pgfqpoint{4.517464in}{1.557025in}}%
\pgfpathlineto{\pgfqpoint{4.579065in}{1.558244in}}%
\pgfpathlineto{\pgfqpoint{4.640666in}{1.559463in}}%
\pgfpathlineto{\pgfqpoint{4.702267in}{1.559463in}}%
\pgfpathlineto{\pgfqpoint{4.763867in}{1.561901in}}%
\pgfpathlineto{\pgfqpoint{4.825468in}{1.564340in}}%
\pgfpathlineto{\pgfqpoint{4.887069in}{1.564340in}}%
\pgfpathlineto{\pgfqpoint{4.948670in}{1.566778in}}%
\pgfpathlineto{\pgfqpoint{5.010270in}{1.567997in}}%
\pgfpathlineto{\pgfqpoint{5.071871in}{1.570435in}}%
\pgfpathlineto{\pgfqpoint{5.133472in}{1.571654in}}%
\pgfpathlineto{\pgfqpoint{5.195073in}{1.571654in}}%
\pgfpathlineto{\pgfqpoint{5.256673in}{1.570435in}}%
\pgfpathlineto{\pgfqpoint{5.318274in}{1.570435in}}%
\pgfpathlineto{\pgfqpoint{5.379875in}{1.571654in}}%
\pgfpathlineto{\pgfqpoint{5.441476in}{1.571654in}}%
\pgfpathlineto{\pgfqpoint{5.503076in}{1.571654in}}%
\pgfpathlineto{\pgfqpoint{5.564677in}{1.571654in}}%
\pgfpathlineto{\pgfqpoint{5.626278in}{1.569216in}}%
\pgfpathlineto{\pgfqpoint{5.687879in}{1.571654in}}%
\pgfpathlineto{\pgfqpoint{5.749479in}{1.571654in}}%
\pgfpathlineto{\pgfqpoint{5.811080in}{1.572874in}}%
\pgfpathlineto{\pgfqpoint{5.872681in}{1.572874in}}%
\pgfpathlineto{\pgfqpoint{5.934282in}{1.574093in}}%
\pgfpathlineto{\pgfqpoint{5.995882in}{1.574093in}}%
\pgfusepath{stroke}%
\end{pgfscope}%
\begin{pgfscope}%
\pgfsetrectcap%
\pgfsetmiterjoin%
\pgfsetlinewidth{0.803000pt}%
\definecolor{currentstroke}{rgb}{0.000000,0.000000,0.000000}%
\pgfsetstrokecolor{currentstroke}%
\pgfsetdash{}{0pt}%
\pgfpathmoveto{\pgfqpoint{0.950781in}{0.706016in}}%
\pgfpathlineto{\pgfqpoint{0.950781in}{4.650000in}}%
\pgfusepath{stroke}%
\end{pgfscope}%
\begin{pgfscope}%
\pgfsetrectcap%
\pgfsetmiterjoin%
\pgfsetlinewidth{0.803000pt}%
\definecolor{currentstroke}{rgb}{0.000000,0.000000,0.000000}%
\pgfsetstrokecolor{currentstroke}%
\pgfsetdash{}{0pt}%
\pgfpathmoveto{\pgfqpoint{6.236125in}{0.706016in}}%
\pgfpathlineto{\pgfqpoint{6.236125in}{4.650000in}}%
\pgfusepath{stroke}%
\end{pgfscope}%
\begin{pgfscope}%
\pgfsetrectcap%
\pgfsetmiterjoin%
\pgfsetlinewidth{0.803000pt}%
\definecolor{currentstroke}{rgb}{0.000000,0.000000,0.000000}%
\pgfsetstrokecolor{currentstroke}%
\pgfsetdash{}{0pt}%
\pgfpathmoveto{\pgfqpoint{0.950781in}{0.706016in}}%
\pgfpathlineto{\pgfqpoint{6.236125in}{0.706016in}}%
\pgfusepath{stroke}%
\end{pgfscope}%
\begin{pgfscope}%
\pgfsetrectcap%
\pgfsetmiterjoin%
\pgfsetlinewidth{0.803000pt}%
\definecolor{currentstroke}{rgb}{0.000000,0.000000,0.000000}%
\pgfsetstrokecolor{currentstroke}%
\pgfsetdash{}{0pt}%
\pgfpathmoveto{\pgfqpoint{0.950781in}{4.650000in}}%
\pgfpathlineto{\pgfqpoint{6.236125in}{4.650000in}}%
\pgfusepath{stroke}%
\end{pgfscope}%
\begin{pgfscope}%
\pgfsetbuttcap%
\pgfsetmiterjoin%
\definecolor{currentfill}{rgb}{1.000000,1.000000,1.000000}%
\pgfsetfillcolor{currentfill}%
\pgfsetfillopacity{0.800000}%
\pgfsetlinewidth{1.003750pt}%
\definecolor{currentstroke}{rgb}{0.800000,0.800000,0.800000}%
\pgfsetstrokecolor{currentstroke}%
\pgfsetstrokeopacity{0.800000}%
\pgfsetdash{}{0pt}%
\pgfpathmoveto{\pgfqpoint{4.488599in}{3.819879in}}%
\pgfpathlineto{\pgfqpoint{6.080570in}{3.819879in}}%
\pgfpathquadraticcurveto{\pgfqpoint{6.125014in}{3.819879in}}{\pgfqpoint{6.125014in}{3.864323in}}%
\pgfpathlineto{\pgfqpoint{6.125014in}{4.494444in}}%
\pgfpathquadraticcurveto{\pgfqpoint{6.125014in}{4.538889in}}{\pgfqpoint{6.080570in}{4.538889in}}%
\pgfpathlineto{\pgfqpoint{4.488599in}{4.538889in}}%
\pgfpathquadraticcurveto{\pgfqpoint{4.444155in}{4.538889in}}{\pgfqpoint{4.444155in}{4.494444in}}%
\pgfpathlineto{\pgfqpoint{4.444155in}{3.864323in}}%
\pgfpathquadraticcurveto{\pgfqpoint{4.444155in}{3.819879in}}{\pgfqpoint{4.488599in}{3.819879in}}%
\pgfpathlineto{\pgfqpoint{4.488599in}{3.819879in}}%
\pgfpathclose%
\pgfusepath{stroke,fill}%
\end{pgfscope}%
\begin{pgfscope}%
\pgfsetrectcap%
\pgfsetroundjoin%
\pgfsetlinewidth{1.505625pt}%
\definecolor{currentstroke}{rgb}{0.235294,0.701961,0.443137}%
\pgfsetstrokecolor{currentstroke}%
\pgfsetdash{}{0pt}%
\pgfpathmoveto{\pgfqpoint{4.533044in}{4.358941in}}%
\pgfpathlineto{\pgfqpoint{4.755266in}{4.358941in}}%
\pgfpathlineto{\pgfqpoint{4.977488in}{4.358941in}}%
\pgfusepath{stroke}%
\end{pgfscope}%
\begin{pgfscope}%
\definecolor{textcolor}{rgb}{0.000000,0.000000,0.000000}%
\pgfsetstrokecolor{textcolor}%
\pgfsetfillcolor{textcolor}%
\pgftext[x=5.155266in,y=4.281163in,left,base]{\color{textcolor}\sffamily\fontsize{16.000000}{19.200000}\selectfont TENT}%
\end{pgfscope}%
\begin{pgfscope}%
\pgfsetrectcap%
\pgfsetroundjoin%
\pgfsetlinewidth{1.505625pt}%
\definecolor{currentstroke}{rgb}{1.000000,0.549020,0.000000}%
\pgfsetstrokecolor{currentstroke}%
\pgfsetdash{}{0pt}%
\pgfpathmoveto{\pgfqpoint{4.533044in}{4.032769in}}%
\pgfpathlineto{\pgfqpoint{4.755266in}{4.032769in}}%
\pgfpathlineto{\pgfqpoint{4.977488in}{4.032769in}}%
\pgfusepath{stroke}%
\end{pgfscope}%
\begin{pgfscope}%
\definecolor{textcolor}{rgb}{0.000000,0.000000,0.000000}%
\pgfsetstrokecolor{textcolor}%
\pgfsetfillcolor{textcolor}%
\pgftext[x=5.155266in,y=3.954992in,left,base]{\color{textcolor}\sffamily\fontsize{16.000000}{19.200000}\selectfont CLIPTTA}%
\end{pgfscope}%
\end{pgfpicture}%
\makeatother%
\endgroup%

%% file: main.bbl
\begin{thebibliography}{10}

\bibitem{clip}
Alec Radford, Jong~Wook Kim, Chris Hallacy, Aditya Ramesh, Gabriel Goh, Sandhini Agarwal, Girish Sastry, Amanda Askell, Pamela Mishkin, Jack Clark, et~al.
\newblock Learning transferable visual models from natural language supervision.
\newblock In {\em International conference on machine learning}, pages 8748--8763. PMLR, 2021.

\bibitem{align}
Chao Jia, Yinfei Yang, Ye~Xia, Yi{-}Ting Chen, Zarana Parekh, Hieu Pham, Quoc~V. Le, Yun{-}Hsuan Sung, Zhen Li, and Tom Duerig.
\newblock Scaling up visual and vision-language representation learning with noisy text supervision.
\newblock In {\em Proceedings of the 38th International Conference on Machine Learning, {ICML} 2021, 18-24 July 2021, Virtual Event}, pages 4904--4916. {PMLR}, 2021.

\bibitem{tpt}
Manli Shu, Weili Nie, De-An Huang, Zhiding Yu, Tom Goldstein, Anima Anandkumar, and Chaowei Xiao.
\newblock Test-time prompt tuning for zero-shot generalization in vision-language models.
\newblock In S.~Koyejo, S.~Mohamed, A.~Agarwal, D.~Belgrave, K.~Cho, and A.~Oh, editors, {\em Advances in Neural Information Processing Systems}, volume~35, pages 14274--14289. Curran Associates, Inc., 2022.

\bibitem{tda}
Adilbek Karmanov, Dayan Guan, Shijian Lu, Abdulmotaleb El~Saddik, and Eric Xing.
\newblock Efficient test-time adaptation of vision-language models.
\newblock In {\em Proceedings of the IEEE/CVF Conference on Computer Vision and Pattern Recognition}, pages 14162--14171, 2024.

\bibitem{calip}
Ziyu Guo, Renrui Zhang, Longtian Qiu, Xianzheng Ma, Xupeng Miao, Xuming He, and Bin Cui.
\newblock Calip: zero-shot enhancement of clip with parameter-free attention.
\newblock In {\em Proceedings of the Thirty-Seventh AAAI Conference on Artificial Intelligence and Thirty-Fifth Conference on Innovative Applications of Artificial Intelligence and Thirteenth Symposium on Educational Advances in Artificial Intelligence}, AAAI'23/IAAI'23/EAAI'23. AAAI Press, 2023.

\bibitem{clipartt}
Gustavo~Adolfo {Vargas Hakim}, David Osowiechi, Mehrdad Noori, Milad Cheraghalikhani, Ali Bahri, Moslem Yazdanpanah, Ismail~Ben Ayed, and Christian Desrosiers.
\newblock Clipartt: Light-weight adaptation of clip to new domains at test time, 2024.

\bibitem{watt}
David Osowiechi, Mehrdad Noori, Gustavo~Adolfo Vargas~Hakim, Moslem Yazdanpanah, Ali Bahri, Milad Cheraghalikhani, Sahar Dastani, Farzad Beizaee, Ismail~Ben Ayed, and Christian Desrosiers.
\newblock Watt: Weight average test-time adaption of clip.
\newblock {\em arXiv:2406.13875}, 2024.

\bibitem{onthefly}
Jeya Maria~Jose Valanarasu, Pengfei Guo, Vibashan VS, and Vishal~M. Patel.
\newblock On-the-fly test-time adaptation for medical image segmentation.
\newblock In Ipek Oguz, Jack Noble, Xiaoxiao Li, Martin Styner, Christian Baumgartner, Mirabela Rusu, Tobias Heinmann, Despina Kontos, Bennett Landman, and Benoit Dawant, editors, {\em Medical Imaging with Deep Learning}, volume 227 of {\em Proceedings of Machine Learning Research}, pages 586--598. PMLR, 10--12 Jul 2024.

\bibitem{posetta}
Qiongjie Cui, Huaijiang Sun, Jianfeng Lu, Bin Li, and Weiqing Li.
\newblock Meta-auxiliary learning for adaptive human pose prediction.
\newblock {\em Proceedings of the AAAI Conference on Artificial Intelligence}, 37(5):6166--6174, Jun. 2023.

\bibitem{federatedtta}
Wenxuan Bao, Tianxin Wei, Haohan Wang, and Jingrui He.
\newblock Adaptive test-time personalization for federated learning.
\newblock In A.~Oh, T.~Naumann, A.~Globerson, K.~Saenko, M.~Hardt, and S.~Levine, editors, {\em Advances in Neural Information Processing Systems}, volume~36, pages 77882--77914. Curran Associates, Inc., 2023.

\bibitem{tent}
Dequan Wang, Evan Shelhamer, Shaoteng Liu, Bruno~A. Olshausen, and Trevor Darrell.
\newblock Tent: Fully test-time adaptation by entropy minimization.
\newblock In {\em 9th International Conference on Learning Representations, {ICLR} 2021, Virtual Event, Austria, May 3-7, 2021}. OpenReview.net, 2021.

\bibitem{eta}
Shuaicheng Niu, Jiaxiang Wu, Yifan Zhang, Yaofo Chen, Shijian Zheng, Peilin Zhao, and Mingkui Tan.
\newblock Efficient test-time model adaptation without forgetting.
\newblock In Kamalika Chaudhuri, Stefanie Jegelka, Le~Song, Csaba Szepesvari, Gang Niu, and Sivan Sabato, editors, {\em Proceedings of the 39th International Conference on Machine Learning}, volume 162 of {\em Proceedings of Machine Learning Research}, pages 16888--16905. PMLR, 17--23 Jul 2022.

\bibitem{sar}
Shuaicheng Niu, Jiaxiang Wu, Yifan Zhang, Zhiquan Wen, Yaofo Chen, Peilin Zhao, and Mingkui Tan.
\newblock Towards stable test-time adaptation in dynamic wild world.
\newblock In {\em Internetional Conference on Learning Representations}, 2023.

\bibitem{rotta}
Longhui Yuan, Binhui Xie, and Shuang Li.
\newblock Robust test-time adaptation in dynamic scenarios.
\newblock In {\em Proceedings of the IEEE/CVF Conference on Computer Vision and Pattern Recognition}, pages 15922--15932, 2023.

\bibitem{flyp}
Sachin Goyal, Ananya Kumar, Sankalp Garg, Zico Kolter, and Aditi Raghunathan.
\newblock Finetune like you pretrain: Improved finetuning of zero-shot vision models.
\newblock In {\em Proceedings of the IEEE/CVF Conference on Computer Vision and Pattern Recognition}, pages 19338--19347, 2023.

\bibitem{survey_collapse}
Zixin Wang, Yadan Luo, Liang Zheng, Zhuoxiao Chen, Sen Wang, and Zi~Huang.
\newblock In search of lost online test-time adaptation: A survey.
\newblock {\em International Journal of Computer Vision}, pages 1--34, 2024.

\bibitem{ostta}
Jungsoo Lee, Debasmit Das, Jaegul Choo, and Sungha Choi.
\newblock Towards open-set test-time adaptation utilizing the wisdom of crowds in entropy minimization.
\newblock In {\em Proceedings of the IEEE/CVF International Conference on Computer Vision (ICCV)}, pages 16380--16389, October 2023.

\bibitem{realm}
Skyler Seto, Barry-John Theobald, Federico Danieli, Navdeep Jaitly, and Dan Busbridge.
\newblock Realm: Robust entropy adaptive loss minimization for improved single-sample test-time adaptation.
\newblock In {\em Proceedings of the IEEE/CVF Winter Conference on Applications of Computer Vision (WACV)}, pages 2062--2071, January 2024.

\bibitem{memo}
Marvin Zhang, Sergey Levine, and Chelsea Finn.
\newblock Memo: Test time robustness via adaptation and augmentation.
\newblock In S.~Koyejo, S.~Mohamed, A.~Agarwal, D.~Belgrave, K.~Cho, and A.~Oh, editors, {\em Advances in Neural Information Processing Systems}, volume~35, pages 38629--38642. Curran Associates, Inc., 2022.

\bibitem{conjugate}
Sachin Goyal, Mingjie Sun, Aditi Raghunanthan, and Zico Kolter.
\newblock Test-time adaptation via conjugate pseudo-labels.
\newblock {\em Advances in Neural Information Processing Systems}, 2022.

\bibitem{ttc}
Guoliang Lin, Hanjiang Lai, Yan Pan, and Jian Yin.
\newblock Improving entropy-based test-time adaptation from a clustering view.
\newblock {\em arXiv:2310.20327}, 2023.

\bibitem{adacontrast}
Dian Chen, Dequan Wang, Trevor Darrell, and Sayna Ebrahimi.
\newblock Contrastive test-time adaptation.
\newblock In {\em Proceedings of the IEEE/CVF Conference on Computer Vision and Pattern Recognition}, pages 295--305, 2022.

\bibitem{moco}
Kaiming He, Haoqi Fan, Yuxin Wu, Saining Xie, and Ross Girshick.
\newblock Momentum contrast for unsupervised visual representation learning.
\newblock In {\em Proceedings of the IEEE/CVF conference on computer vision and pattern recognition}, pages 9729--9738, 2020.

\bibitem{prompt1}
Brian Lester, Rami Al-Rfou, and Noah Constant.
\newblock The power of scale for parameter-efficient prompt tuning.
\newblock {\em arXiv:2104.08691}, 2021.

\bibitem{tip}
Renrui Zhang, Wei Zhang, Rongyao Fang, Peng Gao, Kunchang Li, Jifeng Dai, Yu~Qiao, and Hongsheng Li.
\newblock Tip-adapter: Training-free adaption of clip for few-shot classification.
\newblock In {\em European conference on computer vision}, pages 493--510. Springer, 2022.

\bibitem{sotta}
Taesik Gong, Yewon Kim, Taeckyung Lee, Sorn Chottananurak, and Sung-Ju Lee.
\newblock Sotta: Robust test-time adaptation on noisy data streams.
\newblock {\em Advances in Neural Information Processing Systems}, 36, 2024.

\bibitem{stamp}
Yongcan Yu, Lijun Sheng, Ran He, and Jian Liang.
\newblock Stamp: Outlier-aware test-time adaptation with stable memory replay.
\newblock {\em arXiv:2407.15773}, 2024.

\bibitem{unient}
Zhengqing Gao, Xu-Yao Zhang, and Cheng-Lin Liu.
\newblock Unified entropy optimization for open-set test-time adaptation.
\newblock In {\em Proceedings of the IEEE/CVF Conference on Computer Vision and Pattern Recognition (CVPR)}, pages 23975--23984, June 2024.

\bibitem{SWR}
Sungha Choi, Seunghan Yang, Seokeon Choi, and Sungrack Yun.
\newblock Improving test-time adaptation via shift-agnostic weight regularization and nearest source prototypes.
\newblock In Shai Avidan, Gabriel~J. Brostow, Moustapha Ciss{\'{e}}, Giovanni~Maria Farinella, and Tal Hassner, editors, {\em Computer Vision - {ECCV} 2022 - 17th European Conference, Tel Aviv, Israel, October 23-27, 2022, Proceedings, Part {XXXIII}}, volume 13693 of {\em Lecture Notes in Computer Science}, pages 440--458. Springer, 2022.

\bibitem{mcm}
Yifei Ming, Ziyang Cai, Jiuxiang Gu, Yiyou Sun, Wei Li, and Yixuan Li.
\newblock Delving into out-of-distribution detection with vision-language representations.
\newblock In S.~Koyejo, S.~Mohamed, A.~Agarwal, D.~Belgrave, K.~Cho, and A.~Oh, editors, {\em Advances in Neural Information Processing Systems}, volume~35, pages 35087--35102. Curran Associates, Inc., 2022.

\bibitem{cifar10}
Alex Krizhevsky, Geoffrey Hinton, et~al.
\newblock Learning multiple layers of features from tiny images.
\newblock Technical report, Toronto, ON, Canada, 2009.

\bibitem{imagenet}
Jia Deng, Wei Dong, Richard Socher, Li-Jia Li, Kai Li, and Li~Fei-Fei.
\newblock Imagenet: A large-scale hierarchical image database.
\newblock In {\em 2009 IEEE Conference on Computer Vision and Pattern Recognition}, pages 248--255, 2009.

\bibitem{cifar10c}
Dan Hendrycks and Thomas~G. Dietterich.
\newblock Benchmarking neural network robustness to common corruptions and perturbations.
\newblock In {\em 7th International Conference on Learning Representations, {ICLR} 2019, New Orleans, LA, USA, May 6-9, 2019}. OpenReview.net, 2019.

\bibitem{aircraft}
Subhransu Maji, Esa Rahtu, Juho Kannala, Matthew Blaschko, and Andrea Vedaldi.
\newblock Fine-grained visual classification of aircraft.
\newblock {\em arXiv:1306.5151}, 2013.

\bibitem{caltech}
Li~Fei-Fei, R.~Fergus, and P.~Perona.
\newblock Learning generative visual models from few training examples: An incremental bayesian approach tested on 101 object categories.
\newblock In {\em 2004 Conference on Computer Vision and Pattern Recognition Workshop}, pages 178--178, 2004.

\bibitem{cars}
Jonathan Krause, Michael Stark, Jia Deng, and Li~Fei-Fei.
\newblock 3d object representations for fine-grained categorization.
\newblock In {\em 2013 IEEE International Conference on Computer Vision Workshops}, pages 554--561, 2013.

\bibitem{dtd}
Mircea Cimpoi, Subhransu Maji, Iasonas Kokkinos, Sammy Mohamed, and Andrea Vedaldi.
\newblock Describing textures in the wild.
\newblock In {\em 2014 IEEE Conference on Computer Vision and Pattern Recognition}, pages 3606--3613, 2014.

\bibitem{eurosat}
Patrick Helber, Benjamin Bischke, Andreas Dengel, and Damian Borth.
\newblock Eurosat: A novel dataset and deep learning benchmark for land use and land cover classification.
\newblock {\em IEEE Journal of Selected Topics in Applied Earth Observations and Remote Sensing}, 12(7):2217--2226, 2019.

\bibitem{flowers}
Maria-Elena Nilsback and Andrew Zisserman.
\newblock Automated flower classification over a large number of classes.
\newblock In {\em 2008 Sixth Indian Conference on Computer Vision, Graphics \& Image Processing}, pages 722--729, 2008.

\bibitem{food}
Lukas Bossard, Matthieu Guillaumin, and Luc Van~Gool.
\newblock Food-101 -- mining discriminative components with random forests.
\newblock In {\em European Conference on Computer Vision}, 2014.

\bibitem{pets}
Omkar~M Parkhi, Andrea Vedaldi, Andrew Zisserman, and C.~V. Jawahar.
\newblock Cats and dogs.
\newblock In {\em 2012 IEEE Conference on Computer Vision and Pattern Recognition}, pages 3498--3505, 2012.

\bibitem{sun}
Jianxiong Xiao, James Hays, Krista~A. Ehinger, Aude Oliva, and Antonio Torralba.
\newblock Sun database: Large-scale scene recognition from abbey to zoo.
\newblock In {\em 2010 IEEE Computer Society Conference on Computer Vision and Pattern Recognition}, pages 3485--3492, 2010.

\bibitem{ucf}
K~Soomro.
\newblock Ucf101: A dataset of 101 human actions classes from videos in the wild.
\newblock {\em arXiv:1212.0402}, 2012.

\bibitem{visda}
Xingchao Peng, Ben Usman, Neela Kaushik, Dequan Wang, Judy Hoffman, and Kate Saenko.
\newblock Visda: A synthetic-to-real benchmark for visual domain adaptation.
\newblock In {\em Proceedings of the IEEE Conference on Computer Vision and Pattern Recognition (CVPR) Workshops}, June 2018.

\bibitem{mscoco}
Tsung-Yi Lin, Michael Maire, Serge Belongie, James Hays, Pietro Perona, Deva Ramanan, Piotr Doll{\'a}r, and C~Lawrence Zitnick.
\newblock Microsoft coco: Common objects in context.
\newblock In {\em Computer Vision--ECCV 2014: 13th European Conference, Zurich, Switzerland, September 6-12, 2014, Proceedings, Part V 13}, pages 740--755. Springer, 2014.

\bibitem{pacs}
Da~Li, Yongxin Yang, Yi-Zhe Song, and Timothy~M Hospedales.
\newblock Deeper, broader and artier domain generalization.
\newblock In {\em Proceedings of the IEEE international conference on computer vision}, pages 5542--5550, 2017.

\bibitem{officehome}
Hemanth Venkateswara, Jose Eusebio, Shayok Chakraborty, and Sethuraman Panchanathan.
\newblock Deep hashing network for unsupervised domain adaptation.
\newblock In {\em Proceedings of the IEEE Conference on Computer Vision and Pattern Recognition}, pages 5018--5027, 2017.

\bibitem{imagenetv2}
Benjamin Recht, Rebecca Roelofs, Ludwig Schmidt, and Vaishaal Shankar.
\newblock Do imagenet classifiers generalize to imagenet?
\newblock In {\em International conference on machine learning}, pages 5389--5400. PMLR, 2019.

\bibitem{imagenetr}
Dan Hendrycks, Steven Basart, Norman Mu, Saurav Kadavath, Frank Wang, Evan Dorundo, Rahul Desai, Tyler Zhu, Samyak Parajuli, Mike Guo, et~al.
\newblock The many faces of robustness: A critical analysis of out-of-distribution generalization.
\newblock In {\em Proceedings of the IEEE/CVF international conference on computer vision}, pages 8340--8349, 2021.

\bibitem{imagenets}
Haohan Wang, Songwei Ge, Zachary Lipton, and Eric~P Xing.
\newblock Learning robust global representations by penalizing local predictive power.
\newblock {\em Advances in Neural Information Processing Systems}, 32, 2019.

\bibitem{imageneta}
Dan Hendrycks, Kevin Zhao, Steven Basart, Jacob Steinhardt, and Dawn Song.
\newblock Natural adversarial examples.
\newblock {\em CVPR}, 2021.

\bibitem{svhn}
Yuval Netzer, Tao Wang, Adam Coates, Alessandro Bissacco, Bo~Wu, and Andrew~Y. Ng.
\newblock Reading digits in natural images with unsupervised feature learning.
\newblock In {\em NeurIPS Workshop on Deep Learning and Unsupervised Feature Learning 2011}, 2011.

\bibitem{places}
Bolei Zhou, Agata Lapedriza, Aditya Khosla, Aude Oliva, and Antonio Torralba.
\newblock Places: A 10 million image database for scene recognition.
\newblock {\em IEEE Transactions on Pattern Analysis and Machine Intelligence}, 40(6):1452--1464, 2018.

\end{thebibliography}
